\documentclass[3p,a4paper]{elsarticle}
\usepackage[utf8]{inputenc}
\usepackage{amssymb}
\usepackage{pifont}
\usepackage{physics}
\newcommand{\xmark}{\ding{55}}%

\title{Physics-Guided, Physics-Informed, and Physics-Encoded Neural Networks in Scientific Computing}

\author[mythirdaddress]{Salah A. Faroughi \corref{mycorrespondingauthor}}
\author[mythirdaddress]{Nikhil M. Pawar}
\author[mythirdaddress,mysecondaddress]{Célio Fernandes}
\author[my6address]{Maziar Raissi}
\author[myfifthaddress]{Subasish Das}
\author[myforthaddress]{Nima K. Kalantari}
\author[mythirdaddress]{Seyed Kourosh Mahjour}
\cortext[mycorrespondingauthor]{Corresponding author: salah.faroughi@txstate.edu}

\address[mythirdaddress]{Geo-Intelligence Laboratory, Ingram School of Engineering, Texas State University, San Marcos, Texas, 78666, USA}
\address[mysecondaddress]{Associate Laboratory of Intelligent Systems, Institute for Polymers and Composites, Polymer Engineering Department, School of Engineering, University of Minho, Campus of Azur\'em, 4800-058 Guimar\~aes, Portugal}
\address[my6address]{
Department of Applied Mathematics, University of Colorado Boulder, Boulder, 610101, Colorado, USA}
\address[myfifthaddress]{Artificial Intelligence in Transportation Lab, Ingram School of Engineering, Texas State University, San Marcos, Texas, 78666, USA}
\address[myforthaddress]{Computer Science and Engineering Department, Texas A\&M University, College Station, Texas, 77843, USA}

\date{\today}

\usepackage{mwe}
\usepackage[font=normalsize,labelfont={bf}]{caption}

\usepackage{setspace} 

\makeatletter
\def\ps@pprintTitle{%
 \let\@oddhead\@empty
 \let\@evenhead\@empty
 \def\@oddfoot{\centerline{\thepage}}%
 \let\@evenfoot\@oddfoot}
\makeatother

\usepackage{amssymb}
\usepackage[]{amsmath}
\usepackage{upgreek}

\usepackage{pgfplots}
\pgfplotsset{compat=1.5}
\usepgfplotslibrary{units}
\usetikzlibrary{spy}
\usetikzlibrary{pgfplots.groupplots}
\usepackage{comment}
\usepackage{epstopdf}
\usepackage{units}
\usepackage{lscape}
\usepackage{booktabs}
\usepackage{gensymb}
\usepackage{multirow}
\usepackage{xcolor}
\usepackage{graphicx}
\usepackage{caption}
\usepackage{subcaption}

\usepackage{floatrow} 
\usepackage[flushleft]{threeparttable}
\usepackage{tikz}
\usepackage{placeins}
\usepackage{color}
\usepackage{hyperref}

\usepackage{natbib}

\usepackage{stackengine}

\usepackage{array}
\usepackage{tabularx}
\usepackage{xcolor}  
\usepackage{soul}

\begin{document}

\begin{abstract}

Recent breakthroughs in computing power have made it feasible to use machine learning and deep learning to advance scientific computing in many fields, including fluid mechanics, solid mechanics, materials science, etc. Neural networks, in particular, play a central role in this hybridization. Due to their intrinsic architecture, conventional neural networks cannot be successfully trained and scoped when data is sparse, which is the case in many scientific and engineering domains. Nonetheless, neural networks provide a solid foundation to respect physics-driven or knowledge-based constraints during training. Generally speaking, there are three distinct neural network frameworks to enforce the underlying physics: (i) physics-guided neural networks (PgNNs), (ii) physics-informed neural networks (PiNNs), and (iii) physics-encoded neural networks (PeNNs). These methods provide distinct advantages for accelerating the numerical modeling of complex multiscale multi-physics phenomena. In addition, the recent developments in neural operators (NOs) add another dimension to these new simulation paradigms, especially when the real-time prediction of complex multi-physics systems is required. All these models also come with their own unique drawbacks and limitations that call for further fundamental research. This study aims to present a review of the four neural network frameworks (i.e., PgNNs, PiNNs,  PeNNs, and NOs) used in scientific computing research. The state-of-the-art architectures and their applications are reviewed, limitations are discussed, and future research opportunities in terms of improving algorithms, considering causalities, expanding applications, and coupling scientific and deep learning solvers are presented. This critical review provides researchers and engineers with a solid starting point to comprehend how to integrate different layers of physics into neural networks.

\end{abstract}


\maketitle

\section{Introduction}
\label{sec:Intro}

Machine learning (ML) and deep learning (DL) are becoming the key technologies to advance scientific research and computing in a variety of fields, such as fluid mechanics \cite{vinuesa2022enhancing}, solid mechanics \cite{Raabe2021}, materials science \cite{Kim2021}, etc. The emergence of multiteraflop machines with thousands of processors for scientific computing combined with advanced sensory-based experimentation has heralded an explosive growth of structured and unstructured heterogeneous data in science and engineering fields. ML and DL approaches were first introduced to scientific computing to address the lack of efficient data modeling procedures, which prevented scientists from interacting quickly with heterogeneous and complex data \citep{dino2020impact}. These approaches show transformative potential because they enable the exploration of vast design spaces, the identification of multidimensional connections, and the management of ill-posed issues \cite{im2021surrogate, karniadakis2021physics, arman2022compfluids}. However, conventional ML and DL methods are unable to extract interpretative information and expertise from complex multidimensional data. They may be effective in mapping observational or computational data, but their predictions may be physically irrational or dubious, resulting in poor generalization \cite{innes2019differentiable, brunton2020machine, cai2022physics}. For this reason, scientists initially considered these methodologies as a magic black box devoid of a solid mathematical foundation and incapable of interpretation. Notwithstanding, learning techniques constitute a new paradigm for accurately solving scientific and practical problems orders of magnitude faster than conventional solvers.

\begin{figure}[]
    \centering
    \includegraphics[width=.99\linewidth]{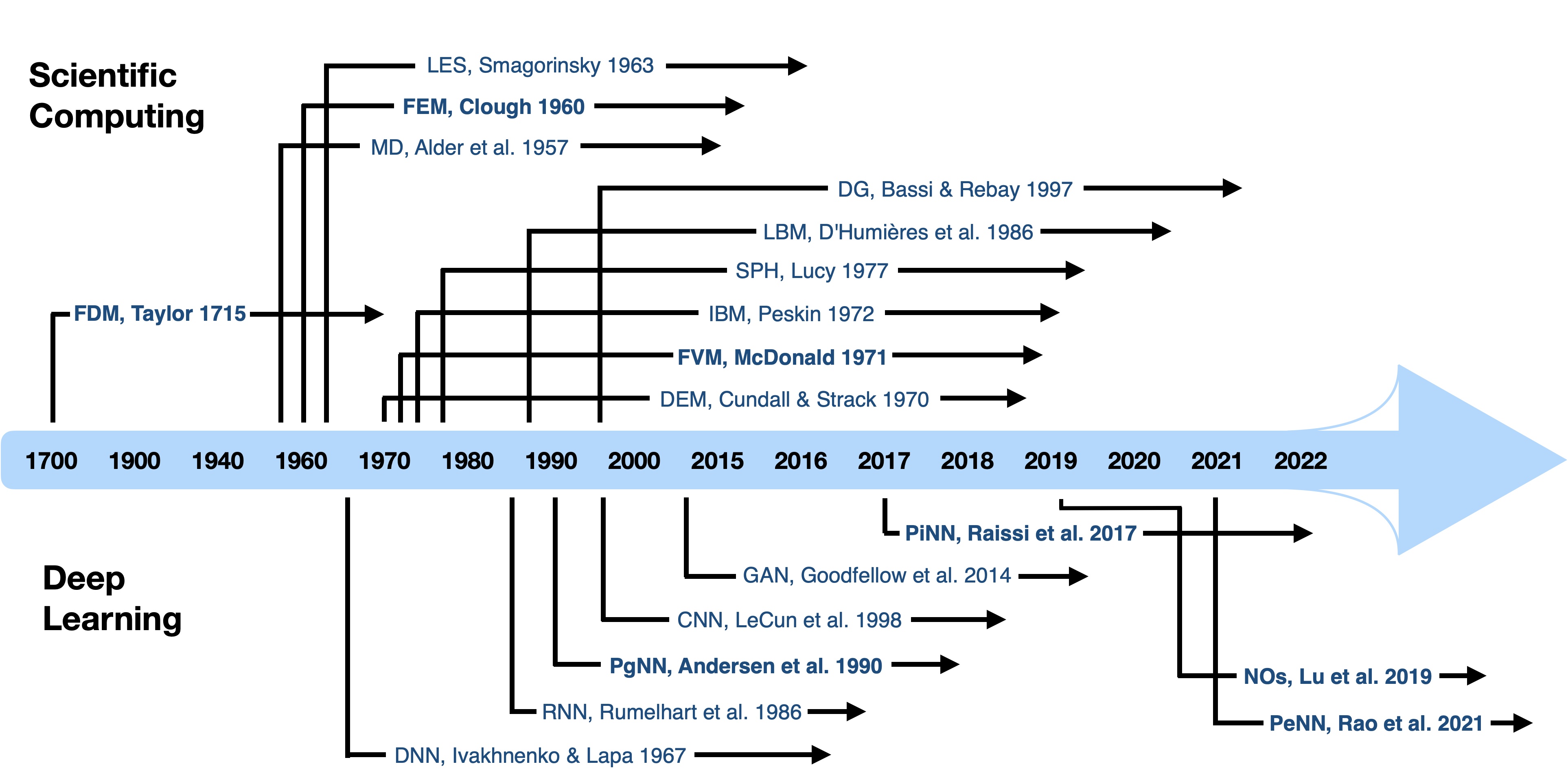}
     \caption{A schematic representation of the history of development, including only seminal works, for scientific computing and DL approaches. For scientific computing, the following are listed: Finite Difference Method (FDM) \cite{taylor1715methodus}, Molecular Dynamics (MD) \cite{alder1957phase}, Finite Element Method (FDM)  \cite{clough1960finite}, Large Eddy Simulation (LES) \cite{smagorinsky1963general}, Discrete Element Method (DEM)  \cite{Cundall197947}, Finite Volume Method (FVM)  \cite{mcdonald1971computation}, Immersed Boundary Method (IBM)  \cite{peskin1972flow}, Smoothed Particle Hydrodynamics (SPH) \cite{lucy1977numerical}, Lattice Boltzmann Method (LBM) \cite{d1986lattice}, and Discontinuous Galerkin (DG) \cite{bassi1997high}. For deep learning, the following are listed: Deep Neural Network (DNN) \cite{ivakhnenko1967cybernetics},  Recurrent Neural Network (RNN) \cite{rumelhart1986learning}, Physics-guided Neural Network (PgNN) \cite{andersen1990artificial}, Convolutional Neural Network (CNN) \cite{lecun1998gradient}, Generative Adversarial Network (GAN) \cite{goodfellow2014generative}, Physics-informed Neural Network (PiNN) \cite{raissi2017physics},  Neural Operators (NOs) \cite{lu2019deeponet}, and Physics-encoded Neural Network (PeNN) \cite{rao2021hard}.    }
         \label{hist}
\end{figure}

Deep learning (i.e., neural networks mimicking the human brain) and scientific computing share common historical and intellectual links that are normally unrealized, e.g., differentiability \cite{innes2019differentiable}. Figure \ref{hist} shows a schematic representation of the history of development for a plethora of scientific computing and DL approaches (only seminal works are included). In the last decade, breakthroughs in DL and computing power have enabled the use of DL in a broad variety of scientific computing, especially in fluid mechanics \cite{vinuesa2022enhancing,cai2022physics,Kutz2017}, solid mechanics \cite{Raabe2021,Shi2019,haghighat2021physicsa}, and materials science  \cite{Pilania2013,Butler2018,Brunton2019}, albeit at the cost of accuracy and loss of generality \citep{bedolla2020machine}. These data-driven methods are routinely applied to fulfill one of the following goals: (i) accelerate direct numerical simulations using surrogate modeling \cite{Kochkov2021}, (ii) accelerate adjoint sensitivity analysis \cite{innes2019differentiable}, (iii) accelerate probabilistic programming \cite{tran2017deep}, and (iv) accelerate inverse problems \cite{jin2017deep}. For example, in the first goal, the physical parameters of the system (e.g., dimensions, mass, momentum, temperature, etc.) are used as inputs to predict the next state of the system or its effects (i.e., outputs), and in the last goal, the outputs of a system (e.g., a material with targeted properties) are used as inputs to infer the intrinsic physical attributes that meet the requirements (i.e., the model's outputs). To accomplish these goals, lightweight DL models can be constructed to partially or fully replace a bottleneck step in the scientific computing processes \citep{bedolla2020machine,lai2022machine, faroughi2022meta}. 

Due to the intrinsic architecture of conventional DL methods, their learning is limited to the scope of the datasets with which the training is conducted (e.g., specific boundary conditions, material types, spatiotemporal discretization, etc.), and inference cannot be successfully scoped under any unseen conditions (e.g., new geometries, new material types, new boundary conditions, etc.). Because the majority of the scientific fields are not (big) data-oriented domains and cannot provide comprehensive datasets that cover all possible conditions, these models trained based on sparse datasets are accelerated but not predictive \cite{faroughi2022meta}. Thus, it is logical to leverage the wealth of prior knowledge, the underlying physics, and domain expertise to further constrain these models while training on available, sparse data points. Neural networks (NNs) are better suited to digest physical-driven or knowledge-based constraints during training. Based on how the underlying physics is incorporated, the authors categorized neural network applications in scientific computing into three separate types: (i) physics-guided neural networks (PgNNs), (ii) physics-informed neural networks (PiNNs), and (iii) physics-encoded neural networks (PeNNs). 

In PgNN-based models,  off-the-shelf supervised DL techniques are used to construct surrogate mappings between formatted inputs and outputs that are generated using experiments and computations in a controlled setting and curated through extensive processes to ensure compliance with physics principles and fundamental rules \cite{faroughi2022meta}. Such models require a rich and sufficient dataset to be trained and used reliably. A PgNN-based model maps a set of inputs $\textbf{x}$ to a related set of outputs $\textbf{y}$ using an appropriate function $\textbf{F}$ with unknown parameters $\textbf{w}$ such that $\textbf{y}=F(\textbf{x};\textbf{w})$. By specifying a particular structure for \textit{F}, a data-driven approach generally attempts to fine-tune the parameters $\textbf{w}$ so that the overall error between true values, $\hat{\textbf{y}}$, and those from model predictions, $\textbf{y}$,  is minimized \cite{arman2022compfluids}. 
For complex physical systems, the data is likely sparse due to the high cost of data acquisition \cite{Lienen2022}. The vast majority of state-of-the-art PgNNs lack robustness and fail to  fulfill any guarantees of generalization (i.e., interpolation \cite{raissi2017physics, hasson2020direct} and extrapolation \cite{li2021physics}). To remediate this issue, PiNNs have been introduced to perform supervised learning tasks while obeying given laws of physics in the form of general non-linear differential equations \cite{raissi2019physics,cai2022physics,nabian2020adaptive, cuomo2022scientific, karniadakis2021physics}. 

The PiNN-based models respect the physical laws by incorporating a weakly imposed loss function consisting of the residuals of physics equations and boundary constraints. They leverage automatic differentiation \cite{baydin2015} to differentiate the neural network outputs with respect to their inputs (i.e., spatiotemporal coordinates and model parameters). By minimizing the loss function, the network can closely approximate the solution \cite{rao2020,faroughi2022physics}. As a result, PiNNs lay the groundwork for a new modeling and computation paradigm that enriches DL with long-standing achievements in mathematical physics \cite{raissi2017physics, raissi2019physics}. The PiNN models  face a number of limitations relating to theoretical considerations (e.g., convergence and stability \cite{mcclenny2020self,karniadakis2021physics,yadav2022distributed}) and implementation considerations
(e.g., neural network design, boundary condition management, and optimization aspects) \cite{rao2021hard,cai2022physics}. In addition, in cases where the explicit form of differential equations governing the complex dynamics is not fully known \textit{a priori}, PiNNs encounter serious limitations \cite{bauer2021digital}. For such cases, another family of DL approaches known as physics-encoded neural networks (PeNN) has been proposed \cite{rao2021hard}.  

The PeNN-based models leverage advanced architectures to address issues with data sparsity and the lack of generalization encountered by both PgNNs and PiNNs models. PeNN-based models forcibly encode the known physics into their core architecture (e.g., NeuralODE \cite{chen2018neural}). By construction, PeNN-based models extend the learning capability of a neural network from instance learning (imposed by PgNN and PiNN architectures) to continuous learning  \cite{chen2018neural}. The encoding mechanisms of the underlying physics in PeNNs are fundamentally different from those in PiNNs \cite{chung2016deep,mattheakis2019physical}, although they can be integrated  to  achieve the desired non-linearity of the model. In comparison to PgNNs and PiNNs, the neural networks generated by the PeNN paradigm offer better performance against data sparsity and model generalizability \cite{rao2021hard}. 

There is another family of supervised learning methods that do not fit well under PgNN, PiNN, and PeNN categories as defined above. These models, dubbed as neural operators, learn the underlying linear and nonlinear continuous operators, such as integrals and fractional Laplacians, using advanced architectures (e.g., DeepONet \cite{lu2019deeponet,lu2021learning}). The data-intensive learning procedure of  a neural operator may resemble the PgNN-based models learning, as both  enforce the physics of the problem using labeled input-output dataset pairs. However, a neural operator is very different from a PgNN-based model that lacks generalization properties due to under-parameterization. A neural operator can be combined with PiNN and PeNN methods to train a model that can learn complex non-linearity in physical systems with extremely high generalization accuracy \cite{li2021physics}. The robustness of neural operators for applications requiring real-time inference is a distinguishing characteristic \cite{goswami2022physicsReview}.

This review paper is primarily intended for the scientific computing community interested in the application of neural networks in computational fluid and solid mechanics. It discusses the general architectures, advantages, and limitations of PgNNs, PiNNs, PeNNs, and neural operators and reviews the most prominent applications of these methods in fluid and solid mechanics. The remainder of this work is structured as follows: In Section 2, the potential of PgNNs to accelerate scientific computing is discussed. Section 3 provides an overview of PiNNs and discusses their potential to advance PgNNs. In Section 4, several leading PeNN architectures to address critical limitations in PgNNs and PiNNs are  discussed. Section 4 reviews the recent developments in neural operators. Finally, in Section 6, an outlook for future research directions is provided.

\section{Physics-guided Neural Networks, PgNNs}

\begin{figure}[htp]
    \centering
    \includegraphics[width=.98\linewidth]{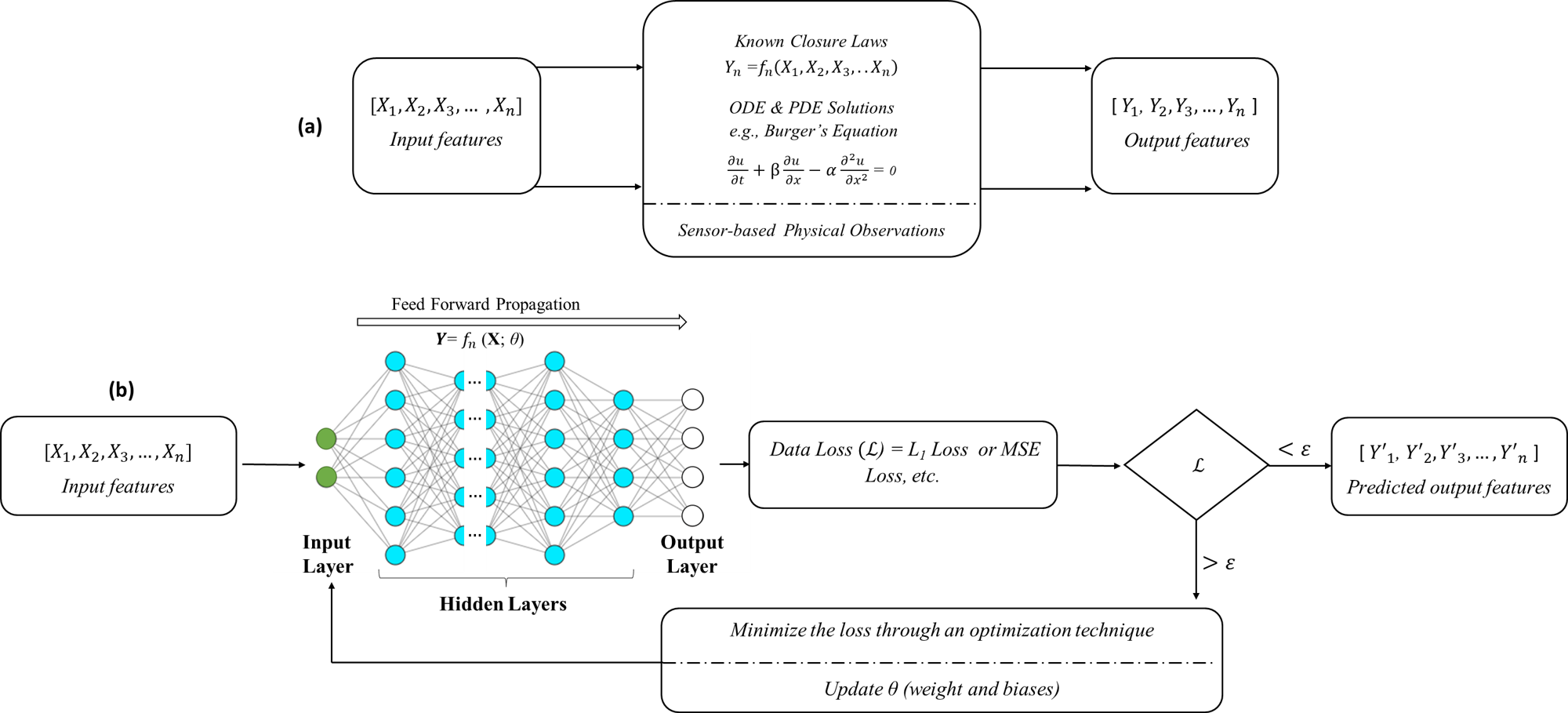}
     \caption{A schematic architecture of PgNNs. Panel (a) shows the typical generation of training datasets using known closure laws, direct numerical simulation of PDEs and ODEs, or experimentation to comply with physical principles. Panel (b) shows the architecture of a PgNN model consisting of a simple feed-forward neural network (which can be replaced with any other network type). The loss function made of  $L_1$, $L_2$ regularization, MSE, or other user-defined error functions is minimized iteratively in the training phase. $\theta$ is the learnable parameter corresponding to weights/biases in the neural network that can be learned simultaneously while minimizing the loss function.}
         \label{PgNN}
\end{figure}

PgNNs use off-the-shelf supervised DL models to statistically learn the known physics of a desired phenomenon by extracting features or attributes from  training datasets obtained through well-controlled experiments and computations \cite{lecun2015deep}. PgNNs consist of one or a combination of Multilayer Perceptron (MLP, alternatively called artificial neural networks, ANN, or deep neural networks, DNN, in different studies relevant to this review) \cite{lecun2015deep}, CNN \cite{lecun2015deep}, RNN \cite{lecun2015deep}, GAN \cite{creswell2018generative}, and graph neural networks (GRNN) \cite{ scarselli2008graph}. Although GAN models are categorized as unsupervised learning, they can be classified as PgNNs, in the context of this paper, because their underlying training is framed as a supervised learning problem \cite{creswell2018generative,goodfellow2020generative}.   A schematic representation of a sample PgNN architecture is illustrated in Fig.~\ref{PgNN}. Any physical problem includes a set of independent features or input features as $\textbf{x} = [{X_1},{X_2},{X_3},...,{X_n}]$  and a set of dependent variables or desired outputs as $\textbf{y} = [{Y_1},{Y_2},{Y_3},...,{Y_n}]$. The data describing this physical phenomenon can be generated by experimentation (e.g., sensor-based observation, etc.), closure laws (e.g., Fourier’s law, Darcy's law, drag force, etc.), or the solution of governing ordinary differential equations (ODE) and/or partial differential equations (PDE), e.g., Burger's equation, Navier-Stokes equations, etc. The dependent variables and independent features thus comply with physics principles, and the trained neural network is guided inherently by physics throughout training. 

In PgNNs, the neurons in each layer are connected to the neurons in the next layer through a set of weights. The output of each node is obtained by applying an activation function (e.g., rectified linear unit (ReLU), Tanh, Sigmoid, Linear, etc.) to the weighted sum of the outputs of the neurons in the preceding layer plus an additional bias \cite{rasamoelina2020review}. This procedure sequentially obtains the output of the neurons in each layer, starting with the input. This process is typically called forward propagation.  A loss function (or, alternatively, a cost function) is subsequently defined and calculated in order to evaluate the accuracy of the prediction. Commonly used loss functions for regression are L1 \cite{he2020extract} and mean-squared-error (MSE) \cite{he2020extract}. The next step in training involves error backpropagation, which calculates the partial derivatives/gradients of the cost function with respect to weights and biases (i.e., $\theta$ as shown in Fig.~\ref{PgNN}). Finally, an optimization technique, such as gradient descent \cite{li2014efficient}, stochastic gradient descent \cite{li2014efficient}, or mini-batch gradient descent \cite{li2014efficient}, is used to minimize the loss function and simultaneously compute and update $\theta$ parameters using the calculated gradients from the backpropagation procedure. The process is iterated until the desired level of accuracy is obtained for a PgNN. 

In recent years, PgNN has been extensively used to accelerate computational fluid dynamics (CFD) \cite{Huang2021}, computational solid mechanics \cite{kumar2022machine}, and multi-functional material  designs \cite{guo2021artificial}. It has been employed in all computationally expensive and time-consuming components of scientific computing, such as (i) pre-processing \cite{Zhang2020,Huang2021,Tingfan2022}, e.g., mesh generation; (ii) discretization and modeling \cite{mendizabal2020simulation,lu2021machine,li2021graph}, e.g., Finite Difference (FDM), Finite Volume (FVM), Finite Element (FEM), Discrete Element Method (DEM), Molecular Dynamics (MD), etc.; and (iii) post-processing, e.g., output assimilation and visualization \cite{menke2021upscaling,cheng2022generalised, zawawi2018review}. These studies are arranged (i) to train shallow networks on small datasets to replace a bottleneck (i.e., a computationally expensive step) in conventional forward numerical modeling, e.g., drag coefficient calculation in concentrated complex fluid flow modeling \citep{faroughi2022meta,he2019supervised,zhu2020machine,roriz2021ml,loiro2021digital}; or (ii) to train relatively deep networks on larger datasets generated for a particular problem, e.g., targeted sequence design within the coarse-grained polymer genome \citep{webb2020targeted}. These networks acknowledge the physical principles upon which the training data is generated and accelerate the simulation process \citep{zawawi2018review,faroughi2022meta}. 

Although the training of PgNNs appears to be straightforward, generating the data by tackling the underlying physics for complex physical problems could require a substantial computational cost \cite{karniadakis2021physics, haghighat2021physicsa}. Once trained, a PgNN can significantly accelerate the computation speed for the phenomena of interest. It is worth noting that while a PgNN model may achieve a good accuracy on the training set based on numerous attempts, it is more likely to memorize the trends, noise, and detail in the training set rather than intuitively comprehend the pattern in the dataset. This is one of the reasons that PgNNs  lose their prediction ability when inferred/tested outside the scope of the training datasets. PgNNs' overfitting can be mitigated in different ways \cite{srivastava2014dropout, bejani2021systematic,ying2019overview} to enhance the predictability of the model within the scope of the training data. In the following subsections, we review the existing literature and highlight some of the most recent studies that applied PgNNs to accelerate different steps in scientific computing for applications in fluid and solid mechanics.

\subsection{Pre-Processing}

Pre-processing is often the most work-intensive component in scientific computing, regardless of the numerical model type (e.g., FEM, FDM, FVM, etc.). The main steps in this component are the disassembly of the domain into small, but finite, parts (i.e., mesh generation, evaluation, and optimization) and the upscaling and/or downscaling of the mesh properties to use a spatiotemporally coarse mesh while implicitly solving for unresolved fine-scale physics. These two steps are time-consuming and require expert-level knowledge; hence, they are potential candidates to be replaced by accelerated PgNN-based models.  

\subsubsection{Mesh Generation}
Mesh generation is a critical step for numerical simulations. \citet{Zhang2020} proposed the automatic generation of an unstructured mesh based on the prediction of the required local mesh density throughout the domain. For that purpose, an ANN was trained to guide a standard mesh generation algorithm. They also proposed extending the study to other architectures, such as CNN or GRNN, for future studies including larger datasets and/or higher-dimensional problems. \citet{Huang2021} adopted a DL approach to identify optimal mesh densities. They generated optimized meshes using classical CFD tools (e.g., Simcenter STAR-CCM+ \cite{cati2022numerical}) and proposed training a CNN to predict optimal mesh densities for arbitrary geometries. The addition of an adaptive mesh refinement version accelerated the overall process without compromising accuracy and resolution. The authors proposed learning optimal meshes (generated by corresponding solvers with adjoint functionality) using ANN, which may be utilized as a starting point in other simulation tools irrespective of the specific numerical approach \cite{Huang2021}. \citet{Tingfan2022} also proposed a mesh optimization method by integrating the moving mesh method with DL in order to solve the mesh optimization problem. With the experiments carried out, a neural network with  high accuracy was constructed to optimize the mesh while preserving the specified number of nodes and topology of the initially given mesh. Using this technique, they also demonstrated that the moving mesh algorithm is independent of the CFD computation \cite{Tingfan2022}. 

In mesh generation, a critical issue has been the evaluation of mesh quality due to a lack of general and effective criteria. \citet{Chen2020} presented a benchmark dataset (i.e., the NACA-Market reference dataset) to facilitate the evaluation of a mesh's quality. They presented GridNet, a technique that uses a deep CNN to perform an automatic evaluation of the mesh's quality. This method receives the mesh as input and conducts the evaluation. The mesh quality evaluation using a deep CNN model trained on the NACA-Market dataset proved to be viable with an accuracy of 92.5 percent \cite{Chen2020}. 

\subsubsection{Cross-scaling Techniques}

It is always desirable to numerically solve a multi-physics problem on a spatiotemporally coarser mesh to minimize computational cost. For this reason, different upscaling \cite{maddu2021stencil, bar2019learning}, downscaling \cite{bernardin2010stochastic}, and cross-scaling \cite{wei2016effect} methods have been developed to determine accurate numerical solutions to non-linear problems across a broad range of length- and time-scales. One viable choice is to use a coarse mesh that reliably depicts long-wavelength dynamics and accounts for unresolved small-scale physics. Deriving the mathematical model (e.g., boundary conditions) for coarse representations, on the other hand, is relatively hard. \citet{bar2019learning} proposed a PgNN model for learning optimum PDE approximations based on actual solutions to known underlying equations. The ANN outputs spatial derivatives, which are then optimized in order to best satisfy the equations on a low-resolution grid. Compared to typical discretization methods (e.g., finite difference), the recommended ANN method was considerably more accurate while integrating the set of non-linear equations at a resolution that was 4 to 8 times coarser \cite{bar2019learning}. The main challenge in this approach, however, is to systematically derive these kinds of solution-adaptive discrete operators. \citet{maddu2021stencil} developed a PgNN, dubbed as STENCIL-NET, for learning resolution-specific local discretization of non-linear PDEs. By combining spatially and temporally adaptive parametric pooling on regular Cartesian grids with knowledge about discrete time integration, STENCIL-NET can accomplish numerically stable discretization of the operators for any arbitrary non-linear PDE. The STENCIL-NET model can also be used to determine PDE solutions over a wider spatiotemporal scale than the training dataset. In their paper, the authors employed STENCIL-NET for long-term forecasting of chaotic PDE solutions on coarse spatiotemporal grids to test their hypothesis. Comparing the STENCIL-NET model to baseline numerical techniques (e.g., fully vectorized WENO \cite{shu1998essentially}), the predictions on coarser grids were faster by up to 25 to 150 times on GPUs and 2 to 14 times on CPUs, while maintaining the same accuracy \cite{maddu2021stencil}. 

Table \ref{Table_pre} reports a non-exhaustive list of recent works that leveraged PgNNs to accelerate the pre-processing part of scientific computing.
These studies collectively concluded that PgNN can be successfully integrated to achieve a considerable speed-up factor in mesh generation, mesh evaluation, and cross-scaling, which are vital for many complex problems explored using scientific computing techniques. The next subsection discusses the potential of PgNN to be incorporated into the modeling components, hence yielding a higher speed-up factor or greater  accuracy. 
 
\begin{table}[]
\centering
\caption {A non-exhaustive list of recent studies that leveraged PgNNs to accelerate the pre-processing part in scientific computing.}
\label{Table_pre}
\begin{tabularx}{\textwidth}{ >{\hsize=.6\hsize}X >{\hsize=0.3\hsize}X>{\hsize=1.2\hsize}X>{\hsize=10\hsize}r}
\toprule  
{Area of application} & {NN Type} & {Objective }  &  {Reference}  \\ 
\midrule
  Mesh Generation & ANN & Generating unstructured mesh & \cite{Zhang2020} \\
\addlinespace
\addlinespace
  & CNN  & Predicting meshes with optimal  density and accelerating meshing process without compromising performance or resolution & \cite{Huang2021}  \\
\addlinespace
\addlinespace
  &  ANN & Generating high quality tetrahedral meshes  & \cite{zhang2021meshingnet3d} \\
\addlinespace
\addlinespace
   & ANN & Developing a mesh generator tool to produce high-quality FEM meshes & \cite{triantafyllidis2002finite} \\
\addlinespace
\addlinespace
   & ANN & Generating  finite element mesh with less complexities  & \cite{srasuay2010mesh} \\
\addlinespace
\addlinespace
 Mesh  Evaluation & CNN & Conducting automatic mesh evaluation and  quality assessment & \cite{Chen2020} 
\\

\addlinespace
\addlinespace
 Mesh Optimisation  & ANN & Optimizing mesh while retaining the same number of nodes and topology as the initially given mesh &  \cite{Tingfan2022} \\
\addlinespace
\addlinespace
Cross-scaling  & ANN  &  Utilizing data-driven discretization to estimate spatial derivatives that are tuned to best fulfill the equations on a low-resolution grid. & \cite{bar2019learning} \\
\addlinespace
\addlinespace
  & ANN \footnotesize(STENCIL-NET)  &  Providing solution-adaptive discrete operators to predict PDE solutions on bigger spatial domains and for longer time frames than it was trained  & \cite{maddu2021stencil} \\

\bottomrule
    \end{tabularx}
\end{table}

\subsection{Modeling and Post-processing}

\subsubsection{PgNNs for Fluid Mechanics}

PgNN has gained considerable attention from the fluid mechanics' community. The study by \citet{lee1993fluid} on estimating fluid properties using ANN was among the first studies that applied PgNN to fluid mechanics. Since then, the application of PgNNs in fluid mechanics has been extended to a wide range of applications, e.g., laminar and turbulent flows, non-Newtonian fluid flows, aerodynamics, etc., especially to speed up the traditional computational fluid dynamics (CFD) solvers. 

For incompressible laminar flow simulations, the numerical procedure to solve Navier–Stokes equations is considered as the main bottleneck. To alleviate this issue, PgNNs have been used as a part of the resolution process. For example, \citet{Yang2016} proposed a novel data-driven projection method using an ANN to avoid iterative computation of the projection step in grid-based fluid simulations. The efficiency of the proposed data-driven projection method was shown to be significant, especially in large-scale fluid flow simulations. \citet{Jonathan2016} used a CNN for predicting the numerical solutions to the inviscid Euler equations for fluid flows. An unsupervised training that incorporates multi-frame information was proposed to improve long-term stability. The CNN model produced very stable divergence-free velocity fields with improved accuracy when compared to the ones obtained by the commonly used Jacobi method \cite{Jacobs1980}. \citet{chen2019u} later developed a U-net-based architecture, a particular case of a CNN model, for the prediction of velocity and pressure field maps around arbitrary 2D shapes in laminar flows. The CNN model is trained with a dataset composed of random shapes constructed using Bézier curves and then by solving Navier-Stokes equations using a CFD solver. The predictive efficiency of the CNN model  was also assessed on unseen shapes, using \textit{ad hoc} error functions, specifically, the MSE levels for these predictions were found to be in the same order of magnitude as those obtained on the test subset, i.e., between $1.0\times 10^{-5}$ and $5.0\times 10^{-5}$ for both pressure and velocity, respectively. 

\begin{figure}[h]
    \centering
    \includegraphics[width=0.99\linewidth]{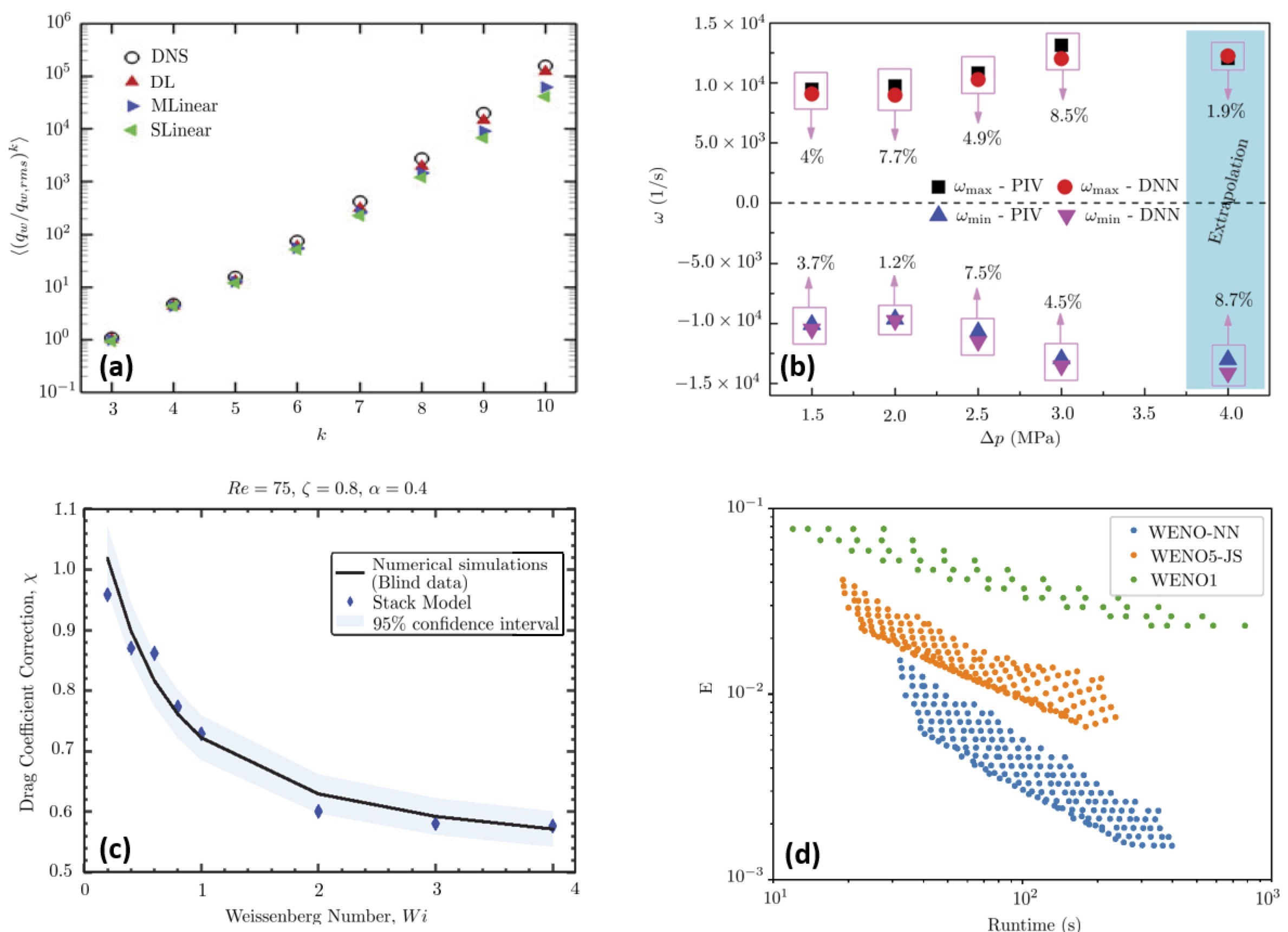}
    \caption{Panel (a) shows a comparison of the high-order moments of the heat flux data obtained from DNS of turbulent heat transfer and predictions obtained by SLinear, MLinear, and CNN (i.e., PgNN) models developed by \citet{Kim2020}. Notice that $q_{w,rms}$ is the root-mean-squared-error (RMSE) of $q_w$, $k$ denotes the index of the weights in the network and the angle bracket denotes the average over all test points. Panel (b) shows the comparison of a PgNN model and PIV technique for the prediction of the swirling flow field in a combustor \cite{WANG2022302}. The changes in maximum and minimum vorticity, $\omega$ (1/s), in a swirling flow field are shown for several pressure drops, $\Delta p$ (MPa). Panel (c) shows the performance of the PgNN model developed by \citet{faroughi2022meta} against the blind dataset generated to predict the drag coefficient of a spherical particle translating in a Giesekus fluid at Reynolds number $Re = 75$, retardation ratio $\zeta = 0.8$ and mobility parameter $\alpha = 0.4$. Panel (d) shows a comparison of the $L_2^2$ error ($E$) and simulation run time of WENO-NN, weighted ENO-Jiang Shu (WENO5-JS) scheme convergent at fifth order, and weighted ENO (WENO1) scheme convergent at first order, to simulate shock wave interactions \citet{Stevens2020}.}
    \label{fig:PgNN}
\end{figure}

Moving from laminar to turbulent flow regimes, PgNNs have been extensively used for the formulation of turbulence closure models \cite{Deng2019}. \citet{Ling2016} used a feed-forward MLP and a specialized neural network to predict Reynolds-averaged Navier–Stokes (RANS) and Large Eddy Simulation (LES) turbulence problems. Their specialized neural network embeds Galilean invariance \cite{levy1971galilei} using a higher-order multiplicative layer. The performance of this model was compared with that of MLP and ground truth simulations. They concluded that the specialized neural network can predict the anisotropy tensor on an invariant tensor basis, resulting in significantly more accurate predictions than MLP. \citet{Maulik2019} presented a closure framework for subgrid modeling of Kraichnan turbulence \cite{Kraichnan1967}. To determine the dynamic closure strength, the proposed framework used an implicit map with inputs as grid-resolved variables and eddy viscosities. Training an ANN with extremely subsampled data obtained from high-fidelity direct numerical simulations (DNSs) yields the optimal map. The ANN model was found to be successful in imbuing the decaying turbulence problem with dynamic kinetic energy dissipation, allowing accurate capture of coherent structures and inertial range fidelity.
Later, \citet{Kim2020} used simple linear regression, SLinear, multiple linear regression, MLinear, and a CNN to predict the turbulent heat transfer (i.e., the wall-normal heat flux, $q_w$) using other wall information, including the streamwise wall-shear stress, spanwise wall-shear stress or streamwise vorticity, and pressure fluctuations, obtained by DNSs of a channel flow (see Fig.~\ref{fig:PgNN}(a)). The constructed network was trained using adaptive moment estimation (ADAM) \cite{kingma2014adam,hoang2020image}, and the grid searching method \cite{priyadarshini2021novel,sun2021improved} was performed to optimize the depth and width of the CNN. Their finding showed that the PgNN model is less sensitive to the input resolution, indicating its potential as a good heat flux model in turbulent flow simulation. \citet{Yousif2022} also proposed an efficient method for generating turbulent inflow conditions based on a PgNN formed by a combination of a multiscale convolutional auto-encoder with a subpixel convolution layer (MSCSP-AE) \cite{shi2016real,talab2019super} and long short-term memory (LSTM) \cite{huang2015bidirectional,sherstinsky2020fundamentals} model. The proposed model was found to have the capability to deal with the spatial mapping of turbulent flow fields.

PgNNs have also been applied in the field of aerodynamics. \citet{Kou2021} presented a review paper on typical data-driven methods, including system identification, feature extraction, and data fusion, that have been employed to model unsteady aerodynamics. The efficacy of those data-driven methods is described by several benchmark cases in aeroelasticity. \citet{WANG2022302} described the application of ANN to the modeling of the swirling flow field in a combustor (see Fig.~\ref{fig:PgNN}(b)). Swirling flow field data from particle image velocimetry (PIV) was used to train an ANN model. The trained PgNN model was successfully tested to predict the swirling flow field under unknown inlet conditions. \citet{CHOWDHARY2022115396} studied the efficacy of combining ANN models with projection-based (PB) model reduction techniques \cite{bond2008guaranteed,beli2018projection} to develop an ANN-surrogate model for computationally expensive, high-fidelity physics models, specifically for complex hypersonic turbulent flows. The surrogate model was used to perform Bayesian estimation of freestream conditions and parameters of the SST (shear stress transport) turbulence model. The surrogate model was then embedded in the high-fidelity (Reynolds-averaged Navier–Stokes) flow simulator, using shock-tunnel data. \citet{SIDDIQUI2022103706} developed a non-linear data-driven model, encompassing Time Delay Neural Networks (TDNN), for a pitching wing. The pitch angle was considered as the input to the model, while the lift coefficient was considered as the output. The results showed that the trained models were able to capture the non-linear aerodynamic forces more accurately than linear and semi-empirical models, especially at higher offset angles. \citet{Wang2022} also proposed a multi-fidelity reduced-order model based on multi-task learning ANNs to efficiently predict the unsteady aerodynamic performance of an iced airfoil. The results indicated that the proposed model  achieves higher accuracy and better generalization capability compared with single-fidelity and single-task modeling approaches.

The simulation of complex fluid flows, specifically using fluids that exhibit viscoelastic nature and non-linear rheological behaviors, is another topic where PgNNs have been applied \cite{pawar2022complex, fernandes2022advanced}. The dynamics  of these fluids are generally governed by non-linear constitutive equations that lead to stiff numerical problems \cite{faroughi2020closure,Fernandes2019}. \citet{faroughi2022meta} developed a PgNN model to predict the drag coefficient of a spherical particle translating in viscoelastic fluids (see Fig.~\ref{fig:PgNN}(c)). The PgNN considered a stacking technique (i.e., ensembling Random Forrest \cite{lin2017ensemble}, Extreme Gradient Boosting \cite{chen2015xgboost} and ANN models) to digest inputs (Reynolds number, Weissenberg number, viscosity ratio, and  mobility factor considering both Oldroyd-B and Giesekus fluids) and outputs drag predictions based on the individual learner’s predictions and an ANN meta-regressor. The accuracy of the model was successfully checked against blind datasets generated by DNSs. \citet{lennon2022MLcomplex} also developed a tensor basis neural network (TBNN) allowing rheologists to construct learnable constitutive models that incorporate essential physical information while remaining agnostic to details regarding particular experimental protocols or flow kinematics. The TBNN model incorporates a universal approximator within a materially objective tensorial constitutive framework that, by construction, respects physical constraints, such as frame-invariance and tensor symmetry, required by continuum mechanics. Due to the embedded TBNN, the developed rheological universal differential equation quickly learns simple yet accurate and highly general models for describing the provided training data, allowing a rapid discovery of  constitutive equations.

Lastly, PgNNs have also been extensively used to improve both the accuracy and speed of CFD solvers. \citet{Stevens2020} developed a DL model (the weighted essentially non-oscillatory neural network, WENO-NN) to enhance a finite-volume method used to discretize PDEs with discontinuous solutions, such as the turbulence–shock wave interactions (see Fig.~\ref{fig:PgNN}(d)). \citet{Kochkov2021} used hybrid discretizations, combining CNNs and subcomponents of a numerical solver, to interpolate differential operators onto a coarse mesh with high accuracy. The training of the model was performed within a standard numerical method for solving the underlying PDEs as a differentiable program, and the method allows for end-to-end gradient-based optimization of the entire algorithm. The method learns accurate local operators for convective fluxes and residual terms and matches the accuracy of an advanced numerical solver running at 8 to 10 times finer resolution while performing the computation 40 to 80 times faster. \citet{Cai2022} implemented a least-squares ReLU neural network (LSNN)  for solving the linear advection-reaction problem with a discontinuous solution. They showed that the proposed method outperformed mesh-based numerical methods in terms of the number of DOFs (degrees of freedom). \citet{Haber2022} suggested an auto-encoder CNN to reduce the resolution cost of a scalar transport equation coupled to the Navier–Stokes equations. \citet{Lara2022} proposed to accelerate high-order discontinuous Galerkin methods using neural networks. The methodology and bounds were examined for a variety of meshes, polynomial orders, and viscosity values for the 1D Burgers’ equation. \citet{List2022} employed CNN to train turbulence models to improve under-resolved, low-resolution solutions to the incompressible Navier–Stokes equations at simulation time. The developed method consistently outperforms simulations with a two-fold higher resolution in both spatial and temporal dimensions. For mixing layer cases, the hybrid model on average resembles the performance of three-fold reference simulations, which corresponds to a speed-up of 7.0 times for the temporal layer and 3.7 times for the spatial mixing layer.

Table \ref{Table_PgNN_Fluids} reports a non-exhaustive list of recent studies that leveraged PgNN to model fluid flow problems. These studies collectively concluded that PgNNs can be successfully integrated with CFD solvers or used as standalone surrogate models to develop accurate and yet faster modeling components for scientific computing in fluid mechanics. In the next section, the potential application of PgNNs in computational solid mechanics is discussed.

\begin{table}[]
\centering
\caption {A non-exhaustive list of studies that leveraged PgNNs to model fluid computational flow problems.}
\label{Table_PgNN_Fluids}
\begin{tabularx}{\textwidth}{ >{\hsize=.2\hsize}l >{\hsize=0.15\hsize}X>{\hsize=0.6\hsize}X>{\hsize=0.1\hsize}r}
\toprule  
{Area of application} & {NN Type} & {Objective}  &  {Reference}  \\ 
\midrule
Laminar Flows & CNN & Calculating numerical solutions to the inviscid Euler equations & \cite{Jonathan2016} \\ 
\addlinespace
\addlinespace  
 & CNN & Predicting the velocity and pressure fields around arbitrary 2D shapes & \cite{chen2019u} \\
\addlinespace
\addlinespace  
Turbulent Flows & ANN &  Developing a model for the Reynolds stress anisotropy tensor using high-fidelity simulation data & \cite{Ling2016}
 \\
\addlinespace
\addlinespace  
  & ANN & Modelling of LESs of a turbulent plane jet flow configuration  & \cite{Vollant2017} \\
\addlinespace
\addlinespace  
   &  CNN & Designing and training artificial neural networks based on local convolution filters for LES & \cite{Beck2018} \\
\addlinespace
\addlinespace
   & ANN & Developing subgrid modelling of Kraichnan turbulence & \cite{Maulik2019}  \\
\addlinespace
\addlinespace
   & CNN & Estimating turbulent heat transfer based on other wall information acquired from channel flow DNSs  & \cite{Kim2020} \\
\addlinespace
\addlinespace
  & CNN-LSTM & Generating turbulent inflow conditions with accurate statistics and spectra & \cite{Yousif2022} \\
\addlinespace
\addlinespace
Aerodynamics  & CNN-MLP & Predicting incompressible laminar steady flow field over airfoils & \cite{Sekar2019} \\
\addlinespace
\addlinespace
 & ANN & Developing a high-dimensional PgNN model for high Reynolds number turbulent flows around airfoils & \cite{Zhu2019} \\
\addlinespace
\addlinespace
  & ANN & Modeling the swirling flow field in a combustor & \cite{WANG2022302} \\
\addlinespace
\addlinespace
  & PCA-ANN & Creating surrogate models of computationally expensive, high-fidelity physics models for complex hypersonic turbulent flows & \cite{CHOWDHARY2022115396} \\
\addlinespace
\addlinespace
  & ANN & Predicting unsteady aerodynamic performance of iced airfoil & \cite{Wang2022} \\
\addlinespace
\addlinespace  
  Viscoelastic Flows & ANN & Predicting drag coefficient of a spherical particle translating in viscoelastic fluids & \cite{faroughi2022meta}\\
\addlinespace
\addlinespace
 & ANN  & Constructing learnable constitutive models using a universal approximator within a materially objective tensorial constitutive framework & \cite{lennon2022MLcomplex}\\
\addlinespace
\addlinespace
 Enhance CFD Solvers  & ANN & Developing an improved finite-volume method for simulating PDEs with discontinuous solutions &  \cite{Stevens2020} 
\\
\addlinespace
\addlinespace
  & CNN &  Interpolating differential operators onto a coarse mesh with high accuracy & \cite{Kochkov2021} 
\\
\addlinespace
\addlinespace
  & LSNN &  Solving the linear advection-reaction problem with discontinuous solution & \cite{Cai2022} \\
\addlinespace
\addlinespace
   & CNN & Modeling the scalar transport equation to reduce the resolution cost of forced cooling of a hot workpiece in a confined environment & \cite{Haber2022} \\
\addlinespace
\addlinespace
  & CNN  & Accelerating high order discontinuous Galerkin methods & \cite{Lara2022} \\ 
\addlinespace
\addlinespace
   & CNN & Developing turbulence model to improve under-resolved low-resolution solutions to the incompressible Navier–Stokes equations at simulation time & \cite{List2022} \\ 
\bottomrule
 \end{tabularx}
\end{table}

\subsubsection{PgNNs for Solid Mechanics}

Physics-guided neural networks (PgNNs) have also been extensively adopted by the computational solid mechanics' community. The study by \citet{andersen1990artificial} on welding modeling using ANN was among the first studies that applied PgNN to solid mechanics. Since then, the application of PgNN  has been extended to a wide range of problems, e.g., structural analysis, topology optimization, inverse materials design and modeling, health condition assessment, etc., especially to speed up the traditional forward and inverse modeling methods in   computational mechanics. 

\begin{figure}[htp]
    \centering
    \includegraphics[width=0.99\linewidth]{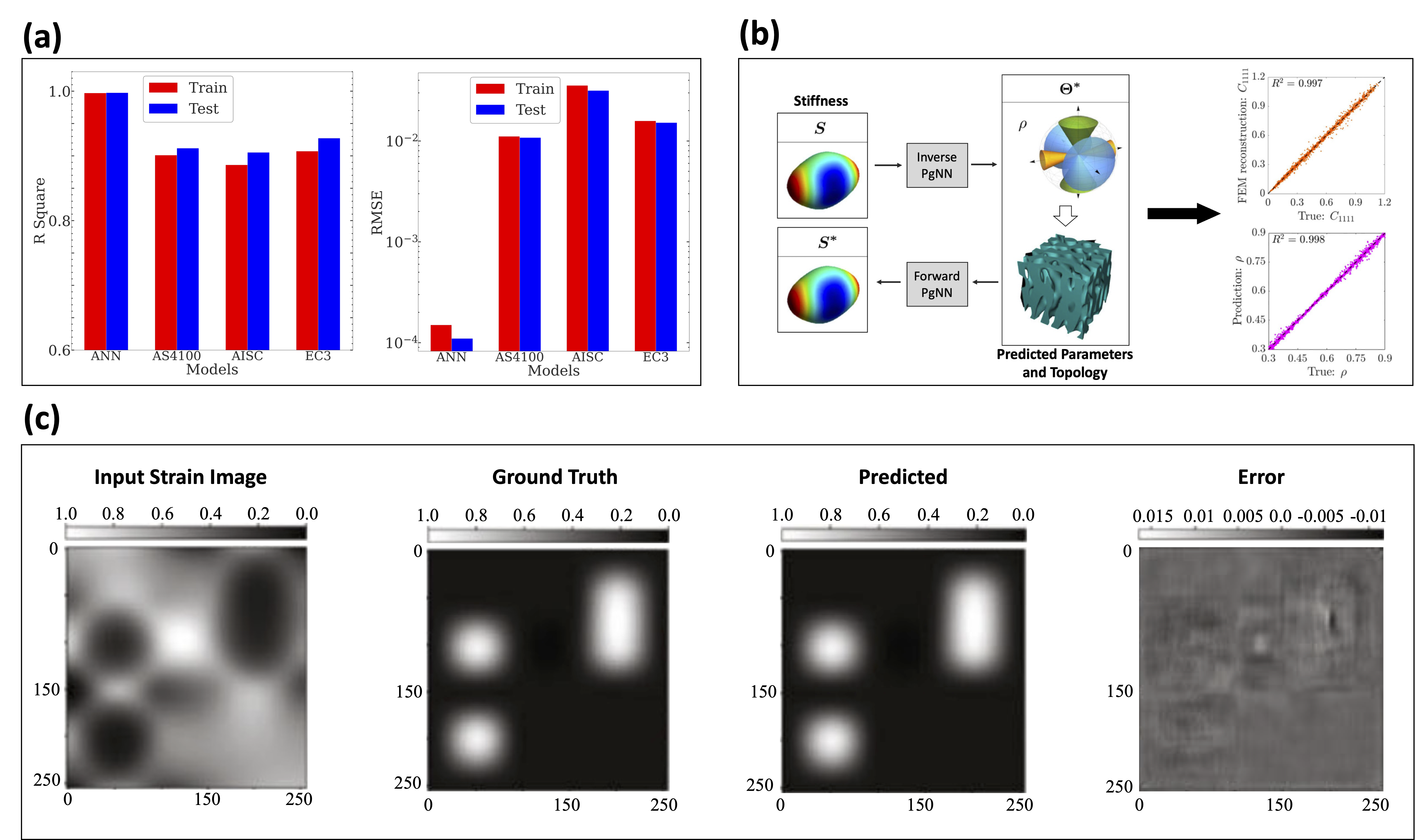}
    \caption{Panel (a) shows a comparison between ANN and other international codes' accuracy (e.g., R-squared and RMSE) to predict the ultimate moment capacities of castellated beams subjected to lateral-distortional buckling (adapted from \citet{hosseinpour2020neural}). Panel (b) shows a two-stage PgNN architecture for predicting the design parameters of meta-materials for spindoid topologies. The first ANN (i.e., inverse PgNN) takes the query stiffness as input and outputs the design parameters. The second ANN (i.e., forward PgNN) takes the predicted design parameters and reconstructs the stiffness to verify inverse network accuracy. R-squared values for prediction of stiffness component  $C_{1111}$ and design parameter $\rho$ are shown in the subsets (adapted from \citet{kumar2020inverse}). Panel (c) shows a comparison between conditional GAN and ground truth made by elastography to predict elastic modulus from strain data (adapted from \citet{ni2021deep}).}
    \label{fig:PgNNsolids}
\end{figure}

In the area of structural analysis, \citet{tadesse2012neural} proposed an ANN for predicting mid-span deflections of a composite bridge with flexible shear connectors. The ANN was tested on six different bridges, yielding a maximum root-mean-squared error (RMSE) of 3.79\%, which can be negligible in practice. The authors also developed ANN-based close-form solutions to be used for rapid prediction of deflection in everyday design. \citet{guneyisi2014prediction} employed ANN to develop a new formulation for the flexural overstrength factor for steel beams. They considered 141 experimental data samples with different cross-sectional typologies to train the model. The results showed a comparable training and testing accuracy of 99 percent, indicating that the ANN model provided a reliable tool to estimate beams' over-strength. \citet{hung2019deep} leveraged ANN to predict the ultimate load factor of a non-linear, inelastic steel truss. They considered a planar 39-bar steel truss to demonstrate the efficiency of the proposed ANN. They used the cross-sections of members as the input and the load-factor as the output. The ANN-based model yielded a high degree of accuracy, with an average loss of less than 0.02, in predicting the ultimate load-factor of the non-linear inelastic steel truss. \citet{chen2019applicationshell} also used ANN to solve a three-dimensional (3D) inverse problem of a collision between an elastoplastic hemispherical metal shell and a rigid impactor. The goal was to predict the position, velocity, and duration of the collision  based on the shell's permanent plastic deformation. For static and dynamic loading, the ANN model predicted the location, velocity, and collision duration with high accuracy. \citet{hosseinpour2020neural} used PgNN for buckling capacity assessment of castellated steel beams subjected to lateral-distortional buckling. As shown in Fig.~\ref{fig:PgNNsolids}(a), the ANN-based model provided higher accuracy than well-known design codes, such as AS4100 \cite{trahair2017behaviour}, AISC \cite{white2006stability}, and EC3 \cite{Eurocode3} for modeling and predicting the ultimate moment capacities.

\color{black}
Topology optimization of materials and meta-materials is yet another domain where PgNNs have been employed \cite{white2019multiscale,zhang2019concurrent}. Topology optimization is a technique that identifies the optimal materials placed inside a prescribed domain to achieve the optimal structural performance \cite{sigmund2013topology}. For example,  \citet{abueidda2020topology} developed a CNN model that performs real-time topology optimization of linear and non-linear elastic materials  under large  and small deformations. The trained model can predict the optimal designs with great accuracy without the need for an iterative process scheme and with very low inference computation time. \citet{yu2019deep} suggested an integrated two-stage technique made up of a CNN-based encoder and decoder (as the first stage) and a conditional GAN (as the second stage) that allows for the determination of a near-optimal topological design. This integration resulted in a model that determines a near-optimal structure in terms of pixel values and compliance with considerably reduced computational time.  \citet{banga20183d} also proposed a 3D encoder-decoder CNN to speed up 3D topology optimization and determine the optimal computational strategy for its deployment. Their findings showed that the proposed model can reduce the overall computation time by 40\% while achieving accuracy in the range of 96\%. \citet{li2019non} then presented a GAN-based non-iterative near-optimal topology optimizer for conductive heat transfer structures trained on black-and-white density distributions. A GAN for low resolution topology was combined with a super resolution generative adversarial network, SRGAN,  \cite{takano2019srgan,nagano2018srgan} for a high resolution topology solution in a two-stage hierarchical prediction-refinement pipeline. When compared to conventional topology optimization techniques, they showed this strategy has clear advantages in terms of computational cost and efficiency. 

PgNN has also been applied for inverse design and modeling in solid mechanics. \citet{messner2020convolutional} employed a CNN to develop surrogate models that estimate the effective mechanical properties of periodic composites. As an example, the CNN-based model was applied to solve the inverse design problem of finding structures with optimal mechanical properties. The surrogate models were in good agreement with well-established topology optimization methods, such as solid isotropic material with penalization (SIMP) \cite{tcherniak2002topology}, and were sufficiently accurate to recover optimal solutions for topology optimization. \citet{lininger2021general} also used CNN to solve an inverse design problem for meta-materials made of thin film stacks. The authors demonstrated the CNN's remarkable ability to explore the large global design space (up to 1012 parameter combinations) and resolve all relationships between meta-material structure and associated ellipsometric and reflectance/transmittance spectra \cite{loper2014complex,lininger2021general}. \citet{kumar2020inverse} proposed a two-stage ANN model, as shown in Fig.~\ref{fig:PgNNsolids}(b), for inverse design of meta-materials. The model generates uniform and functionally graded cellular mechanical meta-materials with tailored anisotropic stiffness and density for spinodoid topologies. The ANN model used in this study is a combination of  two-stage ANN, first ANN (i.e., inverse PgNN) takes query stiffness as input and outputs design parameters, e.g., $\Theta$. The second ANN (i.e., forward PgNN) takes the predicted design parameters as input and reconstructs the stiffness to verify the first ANN results. The prediction accuracy for stiffness and the design parameter was validated against ground truth data for both networks; sample comparisons and their corresponding R-squared values are shown in  Fig.~(\ref{fig:PgNNsolids}(b). \citet{ni2021deep} proposed a combination of representative sampling spaces and conditional GAN, cGAN  \cite{smith2020conditional,balaji2019conditional}, to address the inverse problem of modulus identification in the field of elasticity. They showed that the proposed approach can be deployed with high accuracy, as shown in Fig.~\ref{fig:PgNNsolids}(c) while avoiding the use of costly iterative solvers used in conventional methods, such as the adjoint weighted approach \cite{oberai2003solution}. This model is especially suitable for real-time elastography and high-throughput non-destructive testing techniques used in geological exploration, quality control,  composite material evaluation, etc.

The PgNN models have also been used to overcome some of the computational limitations of multiscale simulations in solid mechanics. This is achieved by (i)  bypassing the costly lower-scale calculations and thereby speeding the macro-scale simulations \cite{kumar2022machine}, or (ii)  replacing a step or the complete simulation with surrogate models \cite{kumar2022machine}. For example,  \citet{liang2018deep} developed an ANN model that takes  finite element-based aorta geometry as input and output the aortic wall stress distribution directly, bypassing FEM  calculation. The difference between the stress calculated by FEM and the one estimated by the PgNN model is practically negligible, while the PgNN model produces output in just a fraction of the FEM computational time. \citet{Mozaffar2019} successfully employed RNN-based surrogate models for material modeling by learning the reversible, irreversible, and history-dependent phenomena that occur when studying material plasticity. \citet{Raabe2021} used a CNN-based solver to predict the local stresses in heterogeneous solids with the highly non-linear material response and mechanical contrast features. When compared to common solvers like FEM, the CNN-based solver offered an acceleration factor of 8300x for elasto-plastic materials. \citet{im2021surrogate} proposed  a PgNN framework to construct a surrogate model for a high-dimensional elasto-plastic FEM model  by integrating an LSTM network with the proper orthogonal decomposition (POD) method \cite{chatterjee2000introduction,liang2002proper}. The suggested POD-LSTM surrogate model allows rapid, precise, and reliable predictions of elasto-plastic structures based on the provided training dataset exclusively. For the first time, \citet{long2021deep} used a CNN to estimate the stress intensity factor of planar cracks. Compared to FEM, the key benefit of the proposed light-weight CNN-based crack evaluation methodology is that it can be installed on an unmanned machine to automatically monitor the severity of a crack in real-time.

Table \ref{Table_PgNN_Solid} reports a non-exhaustive list of recent studies that leveraged PgNNs in solid mechanics and materials design problems. These studies collectively concluded that PgNNs can be successfully integrated with conventional solvers (e.g., FEM solvers) or used as standalone surrogate models to develop  accurate and yet faster modeling components 
for scientific computing in solid mechanics. Albeit, PgNNs come with their own limitations and shortcomings that might compromise solutions under different conditions, as discussed in the next section.

\color{black}

\begin{table}[]
\centering
\caption {A non-exhaustive list of studies that leveraged PgNNs to model solid mechanics problems.}
\label{Table_PgNN_Solid}
\begin{tabularx}{\textwidth}{ >{\hsize=.2\hsize}l >{\hsize=0.15\hsize}X>{\hsize=0.6\hsize}X>{\hsize=0.1\hsize}r}
\toprule  
{Area of application} & {NN Type}  & {Objective }  &  {Reference}  \\ 
\midrule
 Accelerating Simulations & ANN &  Predicting the aortic wall stress distribution using FEM aorta geometry   & \cite{liang2018deep} \\
\addlinespace
\addlinespace
  & RNN  & Developing surrogate models for material modeling by learning reversible, irreversible, and history-dependent phenomena  & \cite{Mozaffar2019}  \\
\addlinespace
\addlinespace
   & CNN & Predicting local stresses in heterogeneous solids with the highly non-linear material response and mechanical contrast features   & \cite{Raabe2021} \\
\addlinespace
\addlinespace
  & CNN  & Estimating stress intensity factor of planar cracks & \cite{long2021deep} \\
\addlinespace
\addlinespace
 Topology Optimization & CNN & Optimizing topology  of linear and non-linear elastic materials under large and small deformations  & \cite{abueidda2020topology} \\
\addlinespace
\addlinespace
  & CNN-GAN &  Determining near-optimal topological design &  \cite{yu2019deep} 
\\
\addlinespace
\addlinespace
  & CNN &  Accelerating 3D topology optimization & \cite{banga20183d} 
\\
\addlinespace
\addlinespace
 & GAN-SRGAN &  Generating near-optimal topologies for conductive heat transfer structures & \cite{li2019non} \\
\addlinespace
\addlinespace
Inverse Modeling  & CNN  &   Estimating effective mechanical properties for periodic composites  & \cite{messner2020convolutional} \\
\addlinespace
\addlinespace
   & CNN  & Solving an inverse design problem for meta-materials made of thin film stacks & \cite{lininger2021general} \\ 
\addlinespace
\addlinespace
 &   cGAN  & Addressing inverse problem of modulus identification in elasticity & \cite{ni2021deep} \\ 
\addlinespace
\addlinespace
  & CVAE  & Designing nano-patterned power splitters for photonic integrated circuits & \cite{ni2021deep} \\ 

\addlinespace
\addlinespace
Structural Elements & ANN & Predicting non-linear buckling load of an imperfect reticulated shell & \cite{zhu2021prediction} \\ 
\addlinespace
\addlinespace
 & ANN & Optimizing dynamic behavior of thin-walled laminated cylindrical shells & \cite{miller2020optimization} \\ 
\addlinespace
\addlinespace
 & ANN & Determining and identifying loading conditions for shell structures  & \cite{chen2019applicationshell} \\ 
\addlinespace
\addlinespace
Structural Analysis  & CNN & Forecasting stress fields in 2D linear elastic cantilevered structures subjected to external static loads & \cite{nie2020stress} \\ 
\addlinespace
\addlinespace
 & ANN &  Estimating the  thickness and length of reinforced walls based on previous architectural  projects  & \cite{pizarro2021structural} \\ 
 \addlinespace
\addlinespace

Condition Assessment  & Auto-encoder-NN & Learning  mapping between vibration characteristics and structural damage   & \cite{pathirage2018structural} \\ 
\addlinespace
\addlinespace
 & CNN &  Providing a real-time crack assessment method  & \cite{jiang2020real} \\ 
\addlinespace
\addlinespace
 & RNN & Nonparametric identification of large civil structures subjected to dynamic loadings    & \cite{perez2019recurrent} \\ 
\addlinespace
\addlinespace
 & CNN & Damage Identification of truss structures using noisy incomplete modal data   & \cite{truong2020effective} \\ 
\bottomrule
    \end{tabularx}
\end{table}

\subsection{PgNNs Limitations} \label{pgnn_limittaion}

Even though PgNN-based models show great potential to accelerate the modeling of non-linear phenomena described by input-output interdependencies, they suffer from several critical limitations and shortcomings. Some of these limitations become more pronounced when the training datasets are sparse. 

\begin{itemize}
\item The main PgNNs' limitation stems from the fact that their training process is solely based on statistics \cite{lecun2015deep}. Even though the training datasets are inherently constrained by physics (e.g., developed by direct numerical simulation, closure laws, and de-noised experimentation), PgNN generates models based on correlations in statistical variations. The outputs (predictions), thus, are naturally physics-agnostic \cite{raissi2017physics,raissi2018deephidden} and may violate the underlying physics \cite{karniadakis2021physics}. 

\item Another important limitation of PgNNs stems from the fact that training datasets are usually sparse, especially in the scientific fields discussed in this paper. When the training data is sparse and does not cover the entire range of underlying physiochemical attributes, the PgNN-based models fail in blind-testing on conditions outside the scope of training \cite{li2021physics}, i.e., they do not offer extrapolation capabilities in terms of  spatiotemporal variables and/or other physical attributes. 

\item PgNN's predictions might be severely compromised, even for inputs within the scope of sparse training datasets \citep{faroughi2022meta}. The lack of interpolation capabilities is more pronounced  in complex and non-linear problems where the range of the physiochemical attributes is extremely wide (e.g., the range of Reynolds numbers from creeping flow to turbulent flow).  

 \item PgNNs may not fully satisfy the initial conditions and boundary conditions using which the training datasets are generated \citep{raissi2017physics}. The boundary conditions and computational domain vary from one problem to another, making the data generation and training process prohibitively costly. In addition, a significant portion of scientific computing research involves inverse problems in which unknown physiochemical attributes of interest are estimated from measurements or calculations that are only indirectly related to these attributes \cite{biros2011large,vogel2002computational,cai2022physics,haghighat2021physicsa}. For instance, in groundwater flow modeling, we leverage measurements of the pressure of a fluid immersed in an aquifer to estimate the aquifer's geometry and/or material characteristics \cite{franssen2009comparison}. Such requirements further complicate the process of developing a simple neural network that is predictive under any conditions. 

\item  PgNNs-based models are not resolution-invariant by construction \cite{li2020fourier}, hence they cannot be trained on a lower resolution and be directly inferred on a higher resolution. This shortcoming is due to the fact that PgNN is only designed to learn the solution of physical phenomena for a single instance (i.e., inputs-outputs).

\item Through the training process, PgNN-based networks learn the input-output interdependencies across the entire dataset. Such a process could potentially consider slight variations in the functional dependencies between different input and output pairs as noise, and produce an average solution. Consequently, while these models are optimal with respect to the entire dataset, they may produce suboptimal results in individual cases.

\item PgNN models may struggle to learn the underlying process when the training dataset is diverse, i.e., when the interdependencies between different input and output pairs are drastically different. Although this issue can be mitigated by increasing the model size,  more data is required to train such a network, making the training costly and, in some cases, impractical.

\end{itemize}

One way to resolve some of the PgNNs' limitations is to generate more training data. However, this is not always a feasible solution due to the high cost of data acquisition. Alternatively, PgNNs can be further constrained by governing physical laws without any prior assumptions, reducing the need for large datasets. The latter is a plausible solution because, in most cases, the physical phenomenon can be fully and partially described using explicit ODEs, PDEs, and/or closure laws. This approach led to the development of a physics-informed neural network \cite{raissi2017physics,raissi2019physics}, which is described and reviewed in the next section.

\section{Physics-informed Neural Networks, PiNNs} 

In scientific computing, physical phenomena are often described using a strong mathematical form consisting of governing differential equations as well as initial and boundary conditions. At each point inside a domain, the strong form specifies the constraints that a solution must meet. The governing equations are usually linear or non-linear PDEs and/or ODEs. Some of the PDEs  are notoriously challenging to solve, e.g., the Navier-Stokes equations to explain a wide range of fluid flows \cite{cai2022physics}, Föppl–von Kármán equations to describe large deflections in solids \cite{randjbaran2015review,randjbaran2015review}, etc. Other important PDE examples are heat equations \cite{triebel2015hybrid}, wave equation \cite{durran2013numerical}, Burgers' equation \cite{prato2004stochastic}, Laplace's equation \cite{medkova2018laplace}, Poisson's equation \cite{genovese2006efficient}, amongst others. This wealth of well-tested knowledge can be logically leveraged to further constrain PgNNs while training on available data points if any \cite{raissi2017physics}. To this end, mesh-free physics-informed neural networks (PiNNs) have been developed \cite{raissi2017physics,raissi2019physics}, quickly extended \cite{jagtap2020conservative, jagtap2021extended}, and extensively deployed in a variety of scientific and applied fields \cite{mahmoudabadbozchelou2021rheology,katsikis2022gentle,salvati2022defect, datta2022physics, nguyen2022modeling, shaier2022data}. Readers are referred to \citet{karniadakis2021physics} and \citet{cai2022physics} for  the foundational review  on how PiNNs function. This section briefly reviews the PiNN's core architecture and its state-of-the-art applications in computational fluid and solid mechanics and discusses some of the major limitations.  

\begin{figure}[htp]
    \centering
    \includegraphics[width=0.98\linewidth]{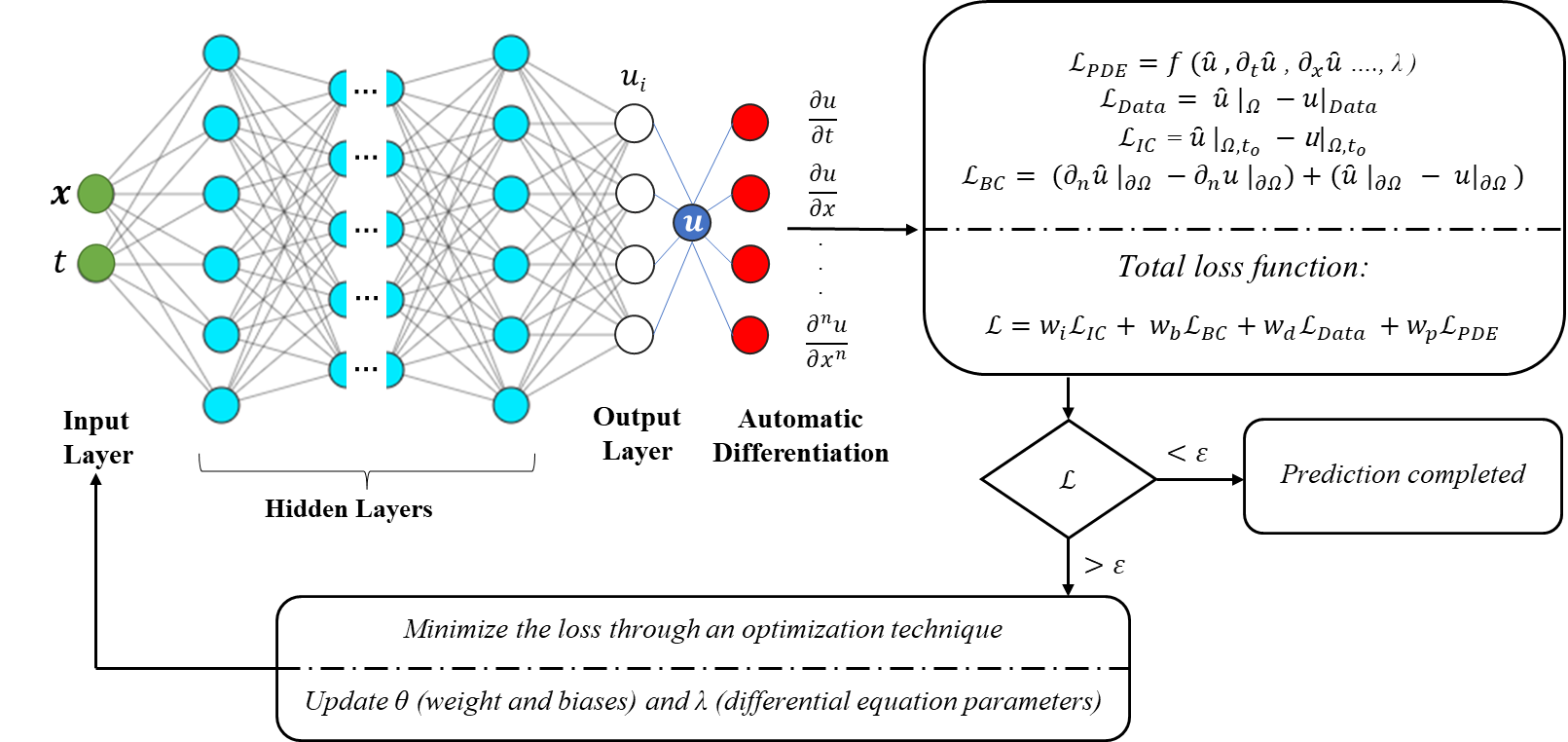}
    \caption{A schematic architecture of Physics-informed Neural Networks (PiNNs). The network digests  spatiotemporal coordinates (\textbf{\textit{x}},\textit{t}) as inputs to approximate the multi-physics solution $\hat{\textbf{\textit{u}}}$. The last layer generates the derivatives of the predicted solution $\textbf{\textit{u}}$ with respect to inputs, which are calculated using automatic differentiation (AD). These derivatives are used to formulate the residuals of the governing equations in the loss function, which is composed of multiple terms weighted by different coefficients. $\theta$ and $\lambda$ are the learnable parameters for weights/biases and unknown PDE parameters, respectively, that can be learned  simultaneously while minimizing the loss function.}
    \label{fig:PiNN}
\end{figure}

A schematic representation of a  vanilla PiNN architecture is illustrated in Fig.~\ref{fig:PiNN}. In PiNNs, the underlying physics is incorporated outside the neural network architecture to constrain the model while training, thereby ensuring outputs follow known physical laws. The most common method to emulate this process is through a weakly imposed penalty loss that penalizes the network for not following the physical constraints. As shown in Fig.~\ref{fig:PiNN}, a neural network with spatiotemporal features (i.e., $\textbf{x}$ and $t$) as input parameters and the PDE solution elements as output parameters (i.e., $\textbf{u}$) can be used to emulate any PDE. 

The network's outputs are then fed into the next layer, which is an automated differentiation layer. In this instance, multiple partial derivatives are generated by differentiating the outputs with regard to the input parameters ($\textbf{x}$ and $t$). With the goal of optimizing the PDE solution, these partial derivatives are used to generate the required terms in the loss function. 
The loss function in PiNN is a combination of the loss owing to labelled data ($\mathcal{L}_{Data}$), governing PDEs ($\mathcal{L}_{PDE}$), applied initial conditions ($\mathcal{L}_{IC}$) and applied boundary conditions ($\mathcal{L}_{BC}$) \cite {cai2022physics}. The $\mathcal{L}_{BC}$ ensures that the PiNN's solution meets the specified boundary constraints, whereas $\mathcal{L}_{Data}$ assures that the PiNN follows the trend in the training dataset (i.e., historical data, if any). Furthermore, the structure of the PDE is enforced in PiNN through the $\mathcal{L}_{PDE}$, which specifies the collocation points where the solution to the PDE holds \cite{raissi2017physics}. The weights for the loss due to the initial conditions, boundary conditions, data, and PDE can be specified as $w_i$, $w_b$, $w_d$, and $w_p$, respectively. The next step is to check, for a given iteration, if the loss is within the accepted tolerance, $\epsilon$. If not, the learnable parameters of the network ($\theta$) and unknown PDE parameters ($\lambda$) are updated through error backpropagation. For a given number of iterations, the entire cycle is repeated until the PiNN model produces learnable parameters with loss functions less than $\epsilon$. Note that the training of PiNNs is more complicated compared to PgNNs, as PiNNs are composed of sophisticated non-convex and multi-objective loss functions that may result in instability during optimization \cite{raissi2017physics,karniadakis2021physics,cai2022physics}. 
\color{black}

\citet{dissanayake1994neural} were the first to investigate the incorporation of prior knowledge into a neural network. Subsequently, \citet{owhadi2015bayesian} introduced the concept of physics-informed learning models as a result of the ever-increasing computing power, which enables the use of increasingly complex networks with more learnable parameters and layers. The PiNN, as a new computing paradigm for both forward and inverse modeling, was introduced by Raissi et al. in a series of papers \citep{raissi2017physics,raissi2018hidden,raissi2019physics}. \citet{raissi2017physics} deployed two PiNN models, a continuous and a discrete-time model, on examples consisting of different boundary conditions, critical non-linearities, and complex-valued solutions such as Burgers', Schrodinger's, and Allen-Cahn's equations. The results for Burgers' equation demonstrated that, given a sufficient number of collocation points (i.e., as the basis for the continuous model), an accurate and data-efficient learning procedure can be obtained \cite{raissi2017physics}. 

In continuous PiNN models, when dealing with higher-dimensional problems, the number of collocation points increases exponentially, making learning processing difficult and computationally expensive \cite{raissi2017physics,karniadakis2021physics}. \citet{raissi2017physics} presented a discrete time model based on the Runge-Kutta technique  \cite{iserles2009first} to address the computational cost issue. This model simply takes a spatial feature as input, and over time steps, PiNN converges to the underlying physics. For all the examples explored by \citet{raissi2017physics}, continuous and discrete PiNN models were able to satisfactorily build physics-informed surrogate models. \citet{nabian2021efficient} proposed an alternate method for managing collocation points. They investigated the effect of sampling collocation points according to distribution and discovered that it was proportional to the loss function. This concept requires no additional hyperparameters and is simpler to deploy in existing PiNN models. In their study, they claimed that a sampling approach for collocation points enhanced the PiNN model's behavior during training. The results were validated by deploying the hypothesis on PDEs for solving problems related to elasticity, diffusion, and plane stress physics. 

In order to use PiNN to handle inverse problems, the loss function of the deep neural network must satisfy both the measured and unknown values at a collection of collocation sites distributed throughout the problem domain. \citet{raissi2019physics}  showcased the potential of both continuous and discrete time PiNN models to solve benchmark inverse problems such as the propagation of non-linear shallow-water waves (Korteweg–De Vries equation) \cite{su1969korteweg} and incompressible fluid flows (Navier-Stokes equations) \cite{constantin2020navier}.

Compared to PgNNs, the PiNN models provide more accurate predictions for forward and inverse modeling, particularly in scenarios with high non-linearities, limited data, or noisy data \cite{stiasny2021physics}. As a result, it has been implemented in several fundamental scientific and applied fields. Aside from forward and inverse problems, the PiNN can also be used to develop partial differential equations for unknown phenomena if training data representing the phenomenon's underlying physics is available \cite{raissi2019physics}. \citet{raissi2019physics} leveraged both continuous time and discrete time PiNN models for generating universal PDEs depending on the type and structure of the available data. In the remainder of this section, we review the recent literature on PiNN's applications in the computational fluid and solid mechanics fields.   

\subsection{PiNNs for Fluid Mechanics}

\begin{figure}[htp]
    \centering
    \includegraphics[width=0.81\linewidth]{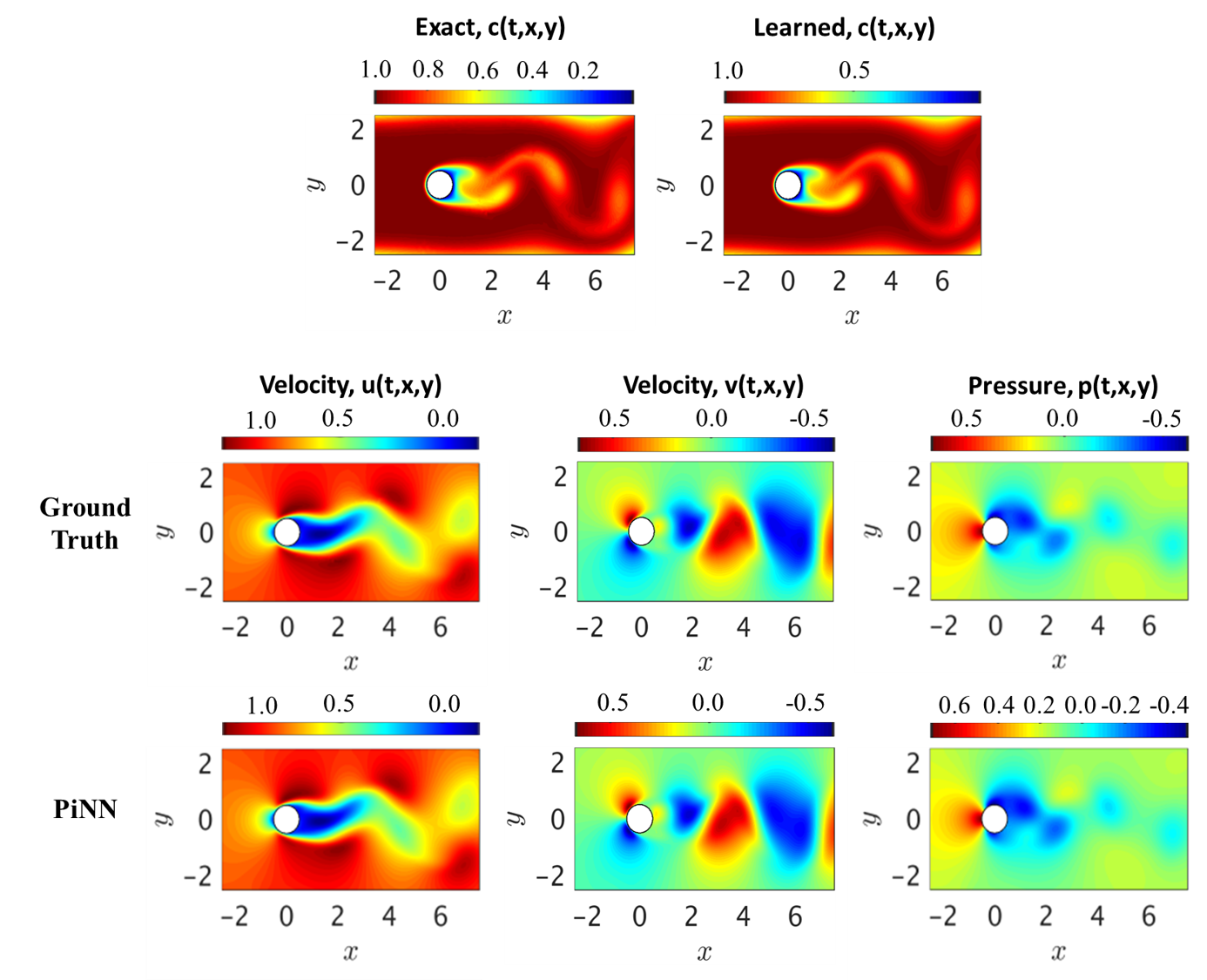}
    \caption{A comparison between ground truth simulation results and PiNN predictions for a 2D flow past a circular cylinder. Comparisons are shown for the concentration of passive scalar, $c(t,x,y)$, and resulting velocity fields, \textit{u,v}, and pressure field, \textit{p} (adapted from \citet{raissi2018hidden}).}
    \label{fig:Results 85}
\end{figure}

The application of PiNNs to problems involving fluid flow is an active, ongoing field of study \cite{mao2020physics, jagtap2022physics}. \citet{raissi2018hidden}, in a seminal work, developed a PiNN, so-called hidden fluid mechanics (HFM), to encode physical laws governing fluid motions, i.e., Navier-Stokes equations. They employed underlying conservation laws to derive hidden quantities of interest such as velocity and pressure fields from spatiotemporal visualizations of a passive scalar concentration, e.g., dye, transported in arbitrarily complex domains. Their algorithm to solve the data assimilation problem is agnostic to the boundary and initial conditions as well as to the geometry. Their model successfully predicted 2D and 3D pressure and velocity fields in benchmark problems inspired by real-world applications. Figure \ref{fig:Results 85}, adapted from \citet{raissi2018hidden}, compares the PiNN prediction with the ground truth for the classical problem of a 2D flow past a cylinder. The model can be used to extract valuable quantitative information such as wall shear stresses and lift and drag forces for which direct measurements are difficult to obtain. 

\citet{zhang2020frequency} also developed a PiNN framework for the incompressible fluid flow past a cylinder governed by Navier-Stokes equations. PiNN learns the relationship between simulation output (i.e., velocity and pressure) and the underlying geometry, boundary, initial conditions, and inherently fluid properties. They demonstrated that the generalization performance is enhanced across both the temporal domain and design space by including Fourier features \cite{tancik2020fourier}, such as frequency and phase offset parameters. \citet{cheng2021deep} developed Res-PiNN (i.e., Resnet blocks along with PiNN) for simulating cavity flow and flow past a cylinder governed by Burgers' and Navier-Stokes equations. Their results showed that Res-PiNN had better predictive ability than conventional PgNN and vanilla PiNN algorithms. \citet{lou2021physics} also demonstrated the potential of PiNN for solving inverse multiscale flow problems. They used PiNN for inverse modeling in both the continuum and rare-field regimes represented by the Boltzmann-Bhatnagar-Gross-Krook (BGK) collision model. The results showed that PiNN-BGK is a unified method (i.e., it can be used for forward and inverse modeling), easy to implement, and effective in solving ill-posed inverse problems \cite{lou2021physics}.

\citet{wessels2020neural} employed PiNN to develop an updated Lagrangian method for the solution of incompressible free surface flow subject to the inviscid Euler equations, the so-called Neural Particle Method (NPM). The method does not require any specific algorithmic treatment, which is usually necessary to accurately resolve the incompressibility constraint. In their work, it was demonstrated that NPM is able to accurately compute a pressure field that satisfies the incompressibility condition while avoiding topological constraints on the discretization process \cite{wessels2020neural}. In addition, PiNN has also been employed to model complex non-Newtonian fluid flows involving non-linear constitutive PDEs able to characterize the fluid's rheological behavior \cite{mahmoudabadbozchelou2022nn}.

\citet{haghighat2021physicsb} trained a PiNN model to solve the dimensionless form of the governing equations of coupled multiphase flow and deformation in porous media. \citet{almajid2022prediction} compared the predictions of PiNN with those of PgNN, i.e., a conventional artificial neural network, for solving the gas drainage problem of water-filled porous media. The study showed that PgNN performs well under certain conditions (i.e., when the observed data consists of early and late time saturation profiles), while the PiNN model performs robustly even when the observed data contains only an early time saturation profile (where extrapolations are needed). \citet{depina2022application} applied PiNN to model unsaturated groundwater flow problems governed by the Richards PDE and van Genuchten constitutive model \citep{tartakovsky2018learning}. They demonstrated that PiNNs can efficiently estimate the van Genuchten model parameters and solve the inverse problem with a relatively accurate approximation of the solution to the Richards equation.

Some of the other variants of PiNN models employed in fluid mechanics are: \textit{\textbf{nn-PiNN}}, where PiNN is employed  to solve constitutive models in conjunction with conservation of mass and momentum for non-Newtonian fluids \cite{mahmoudabadbozchelou2022nn}; \textit{\textbf{ViscoelasticNet}}, where PiNN is used for stress discovery and viscoelastic flow models selection \cite{thakur2022viscoelasticnet}, such as Oldroyd-B \cite{faroughi2020closure}, Giesekus and Linear PTT \cite{fernandes2022finite}; \textit{\textbf{RhINN}} which is a rheology-informed neural networks employed to solve constitutive equations for a Thixotropic-Elasto-Visco-Plastic complex fluid for a series of flow protocols \cite{mahmoudabadbozchelou2021rheology}; \textit{\textbf{CAN-PiNN}}, which is a coupled-automatic-numerical differential framework that combines the benefits of numerical differentiation (ND) and automatic differentiation (AD) for robust and efficient training of PiNN \cite{chiu2022can}; \textit{\textbf{ModalPiNN}}, which is a combination of  PiNN with enforced truncated Fourier decomposition \cite{van2004truncated} for periodic flow reconstruction \cite{raynaud2022modalpinn}; \textit{\textbf{GAPiNN}}, which is a geometry aware PiNN consisted  of variational auto encoder, PiNN and boundary constraining network for real-world applications with irregular geometries without parameterization \cite{oldenburg2022geometry}; \textit{\textbf{Spline-PiNN}}, which is a combination of PiNN and Hermite spline kernels based CNN employed to train a PiNN without any pre-computed training data and provide fast, continuous solutions that generalize to unseen domains \citep{wandel2022spline}; \textit{\textbf{cPiNN}}, which is a conservative physics-informed neural network  consisting of several PiNNs  communicating through the sub-domain interfaces flux continuity for solving conservation laws \citep{jagtap2020conservative};  \textit{\textbf{SA-PiNN}}, which is a self-adaptive PiNN to address the adaptive procedures needed to force  PiNN to fit accurately the stubborn spots in the solution of stiff PDEs \citep{mcclenny2020self}; and \textit{\textbf{XPiNN}}, which is an extended PiNN to enhance the representation and parallelization capacity of PiNN and generalization to any type of PDEs with respect to cPINN \citep{jagtap2021extended}. 

Table \ref{Fluid PiNN Table} reports a non-exhaustive list of recent studies that leveraged  PiNN  to model fluid flow problems. Furthermore, Table \ref{Fluid PiNN Enhanced Architecture Table} reports a non-exhaustive list of recent studies that developed other variants of PiNN architectures to improve the overall prediction accuracy and computational cost in fluid flow problems. 

\begin{table}[]
\centering
\caption {A non-exhaustive list of recent studies that leveraged PiNN to model fluid flow problems.}
\label{Fluid PiNN Table}
\begin{tabularx}{\textwidth}{>{\hsize=.25\hsize}X >{\hsize=.65\hsize}X >{\hsize=0.1\hsize}r}
\toprule  
 {Area of Application} & {Objectives}  &  {Reference}  \\ 
\midrule

 Incompressible  Flows & Accelerating the modeling of Navier-Stokes equations to infer the solution for various 2D and 3D flow problems & \cite{raissi2019physics} \\
\addlinespace
\addlinespace
 & Learning the relationship between output and underlying geometry as well as boundary conditions  &  \cite{zhang2020frequency} \\
  \addlinespace
\addlinespace
 & Simulating ill-posed (e.g., lacking boundary conditions)  or inverse laminar and turbulent  flow problems & \cite{jin2021nsfnets} \\
\addlinespace
\addlinespace
Turbulent Flows & Solving vortex-induced and wake-induced vibration of a cylinder at high Reynolds number   & \cite{cheng2021deep1} \\
 \addlinespace
\addlinespace
 &  Simulating turbulent incompressible flows without using any specific model or making turbulence assumptions & \cite{eivazi2022physics} \\
 \addlinespace
\addlinespace
  & Reconstructing Reynolds stress disparities described by Reynolds-averaged Navier-Stokes equations & \cite{wang2017physics} \\
 \addlinespace
\addlinespace
  Geofluid Flows & Solving well-based groundwater flow equations without utilizing labeled data & \cite{zhang2022gw} \\
\addlinespace
\addlinespace
   & Predicting high-fidelity multi-physics data from low-fidelity fluid flow and transport phenomena in porous media & \cite{aliakbari2022predicting}  \\
 \addlinespace
\addlinespace
    & Estimating Darcy's law-governed hydraulic conductivity for both saturated and unsaturated flows & \cite{tartakovsky2020physics} \\
\addlinespace
\addlinespace
    & Solving solute transport problems in homogeneous and heterogeneous porous media governed by the advection-dispersion equation & \cite{faroughi2022physics} \\
\addlinespace
\addlinespace
    & Predicting fluid flow in porous media by sparse observations and physics-informed PointNet & \cite{kashefi2022prediction} \\
\addlinespace
\addlinespace
Non-Newtonian Flows & Solving systems of coupled PDEs adopted for non-Newtonian fluid flow modeling  & \cite{mahmoudabadbozchelou2022nn} \\
 \addlinespace
\addlinespace
 & Simulating linear viscoelastic flow models such as Oldroyd-B, Giesekus, and Linear PTT  & \cite{thakur2022viscoelasticnet} \\
\addlinespace
\addlinespace
 & Simulating direct and inverse solutions of  rheological constitutive models for complex fluids & \cite{mahmoudabadbozchelou2021rheology} \\
\addlinespace 
\addlinespace
 Biomedical Flows & Enabling the seamless synthesis of non-invasive in-vivo measurement techniques and computational flow dynamics models derived from first physical principles & \cite{kissas2020machine} \\
\addlinespace
\addlinespace
   & Enhancing the quantification of near-wall blood flow and wall shear stress arterial in diseased arterial flows & \cite{arzani2021uncovering} \\
 \addlinespace
\addlinespace
 Supersonic Flows  & Solving inverse supersonic flow problems involving expansion and compression waves  & \cite{jagtap2022physics} \\
 \addlinespace
\addlinespace
 Surface Water Flows   & Solving ill-posed strongly non-linear and weakly-dispersive surface water waves governed by Serre-Green-Naghdi equations using only data of the free surface elevation and depth of the water. & \cite{jagtap2022deep} \\
\bottomrule
\end{tabularx}
\end{table}

\begin{table}[]
\centering
\caption {A non-exhaustive list of different variants of PiNN architectures used in modeling computational fluid flow problems.}
\label{Fluid PiNN Enhanced Architecture Table}
\begin{tabularx}{\textwidth}{ >{\hsize=.2\hsize}X >{\hsize=0.7\hsize}X>{\hsize=0.2\hsize}r}
\toprule  
  {PiNN Structure} &  {Objective }  &  {Reference}  \\ 
\midrule

 CAN-PiNN & Providing a  PiNN  with more accuracy and efficient training by integrating ND- and AD-based approaches  & \cite{chiu2022can}  \\

\addlinespace
\addlinespace
 ModalPiNN  & {Providing a  simpler representation of  PiNN  for oscillating phenomena to improve performance with respect to sparsity, noise and lack of synchronization in the  data }  & \cite{raynaud2022modalpinn} \\

\addlinespace
\addlinespace
GA-PiNN  & {Enhancing PiNN to develop a parameter-free, irregular geometry-based surrogate model for fluid flow modeling}  & \cite{oldenburg2022geometry} \\

\addlinespace
\addlinespace
 Spline-PiNN  & Improving the generalization of PiNN by combining it with Hermite splines CNN to solve the incompressible Navier-Stokes equations  & \cite{wandel2022spline} \\

\addlinespace
\addlinespace

 cPiNN  & Enhancing PiNN to solve  high dimensional non-linear conservation laws requiring high computational and memory requirements  & \cite{jagtap2020conservative} \\
\addlinespace
\addlinespace

SA-PiNN  & Improving the PiNN's convergence and accuracy problem for stiff PDEs using self-adaptive weights in the training & \cite{mcclenny2020self} \\
\addlinespace
\addlinespace

XPiNN  & Improving PiNN and cPiNN in terms of generalization, representation,  parallelization capacity, and computational cost& \cite{jagtap2021extended} \\
\addlinespace
\addlinespace
PiPN  &  overcoming the shortcoming of regular PiNNs that need to be retrained for any single domain with a new geometry & \cite{KASHEFI2022111510} \\

\bottomrule
\end{tabularx}
\end{table}

\color{black}

\subsection{PiNNs for Solid Mechanics}

The application of PiNNs in computational solid mechanics is also an active field of study. 
The study by \citet{haghighat2020deep} on modeling linear elasticity using PiNN  was among the first papers that introduced PiNN in the solid mechanics community. Since then, the framework has been extended to other solid-mechanics problems (e.g., linear and non-linear elastoplasticity, etc.).

\citet{shukla2021physics} used PiNN for surrogate modeling of the micro-structural properties of poly-crystalline nickel. In their study, in addition to employing the PiNN model, they applied an adaptive activation function to accelerate the convergence of numerical modeling. The resulting PiNN-based surrogate model demonstrated a viable strategy for non-destructive material evaluation. \citet{henkes2022physics} modeled non-linear stress and displacement fields induced by inhomogeneities in materials with sharp phase transitions using PiNN. To overcome the PiNN's convergence issues in this problem, they used adaptive training approaches and domain decomposition \cite{wessels2020neural}. According to their results, the domain decomposition approach is capable of properly resolving non-linear stress, displacement, and energy in heterogeneous microstructures derived from real-world  $\mu$CT-scans images \cite{henkes2022physics}. \citet{zhang2021physics} trained a PiNN model with a loss function based on the minimal energy criteria to investigate digital materials. The model tested  on 1D tension, 1D bending, and 2D tensile problems demonstrated equivalent performance when compared to supervised DL methods (i.e., PgNNs). By adding a hinge loss for the Jacobian matrix, the PiNN method was able to properly approximate the logarithmic strain and rectify any erroneous deformation gradient. 

\begin{figure}[]
    \centering
    \includegraphics[width=0.95\linewidth]{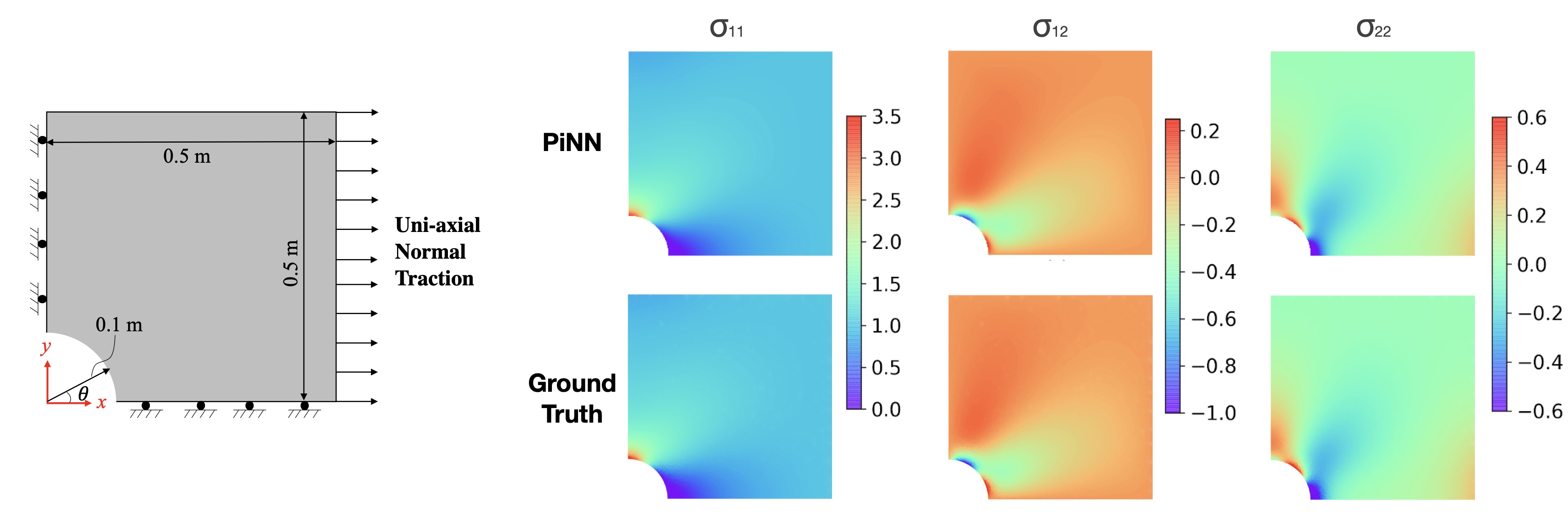}
    \caption{A comparison between mixed-variable PiNN's prediction and ground truth generated by FEM to predict the stress fields in a defected plate under uni-axial load (adapted from \citet{rao2020physics}). }
    \label{fig:Results 103}
\end{figure}

\citet{rao2020physics} proposed a PiNN architecture with mixed-variable (displacement and stress component) outputs to handle elastodynamic problems without labeled data. The method was found to boost the network's accuracy and trainability in contrast to the pure displacement-based PiNN model. Figure \ref{fig:Results 103} compares the ground truth stress fields generated by the FEM with the ones estimated by mixed-variable PiNN for an elastodynamic problem \cite{rao2020physics}. It can be observed that stress components can be accurately estimated by mixed-variable PiNN. \citet{rao2020physics} also proposed a composite scheme of PiNN to enforce the initial and boundary conditions in a hard manner as opposed to the conventional (vanilla) PiNN with soft initial and boundary condition enforcement. This model was tested on a series of dynamics problems (e.g., the defected plate under cyclic uni-axial tension and elastic wave propagation), and resulted in the mitigation of inaccuracies near the boundaries encountered by PiNN. 

\citet{fang2019deep} proposed a PiNN model to design the electromagnetic meta-materials used in  various practical applications such as cloaking, rotators, concentrators, etc. They studied PiNN's inference issues for Maxwell's equation \cite{lax1976maxwell} with a high wave number in the frequency domain and improved the activation function to overcome the high wave number problems. The proposed PiNN recovers not only the continuous functions but also piecewise functions, which is a new contribution to the application of PiNN in practical problems. \citet{zhang2020physics}  employed PiNN to identify nonhomogenous materials in elastic imaging for application in soft tissues. Two PiNNs were used, one for the approximate solution of the forward problem and another for approximating the field of the unknown material parameters. The results showed that the unknown distribution of mechanical properties can be accurately recovered using PiNN. \citet{abueidda2022enhanced} employed PiNN to simulate 3D hyperelasticity problems. They proposed an  Enhanced-PiNN architecture  consisting of  the residuals of the strong form and the potential energy \cite{abueidda2022deep}, producing several loss terms contributing to the definition of the total loss function to be minimized. The enhanced PiNN outperformed both the conventional (vanilla) PiNN and deep energy methods, especially when there were areas of high solution gradients.

\citet{haghighat2021physicsa} tested a different variant of PiNN to handle inverse problems and surrogate modeling in solid mechanics. Instead of employing a single neural network, they implemented a PiNN with multiple neural networks in their study. They deployed the framework on linear elastostatic and non-linear elastoplasticity problems and showed that the improved PiNN model provides a more reliable representation of the physical parameters. In addition, they investigated the domain of transfer learning in PiNN and found that the training phase converges more rapidly when transfer learning is used. \citet{yuan2022pinn} proposed an auxiliary PiNN model (dubbed as A-PiNN) to solve inverse problems of non-linear integro-differential equations (IDEs). A-PiNNs circumvent the limitation of integral discretization by establishing auxiliary output variables in the governing equation to represent the integral(s) and by substituting the integral operator with automated differentiation of the auxiliary output. Therefore, A-PiNN, with its multi-output neural network, is constructed such that it determines both primary and auxiliary outputs to  approximate both the variables and integrals in the governing equations. The A-PiNNs were used to address the inverse issue of non-linear IDEs, including the Volterra equation \citep{lu2021deepxde}. As demonstrated by their findings, the unknown parameters can be determined satisfactorily even with noisy data.

Some of the other variants of PiNN used in computational solid mechanics are:  \textit{\textbf{PhySRNet}}, which is a PiNN-based super-resolution framework for reconstructing high resolution output fields from low resolution counterparts without requiring high-resolution labelled data \cite{arora2022physrnet}; \textit{\textbf{PDDO-PiNN}}, which  is a combination of  peridynamic differential operator (PDDO) \cite{madenci2019peridynamic} and PiNN to overcome degrading performance of PiNN under sharp gradients \cite{haghighat2021nonlocal}; \textit{\textbf{PiELM}}, which is a combination of PiNN and extreme learning machine (ELM) \cite{huang2006extreme} employed to solve direct problems in linear elasticity \cite{yan2022framework}; \textit{\textbf{DPiNN}}, which is a distributed PiNN utilizing a piecewise-neural network representation for the underlying field, instead of the piece-polynomial representation commonly used in FEM \cite{yadav2022distributed}; and \textit{\textbf{PiNN-FEM}}, which is a mixed formulation based on PiNN and FE for computational mechanics in heterogeneous domain \cite{rezaei2022mixed}. 

\color{black}

Table \ref{Solid PiNN Table} reports a non-exhaustive list of recent studies that leveraged PiNN in computational solid mechanics. Furthermore, Table \ref{Solid PiNN Enhanced Architecture Table} reports a non-exhaustive list of recent studies that developed other variants of PiNN architectures to improve overall prediction accuracy and computational cost in solid mechanics modeling.

\begin{table}[]
\centering
\caption {A non-exhaustive list of recent studies that leveraged PiNN in computational solid mechanics.}
\label{Solid PiNN Table}
\begin{tabularx}{\textwidth}{ >{\hsize=.3\hsize}X >{\hsize=.7\hsize}X >{\hsize=0.1\hsize}c}
\toprule  
{Area of Application} & {Objectives }  &  {Reference}  \\  
\midrule
 Elasticity & Solving forward and inverse problems in linear elastostatic and non-linear elasticity problems  & \cite{haghighat2021physicsa} \\
\addlinespace
\addlinespace
  & Simulating forward and discovery problems for linear elasticity  & \cite{haghighat2020deep} \\
\addlinespace
\addlinespace 
 &  Resolving the non-homogeneous material identification problem in elasticity imaging & \cite{zhang2020physics} \\
\addlinespace
\addlinespace
  & Estimating elastic properties of tissues using pre- and post-compression images of  objects mimicking properties of tissues  & \cite{mallampati2021measuring} \\
\addlinespace
\addlinespace
   & Estimating mechanical response of elastic plates under different loading conditions  & \cite{li2021physicselastic} \\
\addlinespace
\addlinespace
   & Finding optimal solutions to reference biharmonic problems of elasticity and elastic plate theory  &
\cite{vahab2021physics} \\
\addlinespace
\addlinespace
   & Simulating elastodynamic problems, e.g., elastic wave propagation, deflected plate under periodic uniaxial strain, without labeled data   & \cite{rao2020physics} \\
\addlinespace
\addlinespace
Heterogeneous Materials  & Inferring the spatial variation of compliance coefficients of materials (e.g., speed of the elastic waves) to identify microstructure & \cite{shukla2021physics} \\
\addlinespace
\addlinespace
  &  Resolving non-linear stress, displacement, and energy fields in heterogeneous microstructures  & \cite{henkes2022physics} \\
 \addlinespace
\addlinespace
 & Solving coupled thermo-mechanics problems in composite materials & \cite{raj2021physics} \\

\addlinespace
\addlinespace
 & Predicting the size, shape, and location of the internal structures (e.g., void, inclusion) using  linear elasticity, hyperelasticity, and plasticity constitutive models  & \cite{zhang2022analyses} \\

 \addlinespace
\addlinespace
 Structural Elements  & Predicting the small-strain response of arbitrarily curved shells  & \cite{bastek2022physics} \\
 \addlinespace
\addlinespace
  & Solving  mechanical problems of elasticity in one-dimensional elements such as rods and beams   & \cite{katsikis2022gentle} \\
\addlinespace
\addlinespace
  & Predicting creep-fatigue life of 
  components (316 stainless steel) at elevated temperatures  & \cite{zhang2021physicsstain} \\
\addlinespace
\addlinespace
Structural Vibrations &  Estimating and optimizing  vibration characteristics and system properties of structural mechanics and vibration problems  & \cite{haghighat2021deepstructuralmech} \\
\addlinespace
\addlinespace
 Digital Materials  & Resolving physical behaviors of digital materials to design next-generation composites   & \cite{zhang2021physics} \\
\addlinespace
\addlinespace
Fracture Mechanics &  Reconstructing the solution of displacement field after damage to predict crack propagation for quasi-brittle materials  & \cite{zheng2022physics} \\
 \addlinespace
\addlinespace
 Elasto-viscoplasticity & Modeling the strain-rate and temperature dependence of the deformation fields (i.e., displacement, stress, plastic strain)  & \cite{arora2022physics} \\
\addlinespace
\addlinespace
 Additive Manufacturing & Predicting finite fatigue life in materials containing defects  & \cite{salvati2022defect} \\
 \addlinespace
\addlinespace
 Solid Mechanics & Providing a detailed introduction to programming PiNN-based computational solid mechanics from 1D to 3D problems  & \cite{bai2022introduction} \\

\bottomrule
\end{tabularx}
\end{table}

\begin{table}[]
\centering
\caption {A non-exhaustive list of different variants of PiNN architectures used in computational solid mechanics problems.}
\label{Solid PiNN Enhanced Architecture Table}
\begin{tabularx}{\textwidth}{ >{\hsize=0.2\hsize}X >{\hsize=0.7\hsize}X>{\hsize=0.2\hsize}r}
\toprule  
   {PiNN Architecture} &  {Objectives }  &  {Reference}  \\  
\midrule
\addlinespace
\addlinespace
PhySRNet & Enhancing PiNN using super resolution techniques to reconstruct downscaled fields  from their upscaled counterparts & \cite{arora2022physrnet} 
\\
 
\addlinespace
\addlinespace
 Enhanced PiNN &  Improving the interpolation and extrapolation ability of PiNN by integrating potential energy, collocation method, and deep learning for hyperelastic problems &  \cite{abueidda2022enhanced} \\
\addlinespace
\addlinespace
 PDDO-PiNN &  Improving the performance of PiNN in presence of sharp gradient by integrating PDDO and PiNN methods to solve elastoplastic deformation problems & \cite{haghighat2021nonlocal} \\
\addlinespace
\addlinespace
PiELM &  Accelerating PiNN's training process by integrating PiNN and ELM methods to model  high dimensional shell structures   & \cite{yan2022framework} \\
\addlinespace
\addlinespace
 PiELM &  Accelerating PiNN's training process by integrating PiNN and ELM methods to model  biharmonic equations  & \cite{dwivedi2020solution} \\
\addlinespace
\addlinespace
 DPiNN & Providing a truly unified framework for addressing problems in solid mechanics Solving high dimensional inverse in heterogeneous media such as linear elasticity   & \cite{yadav2022distributed} \\

\addlinespace
\addlinespace
A-PiNN  &   Improving PiNN model to solve inverse problems of non-linear integro-differential equations   & \cite{yuan2022pinn} \\

\addlinespace
\addlinespace
 PiNN-FEM &  Hybridizing FEM and PiNN to solve  heterogeneous media problems such as elasticity and poisson equation  & \cite{rezaei2022mixed} \\
 
 \bottomrule
    \end{tabularx}
\end{table}

\subsection{PiNNs Limitations}

PiNNs show a great potential to be used in modeling dynamical systems described by ODEs and/or PDEs, however, they come with several limitations and shortcomings that must be considered:  

\begin{itemize}
    \item  Vanilla PiNNs use deep networks consisting of a series of fully connected layers and a variant of gradient descent optimization. The learning process and hyperparameter tuning are conducted manually and are sample size- and problem-dependent. Their training, thus, may face gradient vanishing problems and can be prohibitively slow for practical three-dimensional problems \citep{dwivedi2021distributed}. In addition, vanilla PiNNs impose limitations on low-dimensional spatiotemporal parameterization due to the usage of fully connected layers \citep{rao2021hard}. 
    
    \item For linear, elliptic, and parabolic PDEs, \citet{shin2020convergence} provided the first convergence theory  with respect to the number of training data. They also discussed a set of conditions under which convergence can be guaranteed. However, there is no "solid" theoretical proof of convergence for PiNNs when applied to problems governed by non-linear PDEs. Note that deep learning models generally fail to realize theoretically established global minima; thus, this limitation is not specific to PiNNs and holds for all deep learning models. \citep{karniadakis2021physics}
    
    \item PiNNs contain several terms in the loss function with relative weighting that greatly affects the predicted solution. There are, currently, no guidelines for selecting weights optimally \cite{yadav2022distributed}. Different terms in the loss function may compete with each other during training, and this competition may reduce the stability of the training process. PiNNs also suffer during training when confronted by an ill-posed optimization problem due to their dependence on soft physical constraints \cite{rao2021hard}.
    
    \item  PiNNs suffer from low-frequency induced bias and frequently fail to solve non-linear PDEs for problems  governed by high-frequency or multiscale structures \cite{fuks2020limitations}. In fact, PiNNs may experience difficulty propagating information from initial conditions or boundary conditions to unseen parts of the domain or to future times, especially in large computational domains (e.g., unsteady turbulent flow) \cite{li2021physics}.
    
    \item PiNNs are solution learning algorithms, i.e., they learn the solutions to a given PDE for a single instance. For any given new instance of the functional parameters or  coefficients, PiNNs require training a new neural network \citep{faroughi2022physics}. This is because, by construction, PiNNs cannot learn the physical operation of a given phenomenon, and that limits their generalization (e.g., spatiotemporal extrapolation). The PiNN approach, thus, suffers from the same computational issue as classical solvers, especially in 3D problems (e.g., FEM, FVM, etc.), as the optimization problem needs to be solved for every new instance of PDE parameters, boundary conditions, and initial conditions \cite{goswami2022physicsReview}.  
    
   \item PiNNs encounter difficulties while learning the solutions to inverse problems in heterogeneous media, e.g., a composite slab composed of several materials \citep{dwivedi2021distributed}. In such cases, the parameters of the underlying PDE (e.g., conductivity or permeability coefficients) change across the domain, yet a PiNN outputs unique parameter values over the whole domain due to its inherent design.

\end{itemize}

Despite the shortcomings, PiNNs offer a strong promise for complex domains that are hard to mesh and practical problems where data acquisition is expensive. To circumvent some of the limitations of vanilla PiNN, several techniques have been proposed. For instance, to address the first limitation listed above, discrete learning techniques using convolutional filters, such as HybridNet \cite{long2018hybridnet}, dense convolutional encoder-decoder network \cite{zhu2019physics}, auto-regressive encoder-decoder model \cite{geneva2020modeling}, TF-Net \cite{wang2020towards}, DiscretizationNet \cite{ranade2021discretizationnet}, and PhyGeoNet \cite{gao2021phygeonet}, just to name a few, have been employed that exceed vanilla PiNN in terms of computational efficiency. As another example, to address the last limitation listed above, \citet{dwivedi2021distributed} proposed a Distributed PiNN (DPiNN) that has potential advantages over existing PiNNs to solve the inverse problems in heterogeneous media, which are most likely to be encountered in engineering practices. Some of the other solutions to solve high dimensional inverse problems are Conservative PiNN (cPiNN) \cite{jagtap2020conservative} and Self-Adaptive PiNN  \cite{mcclenny2020self}. Further, XPiNN \cite{jagtap2021extended}, with its intrinsic parallelization capabilities to deploy multiple neural networks in smaller subdomains, can be used to considerably reduce the computational cost of PiNNs in large (three-dimensional) domains. However, these modifications and alternatives do not solve the generalization problem of  PiNNs as the resultant models lack the ability to enforce the existing physical knowledge. To this end, physics-encoded neural networks (PeNNs) have started to emerge. In the next section, we will review the recent literature on physics-encoded neural networks.

\section{Physics-encoded Neural Networks,  PeNNs}

Physics-encoded Neural Networks (PeNNs) are another family of mesh-free algorithms used in scientific computing, mostly in the fluid mechanics and solid mechanics fields, that strive to \textit{hard-encode underlying physics} (i.e., prior knowledge) into the core architecture of the neural networks. Note that, by construction, PeNN-based models extend the learning capability of a neural network from instance learning (imposed by PgNN and PiNN architectures) to continuous learning  \cite{chen2018neural,rao2021hard,rao2022discovering}. To hard-encode physical laws (in terms of ODEs, PDEs, closure laws, etc.) into a neural network, different approaches have been recently proposed \cite{rao2021hard,chen2018neural,li2020fourier,innes2019differentiable}. PeNN is not a completely new notion, as there has been a long trajectory of research that has proposed the philosophy of building physics constraints constructively into the architectures. For example, one can refer to preserving convexity  \cite{wang1994deterministic} using Deterministic Annealing Neural Network (DANN), preserving positivity  \cite{rangarajan1996novel}, enforcing symmetries in physics using  Lagrangian neural networks (LaNN) \cite{cranmer2020lagrangian,allen2020lagnetvip}, capturing trajectories using symplectic recurrent neural networks (SRNNs) \cite{chen2019symplectic,dipietro2020sparse}, enforcing exact physics and extracting  structure-preserving surrogate models using data-driven exterior calculus (DDEC) on graphs \cite{trask2022enforcing}, etc. In this section, we review the most two prominent approaches for encoding physics in neural network architectures and their applications in computational fluid and solid mechanics: (i) Physics-encoded Recurrent Convolutional Neural Network (PeRCNN) \cite{rao2021hard,rao2022discovering}, and (ii) Differential Programming (DP) or Neural Ordinary Differential Equations (NeuralODE) \cite{chen2018neural,innes2019differentiable}.

\subsection{Physics-encoded Recurrent Convolutional Neural Network (PeRCNN)}
\begin{figure}[]
    \centering
    \includegraphics[width=0.98\linewidth]{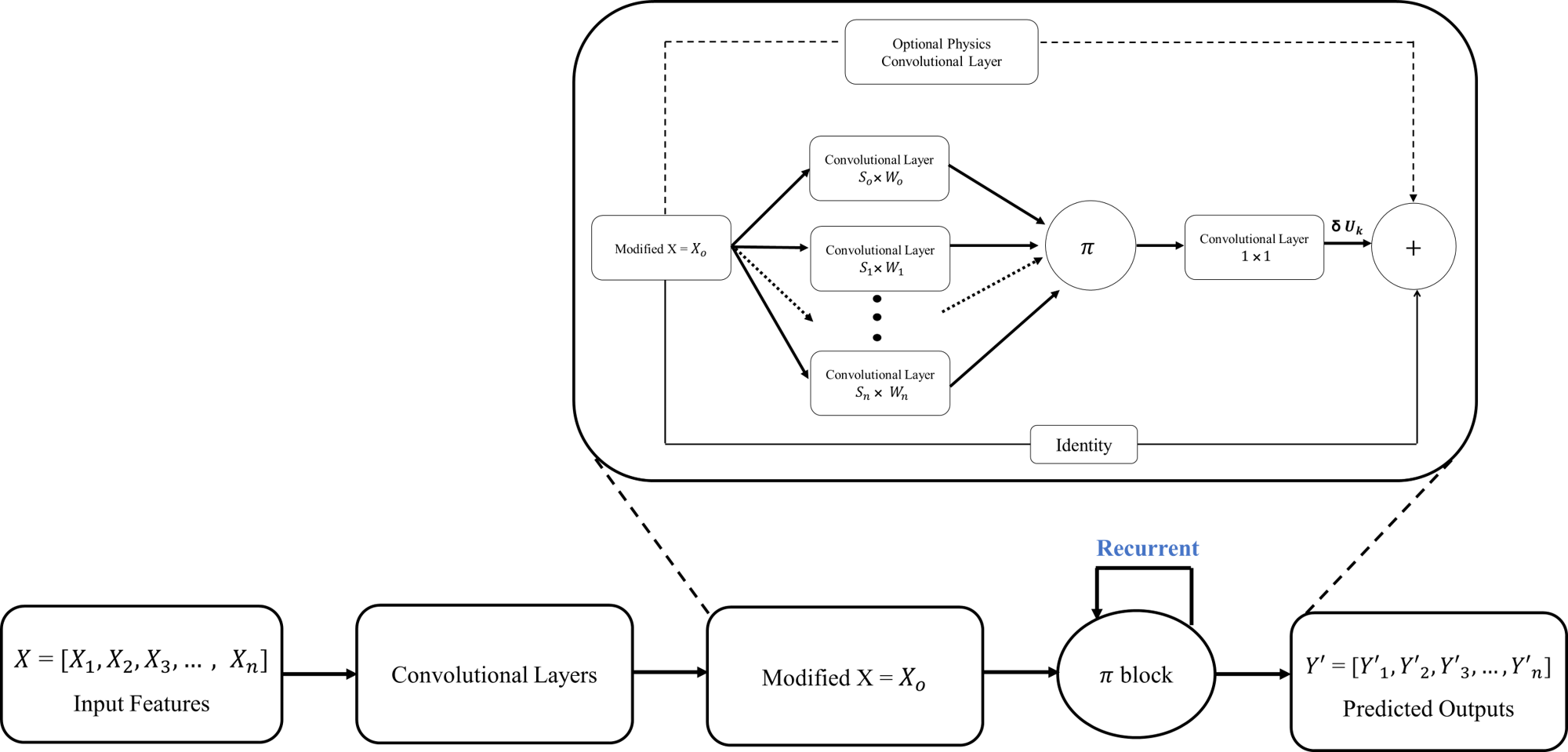}
    \caption {A schematic architecture for Physics-encoded Recurrent Convolutional Neural Network \cite{rao2021hard}. The architecture consists of $\textbf{X}$ as initial inputs, convolutional layers, $X_o$ as full resolution initial state, unconventional convolutional block ($\pi$ block), and the predicted output layer $\textbf{Y'}$. Further, the $\pi$ block consists of multiple parallel convolutional layers whose operations are defined as $S_n \cross W_n$, where n is the number of layers; $\pi$ carries out element-wise product and $+$ carries out element-wise addition.}
    \label{fig:HardEncoded}
\end{figure}

\citet{rao2021hard} introduced the PeRCNN model, which hard encodes prior knowledge governing non-linear systems into a neural network. The PeRCNN architecture shown in Fig.~\ref{fig:HardEncoded} facilitates learning in a data-driven manner while forcibly encoding the known physics knowledge. This model exceeds PgNN and PiNN's capabilities for phenomena in which the explicit formulation of PDEs does not exist and very limited measurement data is available (e.g., Earth or climate system modeling \cite{bauer2021digital}). The proposed encoding mechanism of physics, which is fundamentally different from the penalty-based physics-informed learning, ensures the network rigorously obeys the given physics. Instead of using non-linear activation functions, they proposed a novel element-wise product operation to achieve the non-linearity of the model. Numerical experiments demonstrated that the resulting physics-encoded learning paradigm possesses remarkable robustness against data noise/scarcity and generalizability compared with some state-of-the-art models for data-driven modeling.

As shown in Fig.~\ref{fig:HardEncoded},  PeRCNN is made of: an input layer, which is constituted by low-resolution noisy initial state measurements $X = [X_1, X_2, X_3, ..., X_n]$; a fully convolutional network, as the initial state generator (ISG), which downscales/upsamples the low resolution initial state to a full resolution initial state, dubbed as modified $X_o$, to be used as input to further recurrent computations. For the purpose of recurrent computing, an unconventional convolutional block, dubbed as $\pi$, is employed \cite{rao2021hard}. In the $\pi$ block, which is the core of PeRCNN, the modified $X_o$ goes through multiple parallel convolutional layers, whose feature maps will then be fused via an element-wise product layer. Further, a one-by-one ($1\cross1$) convolutional layer \cite{lin2013network}, is appended after the product operation to aggregate multiple channels into the output of the desired number of channels. Assuming the output of the $1\cross1$ convolution layer approximates the non-linear function, it can be multiplied  by the time spacing $\delta t$ to obtain the residual of the dynamical system at time $t_k$, i.e., $\delta U_k$. Ultimately, the last  layer generates predictions $Y' = [Y'_1, Y'_2, Y'_3,...,Y'_n]$ by element-wise addition. These operations are shown schematically in Fig.~\ref{fig:HardEncoded}.  

\begin{figure}[]
    \centering
    \includegraphics[width=0.999\linewidth]{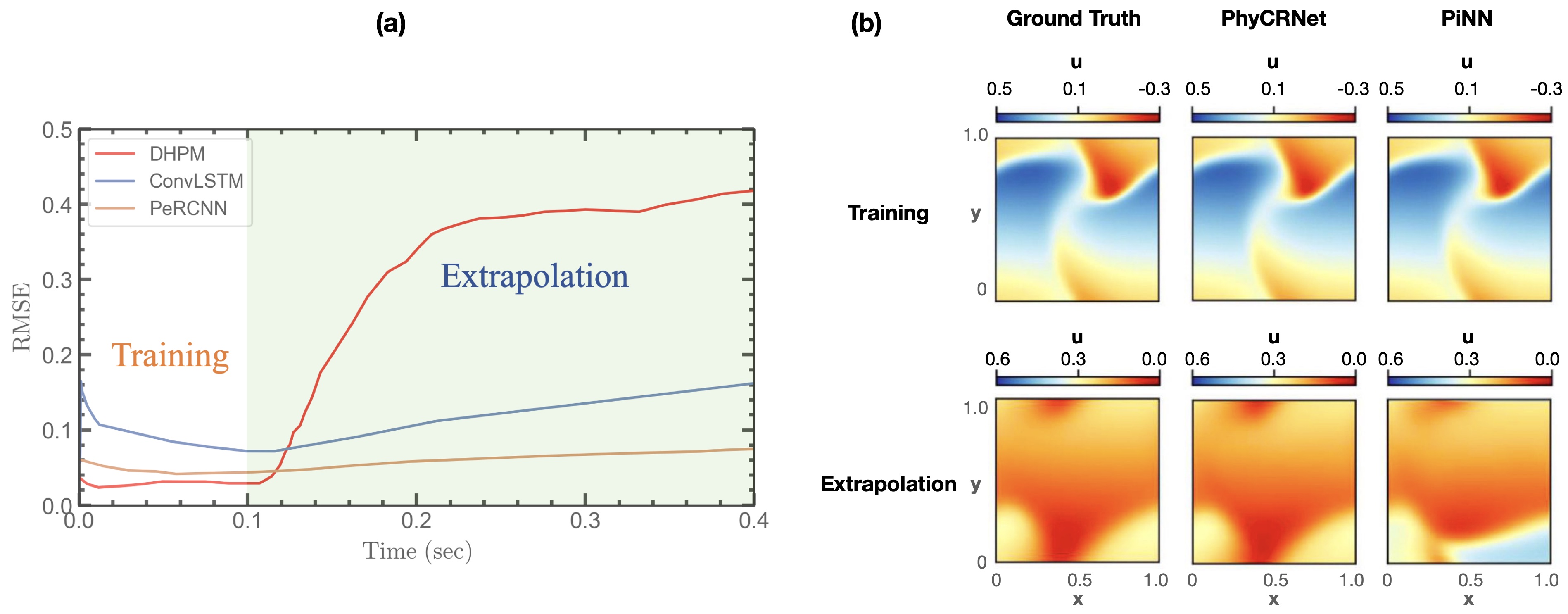}
    \caption {Panel (a) shows a comparison of error propagation in the training and extrapolation phases for a physics encoded recurrent convolutional neural network (PeRCNN), deep hidden physics model (DHPM), and a convolutional LSTM (ConvLSTM) modeling 2D Burgers' dataset (adapted from \citet{rao2021hard}). Panel (b) shows a comparison between the predictions by PhyCRNet and PiNN for 2D Burgers’ equations. The predicted velocity field in x direction is compared at the training time of $t = 1~s$, and at the extrapolation time of $t = 3~s$ (adapted from \citet{ren2022phycrnet}).}
    \label{fig:PhyCRNet}
\end{figure}

The PeRCNN architecture was tested on two datasets representing 2D Burgers and 3D Gray-Scott reaction-diffusion equations  \cite{rao2021hard}. In both cases, PeRCNN was compared with Convolutional LSTM \cite{shi2015convolutional}, Deep Residual Network \cite{he2016deep}, and Deep Hidden Physics Models \citep{raissi2018deephidden} in terms of accuracy (root-mean-squared-error, RMSE), data noise/scarcity, and generalization. The comparison for the 2D Burgers' dataset is shown in Fig.~\ref{fig:PhyCRNet}(a), adapted from \cite{rao2021hard}. The accumulative RMSE for PeRCNN began with a larger value in the training region (due to 10 percent Gaussian noise in the data) and reduced as additional time steps were assessed. The accumulative RMSE for PeRCNN slightly increases in the extrapolation phase (as a measure of the model's generalization), but clearly surpasses all other algorithms in terms of long-term extrapolation. \citet{rao2022discovering} also used PeRCNN for discovering spatiotemporal PDEs from scarce and noisy data and demonstrated its effectiveness and superiority compared to baseline models. 

\citet{ren2022phycrnet} proposed a hybrid algorithm combining PeRCNN and PiNN to solve the limitations in low-dimensional spatiotemporal parameterization encountered by PgNNs and PiNNs. In the resultant physics-informed convolutional-recurrent network, dubbed as PhyCRNet, an encoder-decoder convolutional  LSTM network is proposed for low-dimensional spatial feature extraction and temporal evolution learning. In PhyCRNet, the loss function is specified as aggregated discretized PDE residuals. The boundary conditions are hard-coded in the network via designated padding, and initial conditions are defined as the first input state variable for the network. Autoregressive and residual connections that explicitly simulate time marching were used to enhance the networks. This method ensures generalization to a variety of initial and boundary condition scenarios and yields a well-posed optimization problem in network training. Using PhyCRNet, it is also possible to simultaneously enforce known conservation laws into the network (e.g., mass conservation can be enforced by applying a stream function as the solution variable in the network for fluid dynamics) \cite{ren2022phycrnet}. \citet{ren2022phycrnet} evaluated and validated the performance of PhyCRNet using several non-linear PDEs compared to state-of-the-art baseline algorithms such as the PiNN and auto-regressive dense encoder-decoder model \cite{geneva2020modeling}. A comparison between PhyCRNet and PiNN to solve Burgers' equation is shown in Fig.~\ref{fig:PhyCRNet}(b) \cite{ren2022phycrnet}. Results obtained by \citet{ren2022phycrnet} clearly demonstrated the superiority of the PhyCRNet methodology in terms of solution accuracy, extrapolability, and generalizability. 

 \subsection{Neural Ordinary Differential Equations (NeuralODE)}

The neural ordinary differential equations (NeuralODE) method is another family of PeNN models in which the hidden state of the neural network is transformed from a discrete sequence to a continuous non-linear function by parametrizing the hidden state derivative using a differentiable function \citep{chen2018neural}. The output of the network is then computed using a traditional differential equation solver. During training, the error is back-propagated through the network as well as through the ODE solver without access to its internal operations. This architecture is feasible due to the fact that numerical linear algebra is the common underlying infrastructure for both scientific computing and deep learning, which is bridged by automated differentiation (AD) \citep{baydin2018automatic}. Because differential equations and neural networks are both differentiable, standard optimization and error backpropagation techniques can be used to optimize the network's weights during training. Instead of learning the non-linear transformation directly from the training data, the model in NeuralODE learns the structures of the non-linear transformation. Therefore, due to the fact that the neural network optimization equations are differentiable, the physical differential equations can be encoded directly into a layer as opposed to adding more layers (e.g., deeper networks). This results in a shallower network mimicking an infinitely deep model that can be inferred continuously at any desired accuracy at reduced memory and computational cost 
\citep{rackauckas2019diffeqflux}).

These continuous-depth models offer features that are lacking in PiNN and PgNNs, such as (i) a reduced number of parameters for supervised learning, (ii) constant memory cost as a function of depth, and (iii) continuous time-series learning (i.e., training with datasets acquired at arbitrary time intervals), just to name a few \citep{chen2018neural}. However, the error backpropagation may cause technical difficulties while training such continuous-depth networks. \citet{chen2018neural} computed gradients using the adjoint sensitivity method \cite{pontryagin1987mathematical} while considering the ODE solver as a black box. They demonstrated that this method uses minimal memory, can directly control numerical error, and, most importantly, scales linearly with the problem size.

\citet{ma2021comparison} compared the performance of discrete and continuous adjoint sensitivity analysis. They indicated that forward-mode discrete local sensitivity analysis implemented via AD is more efficient than reverse-mode and continuous forward and/or adjoint sensitivity analysis for problems with approximately fewer than 100 parameters. However, in terms of scalability, they showed that the continuous adjoint method is more efficient than the discrete adjoint and forward methods.

\begin{figure}[htp]
    \centering
    \includegraphics[width=0.99\linewidth]{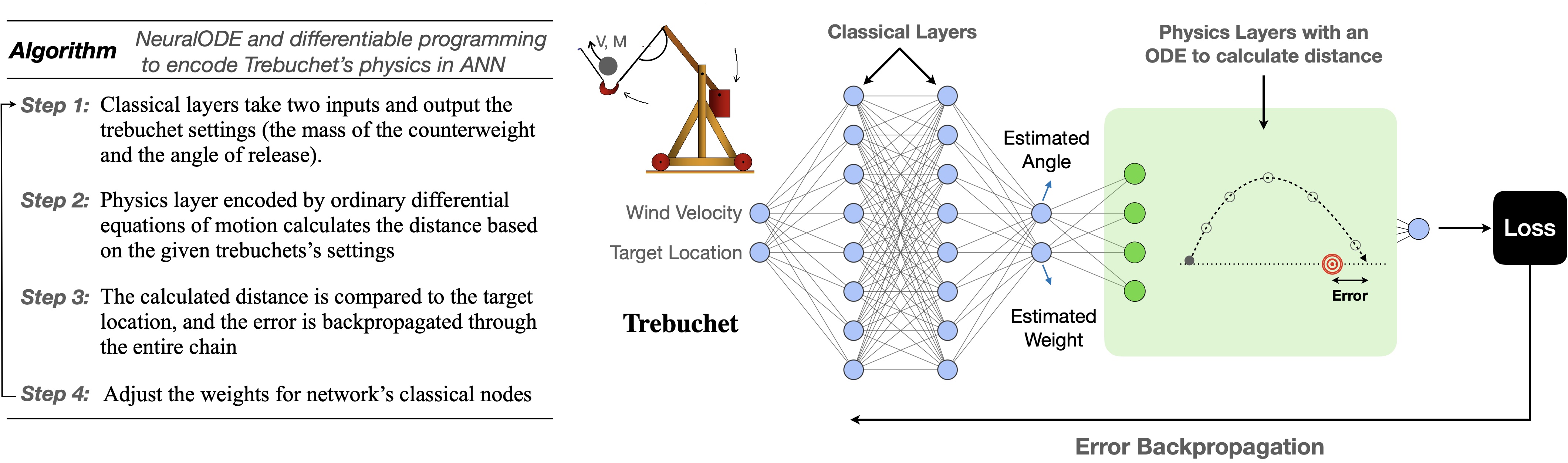}
    \caption {A NeuralODE architecture that leverages differentiable programming to model the inverse dynamics of a trebuchet. This simple network is 100x  faster than direct optimization (adapted from \citet{innes2019differentiable}).}
    \label{fig:nODE}
\end{figure}

Several computational libraries have been implemented to facilitate the practical application of NeuralODE. \citet{poli2020torchdyn} implemented the TorchDyn  library to train NeuralODE models and be as accessible as regular plug-and-play deep learning primitives. \citet{innes2019differentiable} and \citet{rackauckas2019diffeqflux} developed GPU-accelerated Zygote and DiffEqFlux libraries in the Julia coding ecosystem to bring differentiable programming and universal differential equation solver capabilities together. As an example, they encoded the ordinary differential equation of motion, as the transformation function, into a neural network to simulate the trebuchet's inverse dynamics \citep{innes2019differentiable}. As shown in Fig.~\ref{fig:nODE}, the network with classical layers takes the target location and wind speed as input and estimates the weight and angle of the projectile to hit the target. These outputs are fed into the ODE solver to calculate the achieved distance. The model compares the predicted value with the target location and backpropagates the error through the entire chain to adjust the weights of the network. This PeNN model solves the trebuchet's inverse dynamics on a personal computer 100x  faster than a classical optimization algorithm for this inverse problem. Once trained, this network can be used to aim at any blind target, not just the ones it was trained on; hence, the model is both accelerated and predictive.

NeuralODE has also been integrated with PiNN models, dubbed as PiNODE, in order to further constrain the network with known governing physics during training. Such an architecture consists of a neural network whose hidden state is parameterized by an ODE with a loss function similar to the PiNN's loss function (see Fig.~\ref{fig:PiNN}). The loss function penalizes the algorithm based on data and the strong form of the governing ODEs and backpropagates the error through the application of the adjoint sensitivity methods \citep{ma2021comparison}, to update the learnable parameters in the architecture. PiNODE can be deployed to overcome high bias (due to the use of first principles in scientific modeling) and high variance (due to the use of pure data-driven models in scientific modeling) problems. In other words, using PiNODE, prior physics knowledge in terms of ODE is integrated where it is available, and function approximation (e.g., neural networks) is used where it is not available. \citet{lai2021structural} used PiNODE to model the governing equations in the field of structural dynamics (e.g., free vibration of a 4-degree-of-freedom dynamical system with cubic non-linearity). They showed that PiNODE provides an adaptable framework for structural health monitoring  (e.g., damage detection) problems. \citet{roehrl2020modeling} tested PiNODE  using a forward model of an inverted pendulum on a cart and showed the approach can learn the non-conservative forces in a real-world physical system with substantial uncertainty.

The application of neural differential equation has also been extended to learn the dynamics of PDE-described systems. \citet{dulny2021neuralpde} proposed NeuralPDE by combining the Method of Lines (which represents arbitrarily complex PDEs by a system of ODEs) and NeuralODE through the use of a multi-layer convolutional neural network. They tested NeuralPDE on several spatiotemporal datasets generated from the advection-diffusion equation, Burgers' equation, wave propagation equation, climate modeling, etc. They found that NeuralPDE is competitive with other DL-based approaches, e.g., ResNet \cite{he2016identity}. The NeuralPDE's limitations are set by the limitations of the Method of Lines, e.g., it cannot be used to solve elliptical second-order PDEs. Table \ref{PeNN Table} reports a non-exhaustive  list of leading studies that leveraged PeNN to model different scientific problems. 

\begin{table}[H]
\centering
\caption {A non-exhaustive  list of recent studies that leveraged PeNN to model different scientific problems.}
\label{PeNN Table}
\begin{tabularx}{\textwidth}{ >{\hsize=.2\hsize}X >{\hsize=0.7\hsize}X>{\hsize=0.2\hsize}r}
\toprule  
  {PeNN Structure} &  {Objective }  &  {Reference}  \\ 
\midrule
PeRCNN  & Hard-encoding prior knowledge into a neural network to model non-linear systems  & \cite{rao2021hard} \\
\addlinespace
\addlinespace
PhyCRNet &  Leveraging the benefits of PeRCNN and PiNN into a single architecture  & \cite{ren2022phycrnet} \\
 \addlinespace
\addlinespace
 NeuralODE & Developing  a continuous-depth network by parametrizing the hidden state derivative using a differentiable function  & \cite{chen2018neural}  \\
\addlinespace
\addlinespace
 Zygote / DiffEqFlux  &  Providing libraries in the Julia coding ecosystem to facilitate the practical application of NeuralODE.  & \cite{innes2019differentiable,rackauckas2019diffeqflux}  \\
\addlinespace
\addlinespace
PiNODE  & Integrating PiNN and NeuralODE to provide an adaptable framework for structural health monitoring   & \cite{lai2021structural}  \\
\addlinespace
\addlinespace
NeuralPDE  & Combining the Method of Lines  and NeuralODE to learn the dynamics of PDE-described systems & \cite{dulny2021neuralpde}  \\
\addlinespace
\addlinespace
LaNN &  Capturing symmetries in physical problems such as relativistic particle and double pendulum   &  \cite{cranmer2020lagrangian} \\
\addlinespace
\addlinespace
SRNNs &  Capturing dynamics of hamiltonian systems such as three-body and spring-chain system from observed trajectories   &  \cite{chen2019symplectic} \\
\addlinespace
\addlinespace
DDEC &  Enforcing exact physics and extracting structure-preserving surrogate models  &  \cite{trask2022enforcing} \\

\bottomrule
    \end{tabularx}
\end{table}
 
\setlength{\arrayrulewidth}{1.5mm}
\setlength{\tabcolsep}{16pt}

\subsection{PeNNs Limitations}

Despite the advancement of numerous PeNN models and their success in modeling complex physical systems, these new architectures also face several challenges. The most important one is attributed to the training. PeNN-based models promote continuous learning using the development of continuous-depth networks, which makes PeNNs more difficult to train than PgNNs and PiNNs. Considering this, most of the limitations faced by PgNN and PiNN (e.g., convergence rate, stability, scalability, sample size- and problem-dependency) are also faced by PeNN. In addition, PeNNs usually have complex architectures, and their implementation is not as straightforward as PiNNs or PgNNs. In spite of PeNNs' implementation complexity, their efficient algorithms in the finite-dimensional setting, their ability to provide transferable solutions, their robustness against data scarcity, and their generalizability compared to PgNN and PiNN make them have a great potential to significantly accelerate traditional scientific computing for applications in computational fluid and solid mechanics.

\section{Neural Operators, NOs}

Most of the scientific deep learning methods discussed so far, e.g., PgNNs, PiNNs, and PeNNs, are generally designed to map  the solution of a physical phenomenon for a single instance (e.g., a certain spatiotemporal domain and boundary conditions to solve a PDE using PiNN), and thus, must re-trained or further trained (e.g., transfer learning \cite{goswami2020transfer}) to map  the solution under a different instant. Another way to alleviate this problem is to use neural operators that learn nonlinear mappings between function spaces \citep{lu2019deeponet,bhattacharya2020model,li2020neuralb}. Neural operators, thus, form another simulation paradigm that learns the underlying linear and nonlinear continuous operators using advanced architecture. These models, similar to PgNNs, enforce the physics of the problem  using  labelled input-output dataset pairs but provide enhanced generalization, interpretability, continuous learning, and computational efficiency compared to PgNNs as well as PiNNs and PeNNs \cite{li2020fourier,chen2018neural, li2021physics}.   

This new paradigm uses mesh-invariant, infinite-dimensional operators based on neural networks that do not require a prior understanding of PDEs. Neural operators merely work with data to learn the resolution-invariant solution to the problem of interest \citep{li2021physics}. In other words, neural operators can be trained on one spatiotemporal resolution and successfully inferred on any other \cite{li2020neuralb}. This resolution-invariant feature is achieved using the fact that a neural operator learns continuous functions rather than discretized vectors, by parameterizing the model in function spaces \cite{li2021physics,li2020neuralb}. Note that PgNNs and PiNNs, for example using MLP, may also guarantee a small generalization error, but that is only achieved by sufficiently large networks. One distinct feature of neural operators is their robustness for applications requiring real-time inference \cite{goswami2022physicsReview}. Three main neural operators have been proposed recently, namely (i) deep operator networks (DeepONets) \cite{lu2021learning},  (ii) Fourier neural operator (FNO) \cite{li2020fourier}, and (iii) graph neural operator (GNO) \cite{li2020neuralb,migus2022multi}. A recent review by \citet{goswami2022physicsReview} extensively compared these  neural operators. In this section, we briefly review DeepONets and FNO as the two prominent neural operators to be applied in computational fluid and solid mechanics. 

\subsection{Deep Operator Networks (DeepONets)}

\begin{figure}[htp]
    \centering
    \includegraphics[width=0.98\linewidth]{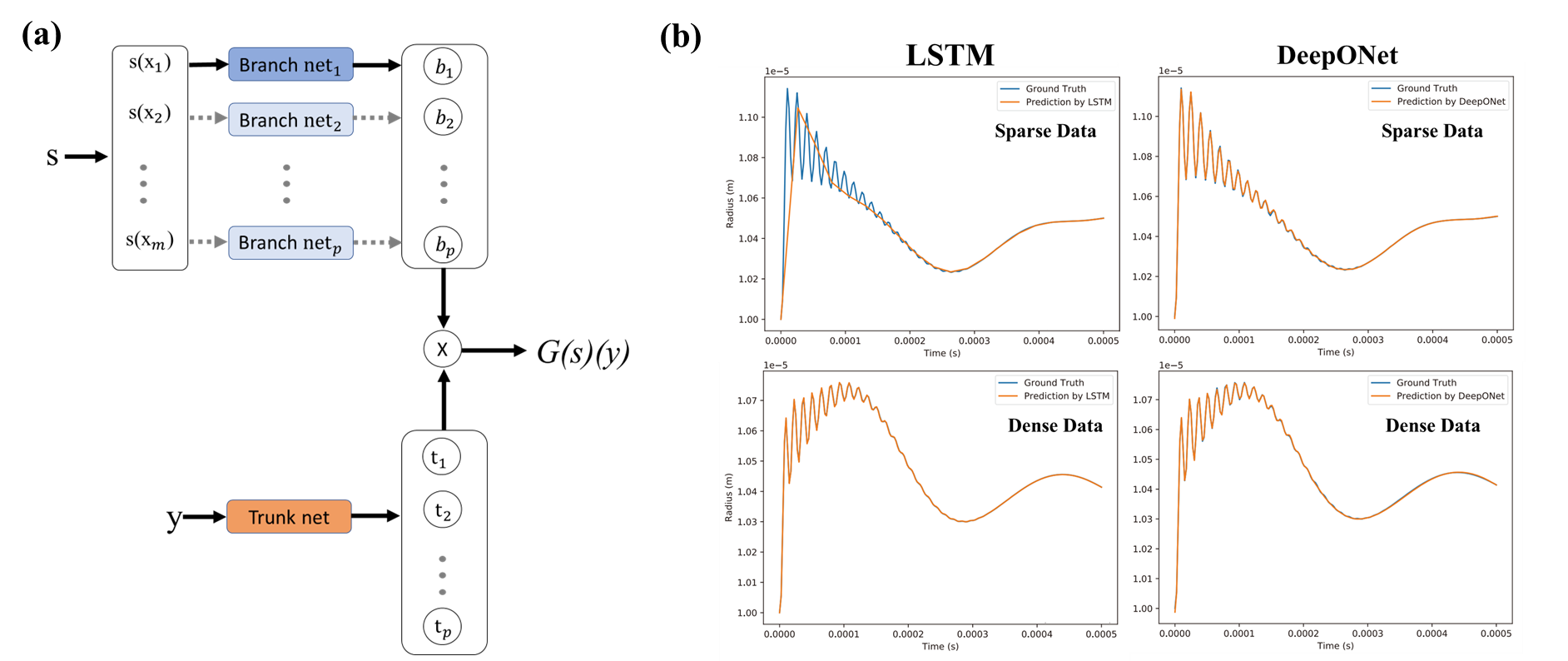}
    \caption {DeepONet for computational mechanics. Panel(a) depicts the general architecture of stacked DeepONets to learn an operator $G$:$s \rightarrow G(s)$.  The stacked DeepONet reduces to an unstacked DeepONet when only one branch network (e.g., the one shown in dark blue) is used. Alternatively, the unstacked DeepONet may be considered as a stacked DeepONet, in which all of the branch networks share the same set of parameters \cite{lu2021learning}. Panel(b) shows a comparison between DeepONet and LSTM (i.e., PgNN) to model sparse and dense datasets representing the formation of a single bubble in response to time-varying changes in the ambient liquid pressure (adapted from \citet{lin2021operator}). }
    \label{fig:CombinedDeepONet}
\end{figure}

\citet{lu2019deeponet} developed deep operator networks (DeepONets) based on the universal approximation theorem for operators \cite{chen1995universal} that can be used  to learn operators accurately and efficiently with very small generalization errors. \citet{lu2021learning} proposed two  architectures known as stacked and unstacked  for  DeepONet. The stacked DeepONet architecture is shown in Fig.~\ref{fig:CombinedDeepONet}(a), which consists of one trunk network and multiple stacked branch networks, $k = 1, 2..., p$. The stacked DeepONet is formed by selecting the trunk network as a one-layer network with width \textit{p} and each branch network as a one-hidden-layer network with width \textit{n}. To learn an operator $G$:$s \rightarrow G(s)$, the stacked DeepONet architecture takes  function $s$ as the input to branch networks and \textit{y} (i.e., points in the domain of $G(s)$) as the input to the trunk network. Here, the vector $[(x_1), (x_2),..., (x_m)]$ represents the finite locations of data, alternatively dubbed as sensors. The trunk network outputs $[t_1 , t_2 ,..., t_p]^T \in \mathbb{R}^p $, and each branch network outputs a scalar  represented by $b_k \in \mathbb{R} $, where $k = 1, 2...,p$. Next, the outputs generated by trunk and branch networks are  integrated together as $G(s)(y)  \approx \sum_{k=1}^{p} b_k (s(x_1), s(x_2),...s(x_m)) t_k (y)$. The unstacked DeepONet architecture is also shown in Fig.~\ref{fig:CombinedDeepONet}(a), which consists of only one branch network (depicted in dark blue) and one trunk network.  The unstacked DeepONet may be considered as a stacked DeepONet, in which all of the branch networks share the same set of parameters \cite{lu2021learning}. DeepONet was first used to learn several explicit operators, including integral and fractional Laplacians, along with implicit operators that represented deterministic and stochastic differential equations \citet{lu2021learning}. The two main advantages of DeepONets discussed by \citet{lu2021learning} are (i) small generalization error and (ii) rapid convergence of training and testing errors with respect to the quantity of the training data. 

\citet{lin2021operator} showed the effectiveness of DeepONet against data density and placement, which is advantageous when no prior knowledge is available on how much training data is required or when there are strict limits on data acquisition (e.g., location accessibility). To this end, they employed DeepONet and LSTM (i.e., PgNN) to model datasets representing the formation of a single bubble in response to time-varying changes in the ambient liquid pressure. To generate datasets, they used Rayleigh–Plesset (R–P) as a macroscopic model and dissipative particle dynamics (DPD) as a microscopic model. They used Gaussian random fields to generate different pressure fields, which serve as input signals for this dynamical system. The results of the comparison are shown in Fig.~\ref{fig:CombinedDeepONet}(b). The top row shows the prediction results for the liquid pressure trajectory when only 20 data points are known per trajectory, i.e., sparse training data, and the  bottom row shows the same but when 200 data points are known per trajectory, i.e., dense training data. As shown, regardless of how sparse the training data was, DeepONet was able to outperform LSTM to predict the liquid pressure trajectory. 

In addition, they examined a case where the input was not contained inside the training input range, i.e., when the correlation length of the pressure field was outside of the training range. In this case, they were initially unable to make accurate predictions, but mitigated the issue by transferring learning to a pre-trained DeepONet trunk network and fine-tuning it with only a few additional data points. They also demonstrated that  DeepONet can learn the mean component of the noisy raw data for the microscopic model without any additional data processing and that the computational time can be reduced from 48 CPU hours to a fraction of a second. These results confirmed that the  DeepONet model can be applied across macroscopic and microscopic regimes of bubble growth dynamics, establishing the foundation for a unified neural network model that can seamlessly predict physics interacting across scales. 

\citet{oommen2022learning} combined convolutional autoencoder architecture with DeepONet (CA-DeepONet) to learn the dynamic development of a two-phase mixture and speed up the time-to-solution for microstructure evolution prediction. In low-dimensional latent space, the convolutional autoencoder was utilized to provide a compact representation of microstructure data, while DeepONet was employed to learn mesoscale dynamics of microstructure evolution from the autoencoder's latent space. Then, the decoder component of the convolutional autoencoder reconstructs the evolution of the microstructure based on DeepONet's predictions. The trained DeepOnet architecture can then be used to speed up the numerical solver in extrapolation tasks or substitute the high-fidelity phase-field numerical solver in interpolation problems.

By taking inspiration from PiNNs for  sparse data domains, DeepONets can also be trained with very sparse labeled datasets  while incorporating known differential equations into the loss function. This approach results in Physics-informed DeepONets (Pi-DeepONets) \cite{wang2021learning,goswami2022physicscrack}. \citet{wang2021learning} employed Pi-DeepONets for benchmark problems such as diffusion reaction, Burger's equation, advection equation, and eikonal equation. In comparison to vanilla DeepONet, the result reveals significant improvements in predictive accuracy, generalization performance, and data efficiency. Furthermore,  Pi-DeepONets can learn a solution operator without any paired input-output training data, allowing them to simulate nonlinear and non-equilibrium processes in computational mechanics up to three orders of magnitude quicker than traditional solvers \cite{wang2021learning}. 

\begin{figure}[htp]
    \centering
    \includegraphics[width=0.98\linewidth]{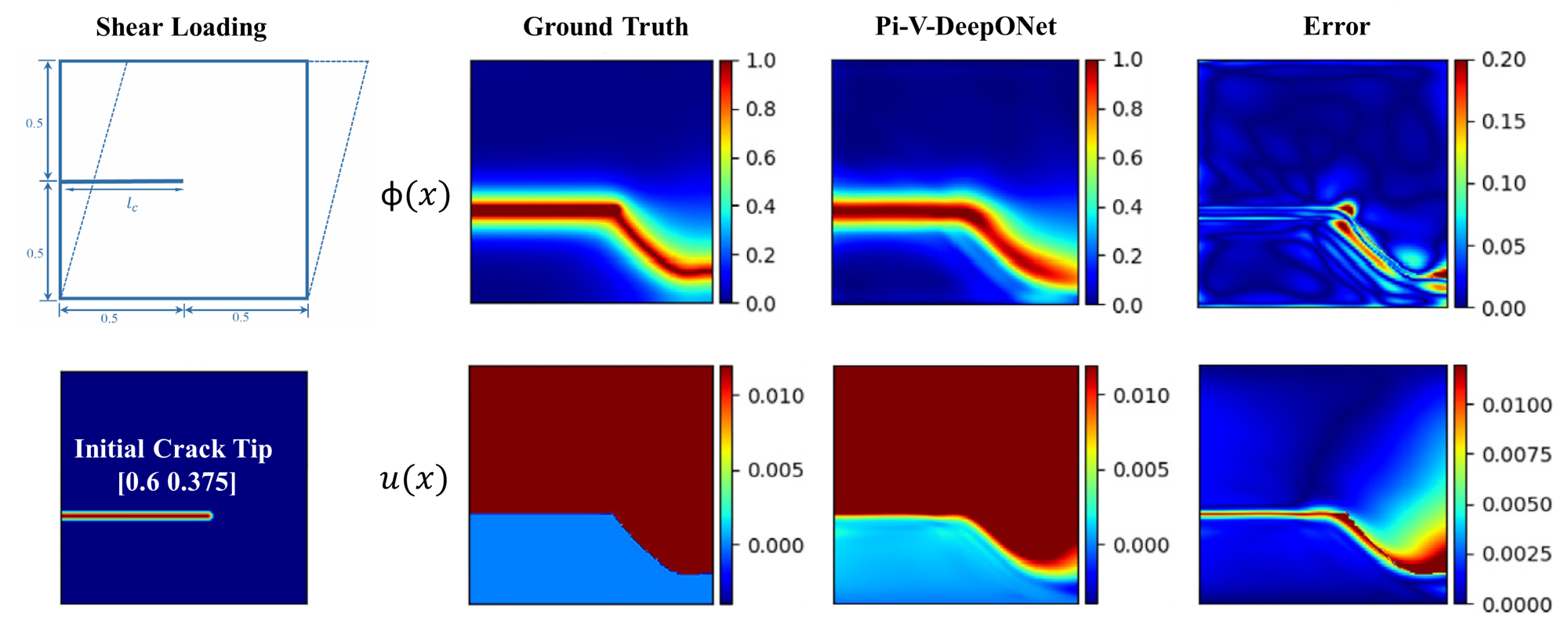}
    \caption {A comparison between Pi-V-DeepONet and isogeometric FEM analysis to predict the final damage path for the shear failure of a single-edge notched plate. The initial geometry of the single-edge notched plate, shear loading, and the initial configuration of the crack are shown on the left panels. The right panels show the comparisons for the phase field,  $\phi(x)$, and displacement along the x-axis, $u(x)$, as well as the corresponding error between the prediction and ground truth (adapted from \citet{goswami2022physicscrack}).}
    \label{fig:VDeepONet}
\end{figure}

\citet{goswami2022physicscrack} used a physics-informed variational formulation of DeepONet (Pi-V-DeepONet) for brittle fracture mechanics. The training of the Pi-V-DeepONet was conducted using the governing equations in a variational form and some labeled data. They used the Pi-V-DeepONet framework to determine failure pathways, failure zones, and damage along failure in brittle fractures for quasi-brittle materials. They trained the model to map the initial configuration of a defect (e.g., crack) to the relevant fields of interest (e.g., damage and displacements, see Fig.~\ref{fig:VDeepONet}). They showed that their model can rapidly predict the solution to any initial crack configuration and loading steps. In brittle fracture mechanics, the proposed model can be employed to enhance the design, evaluate reliability, and quantify uncertainty. 

Due to the high cost of evaluating integral operators,  DeepONets may face difficulty to develop effective numerical algorithms capable of replacing convolutional or recurrent neural networks in an infinite-dimensional context. \citet{li2020fourier} made an effort along this line and  developed an operator regression by  parameterizing the integral kernel in the Fourier space and termed it the Fourier neural operator (FNO). In the next section, we discuss the core architecture of FNO and the recent developments around it.

\subsection{Fourier Neural Operator (FNO)}

\begin{figure}[htp]
    \centering
    \includegraphics[width=.98\linewidth]{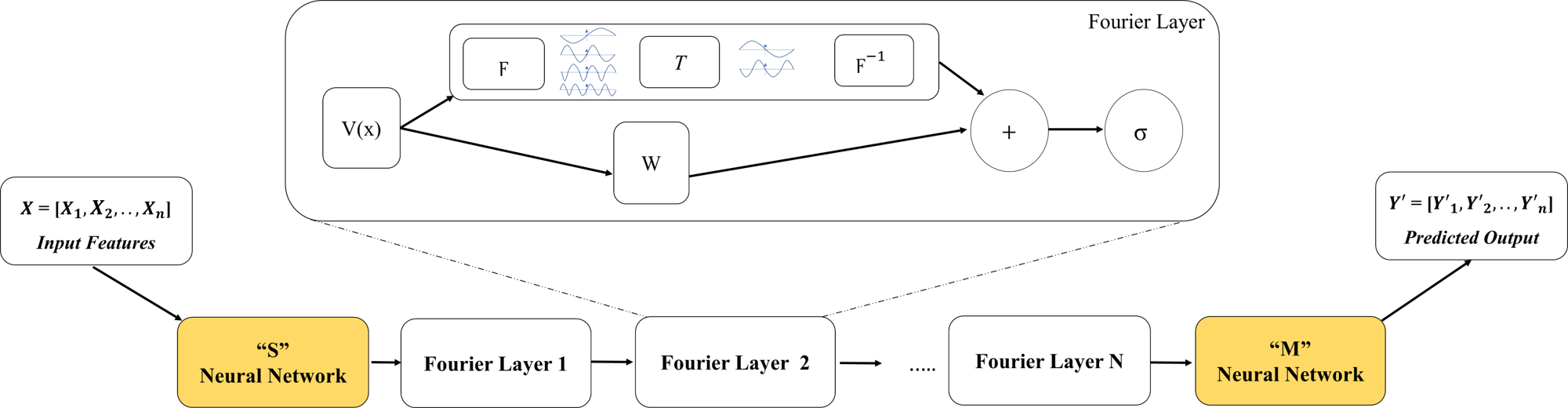}
    \caption {A schematic architecture of the Fourier Neural Operator (FNO) \cite{li2020fourier}. 
    Here, $\textbf{X}$ is the inputs, S and M are the neural network for dimensional space operation, and $\textbf{Y'}$ is the predicted outputs. Each of the Fourier layers consists of v(x) as the initial state, $\mathcal{F}$ layer that performs Fourier transform, $T$ layer that performs linear transform, $\mathcal{F}^{-1}$ layer that performs inverse Fourier transform, W layer for local linear transform, + block that carries out element-wise addition on W, and the final output from the $\mathcal{F}^{-1}$ layer. The final product from element-wise addition is passed on to a $\sigma$ layer. }
    \label{fig:FourierOperator}
\end{figure}

 In order to benefit from neural operators in infinite-dimensional spaces, \citet{li2020fourier} developed a neural operator in the Fourier space, dubbed as FNO, with a core architecture schematically shown in Fig.~\ref{fig:FourierOperator}. The training starts with an input $X$, which is subsequently elevated to a higher dimensional space by a neural network $S$. The second phase entails the use of several Fourier layers of integral operators and activation functions. In each Fourier layer, the input is transformed using (i) a Fourier transform, $F$; (ii) a linear transform, $T$, on the lower Fourier modes that  filters out the higher modes; and (iii) an inverse Fourier transform, $F^{-1}$. The input is also transformed using a local linear transform, $W$, before the application of the activation function, $\sigma$. The Fourier layers are designed to be discretization-invariant due to the fact that they learn from functions that are discretized arbitrarily. Indeed, the integral operator is applied in convolution and is represented as a linear transformation in the Fourier domain, allowing the FNO to learn the mapping over infinite-dimensional spaces. The result of the Fourier layer is  projected back to the target dimension in the third phase using another neural network $M$, which eventually outputs the desired output $Y'$  \cite{li2020fourier}. Unlike other DL methods, the FNO model's error is consistent regardless of the input and output resolutions (e.g., in PgNN methods, the error grows with the resolution). 

\begin{figure}[htp]
    \centering
    \includegraphics[width=.98\linewidth]{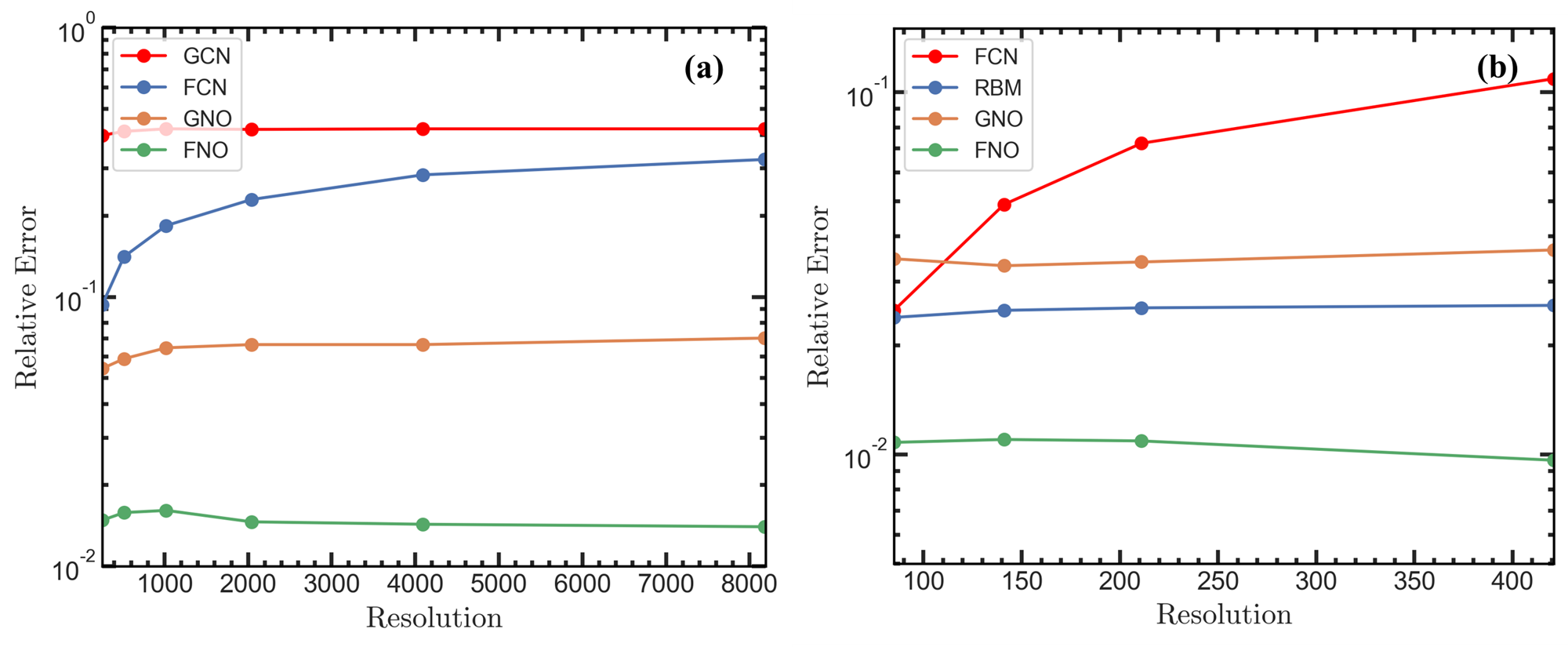}
    \caption {Error comparison between FNO and other state-of-the-art methods (reduced bias method (RBM), fully convolutional network (FCN), graph neural operator (GNO), and graph convolutional network (GCN)) for (a) Burgers' equation and (b) Darcy's Flow equation at different resolutions (adapted from \citet{li2020fourier}).}
    \label{fig:FourierResults}
\end{figure}

\citet{li2020fourier} employed FNO on three different test cases, including the 1D Burgers' equation, the 2D Darcy flow equation, and the 2D Navier-Stokes equations. For each test case, FNO was compared with state-of-the-art models. In particular, for Burgers' and Darcy's test cases, the methods used for comparison were the conventional ANN (i.e., PgNN),  reduced bias method \cite{devore2017theoretical}, fully convolutional networks \cite{zhu2018bayesian}, principal component analysis as an encoder in the neural network \cite{bhattacharya2020model}, graph neural operator \cite{li2020neuralb}, and low-rank decomposition neural operator (i.e., unstacked DeepONet \cite{lu2019deeponet}). In all test cases, FNO yielded the lowest relative error. The models' error comparisons for 1D Burgers' and 2D Darcy flow equations are depicted in Fig. \ref{fig:FourierResults}, adapted from \cite{li2020fourier}.

\begin{figure}[htp]
    \centering
    \includegraphics[width=0.98\linewidth]{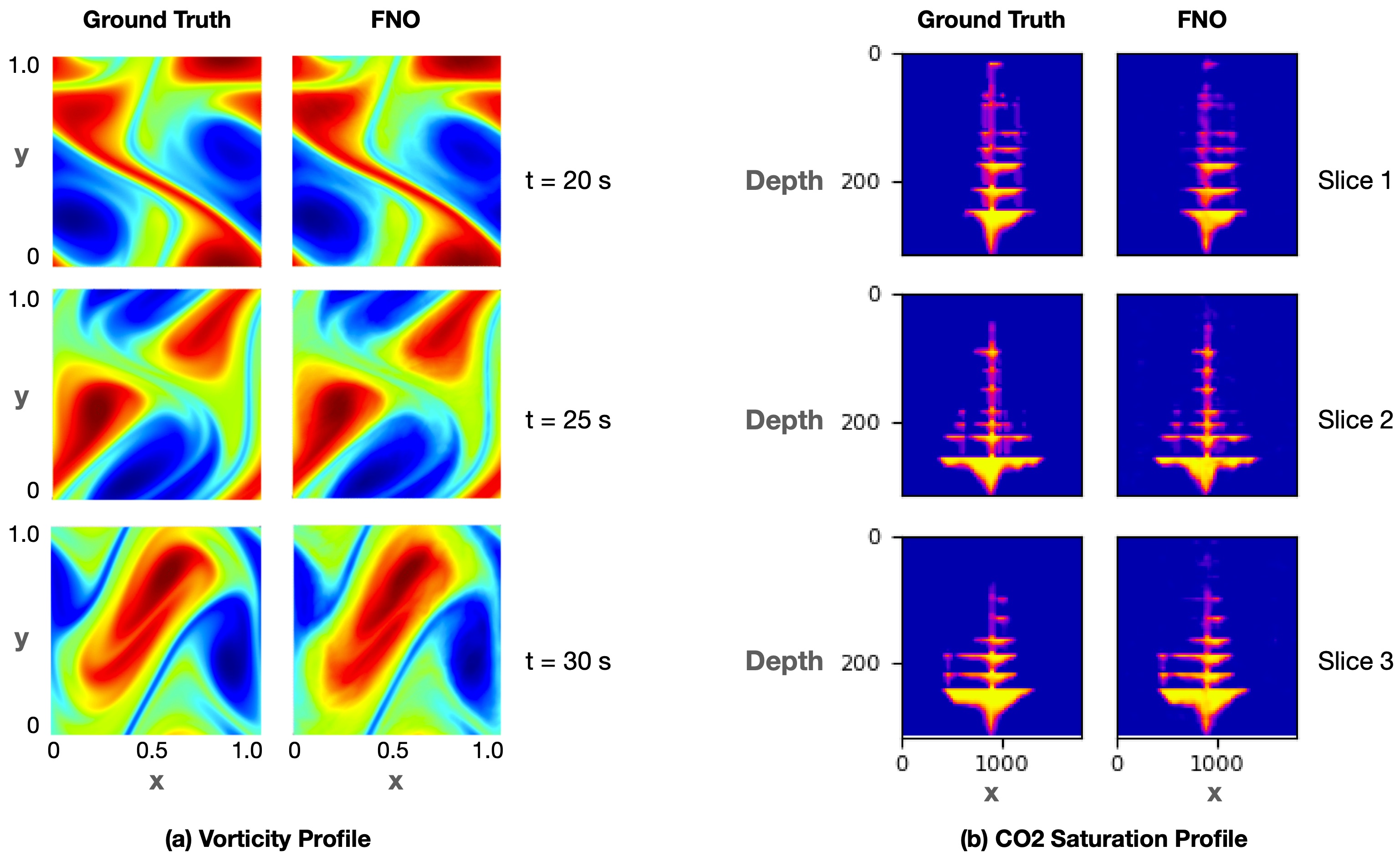}
    \caption {A qualitative comparison between predictions made by the Fourier 
    neural operator (FNO) and ground truth values. Panel (a) shows the comparison for a FNO model trained on 64$\times$64$\times$20 resolution ($n_x,n_y,n_t$) and evaluated on 256$\times$256$\times$80 resolution to solve 2D Navier-Stokes equations (adapted from \citet{li2020fourier}). Panel (b) shows the comparison for a parallelized FNO model trained to solve large-scale 3D subsurface CO2 flow modeling evaluated at 60$\times$60$\times$64$\times n_t$ resolution (adapted from \citet{grady2022towards}).}
    \label{fig:FourierResultsSample}
\end{figure}

The FNO model, as stated, can be trained on a specific resolution and tested on a different resolution. \citet{li2020fourier} demonstrated this claim by training a FNO on the Navier-Stokes equations for a 2D test case with a resolution of 64$\times$64$\times$20 ($n_x,n_y,n_t$) standing for spatial (x, y) and time resolution, and then evaluating it with a resolution of 256$\times$256$\times$80, as shown in Fig.~\ref{fig:FourierResultsSample}(a). In comparison to other models, the FNO was the only technique capable of performing resolution downscaling both spatially and temporally 
\cite{li2020fourier}. FNOs can also achieve several orders of magnitude speedup factors over conventional numerical PDE solvers. However, they have only been used for 2D or small 3D problems due to the large dimensionality of their input data, which increases the number of network weights significantly. With this problem in mind, \citet{grady2022towards} proposed a parallelized version of FNO based on domain-decomposition to resolve this limitation. Using this extension, they were able to use FNO in large-scale modeling, e.g., simulating the transient evolution of the CO2 plume in subsurface heterogeneous reservoirs as a part of the carbon capture and storage (CCS) technology \citep{bui2018carbon}, see Fig.~\ref{fig:FourierResultsSample}(b). The input to the network (with a similar architecture to the one proposed by \citet{li2020fourier}) was designed to be a tensor containing both the permeability and topography fields at each 3D spatial position using a 60$\times$60 $\times$64 ($n_x,n_y,n_z$) resolution and the output was 60$\times$60$\times$64$\times n_t$. For a time resolution of $n_t = 30 ~s$, they found that the parallelized FNO model was 271 times faster (without even leveraging GPU) than the conventional porous media solver while achieving comparable accuracy. \citet{wen2022u} also proposed U-FNO, an extension of FNO, to simulate multiphase flows in porous media, specifically CO2-water multiphase flow through a heterogeneous medium with broad ranges of reservoir conditions, injection configurations, flow rates, and multiphase flow properties. They compared U-FNO with FNO and CNN (i.e., PgNN) and showed that the U-FNO architecture provides the best performance for both gas saturation and pressure buildup predictions in highly heterogeneous geological formations. They also showed that the U-FNO architecture enhances the training accuracy of the original FNO, but does not naturally enable the flexibility of training and testing at multiple discretizations. 

\begin{figure}[htp]
    \centering
    \includegraphics[width=0.9\linewidth]{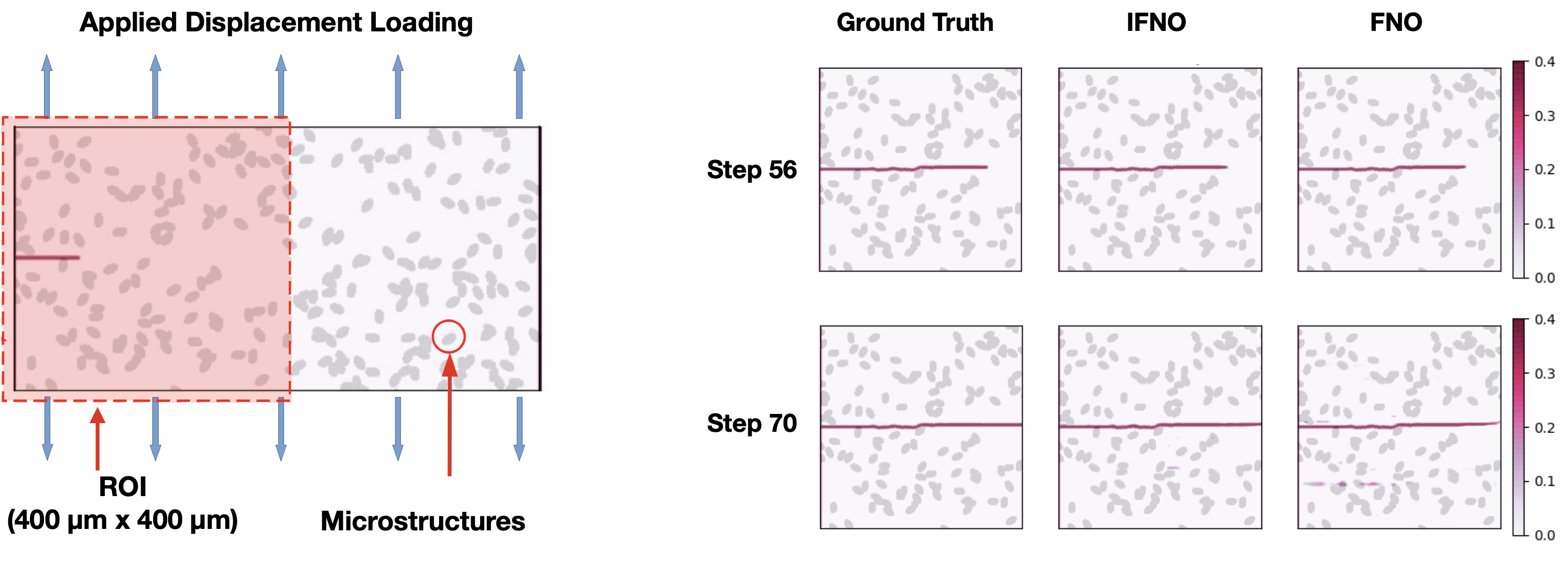}
    \caption {A comparison between the implicit Fourier neural operator (IFNO) and FNO models to predict crack propagation and damage field within a region of interest (ROI) in a pre-cracked glass-ceramics experiment with randomly distributed material property fields (adapted from \citet{you2022learning}).}
    \label{fig:INFO}
\end{figure}

\citet{you2022learning} proposed an implicit Fourier neural operator (IFNO) to model the complex responses of materials due to their heterogeneity and defects without using conventional constitutive models. The IFNO model captures the long-range dependencies in the feature space, and as the network becomes deeper, it becomes a fixed-point equation that yields an implicit neural operator (e.g., it can mimic displacement/damage fields). \citet{you2022learning} demonstrated the performance of IFNO using a series of test cases such as hyperelastic, anisotropic, and brittle materials. Fig.~\ref{fig:INFO} depicts a comparison between IFNO and  FNO for the transient propagation of a glass-ceramic crack \cite{you2022learning}. As demonstrated, IFNO outperforms  FNO (in terms of accuracy) and conventional constitutive models (in terms of computational cost) to predict the displacement field.

The FNO model has also been hybridized with PiNN to create the so-called physics-informed neural operator (PiNO) \cite{li2021physics}. The PiNO framework is a combination of operating-learning (i.e., FNO) and function-optimization (i.e., PiNN) frameworks that improves convergence rates and accuracy over both PiNN and FNO models. This integration was suggested to address the challenges in PiNN (e.g., generalization and optimization, especially for multiscale dynamical systems) and the challenges in FNO (e.g., the need for expensive and impractical large training datasets) \citep{li2021physics}. \citet{li2021physics} deployed the PiNO model on several benchmark problems (e.g., Kolmogorov flow, lid-cavity flow, etc.) to show that PiNO can outperform PiNN and FNO models while maintaining the FNO's exceptional speed-up factor over other solvers.

\begin{table}[H]
\centering
\caption {A non-exhaustive  list of recent studies that leveraged Neural Operators to model different scientific problems.}
\label{PeNN Table}
\begin{tabularx}{\textwidth}{ >{\hsize=.2\hsize}X >{\hsize=0.7\hsize}X>{\hsize=0.2\hsize}r}
\toprule  
  {NO Structure} &  {Objective }  &  {Reference}  \\ 
\midrule
DeepONets &   Learning non-linear operators accurately and efficiently with low generalization error   &  \cite{lu2021learning} \\
\addlinespace
\addlinespace
CADeepONet &   Learning dynamic development of a two-phase mixture and reducing solution time for predicting microstructure evolution &  \cite{oommen2022learning} \\
\addlinespace
\addlinespace
Pi-DeepONets & Integrating DeepONets and PiNN to relax the requirement for a large training dataset while improving generalization and predictive accuracy &  \cite{wang2021learning} \\
\addlinespace
\addlinespace
V-DeepONet & Generating a fast and generalizable surrogate model for brittle fracture mechanics &  \cite{goswami2022physicscrack} \\
\addlinespace
\addlinespace
FNO &   Providing a mesh- and discretization-invariant model to be inferred at any arbitrary spatiotemporal resolutions   &  \cite{li2020fourier} \\
\addlinespace
\addlinespace
Parallelized FNO &   Extending FNO based on domain-decomposition to model large-scale three-dimensional problems  &  \cite{grady2022towards} \\
\addlinespace
\addlinespace
U-FNO & Enhancing the training and testing accuracy of FNO for large-scale, highly heterogeneous geological formations   &  \cite{wen2022u} \\
\addlinespace
\addlinespace
IFNO & capturing the long-range dependencies in the feature space  that yields an implicit neural operator to model the complex responses of materials due to their heterogeneity and defects   &  \cite{you2022learning} \\
\addlinespace
\addlinespace

PINO  & Integrating PiNN and FNO to relax the requirement for a large training dataset while enhancing  generalization and optimization & \cite{li2021physics} \\
\bottomrule
    \end{tabularx}
\end{table}
 
\setlength{\arrayrulewidth}{1.5mm}
\setlength{\tabcolsep}{16pt}

\subsection{NOs Limitations}

DeepONet \cite{lu2019deeponet} and FNO \cite{li2020fourier}, as the two most common neural operators to date, share some commonalities while also having significant differences. The  DeepONet architecture was inspired by \citet{chen1995universal}'s universal approximation theorems, whereas the FNO was architecture established on parameterizing the integral kernel in Fourier space. However, FNO in its continuous form can be viewed as DeepONet with a specific architecture of the trunk  (expressed by a trigonometric basis) and  branch networks \cite{kovachki2021neural}. FNO, unlike DeepONet, discretizes both the input function and output function via point-wise evaluations in an equally spaced mesh.  Therefore, after network training, FNO can only predict the solution in the same mesh as the input function, but DeepONet can make predictions at any arbitrary location. FNO also requires a full field of observation data for training, whereas DeepONet is more flexible, with the exception of POD-DeepONet \cite{lu2022comprehensive}, which requires a full field of observation data to calculate the proper orthogonal decomposition (POD) modes \cite{lu2022comprehensive}. DeepONet, FNO, and their various variants still face some limitations, especially when applied to large multi-physics problems, that  necessitate further investigations.
\begin{itemize}

    \item Neural operators are purely data-driven and require relatively large  training datasets, therefore they face constraints when applied to problems where data acquisition is complex and/or costly   \cite{lu2022comprehensive}. Integration with PiNN can resolve this issue to some extent for problems where the underlying physics is fully known and can be integrated into the loss function \cite{wang2021learning}. Also, for practical applications, training the NOs just based on the governing equations in the loss function may produce inaccurate predictions; instead, a hybrid physics-data training  is recommended \cite{wang2021learning}. 

    \item  DeepONet and FNO are typically limited to basic geometry and/or structured data (e.g., 2D or small 3D problems)  due to the large dimensionality of their input data that increases the number of network weights significantly) \cite{lu2022comprehensive}. They are also prone to over-fitting as the number of trainable parameters increases, making the training process more difficult \cite{you2022nonlocal}. IFNOs have addressed this challenge to some extent \cite{you2022learning}. In IFNO, the solution operator is first formulated as an implicitly defined mapping and then  modeled as a fixed point. The latter aims to overcome the challenge of network training in the case of deep layers, and the former minimizes the number of trainable parameters and memory costs. Nonetheless, due to the finite size of the neural network architecture, the convergence of  NOs (e.g., DeepONet) error with respect to the size of the training data becomes algebraic  for large datasets; it is desired to be exponential \cite{lu2022comprehensive}. 

    \item FNO might not be reliable for discontinuous functions as it relies on the Fourier transformation. This is mitigated to some extent by DeepONet, as it was shown to perform well for functions with discontinuity (e.g., compressible Euler equations) \cite{lu2022comprehensive}. 
    
\end{itemize}

Despite these limitations, neural operators are the leading algorithms in a variety of real-time inference applications, including autonomous systems, surrogates in design problems, and uncertainty quantification \cite{goswami2022physicsReview}.

\section{Conclusions and Future Research Directions}

A considerable number of research topics collectively support the efficacy of combining scientific computing and deep learning approaches. In particular, this combination improves the efficiency of both forward and inverse modeling for high-dimensional problems that are prohibitively expensive, contain noisy data, require a complex mesh, and are governed by non-linear, ill-posed differential equations. The ever-increasing computer power will continue to further furnish this combination by allowing the use of deeper neural networks and considering higher-dimensional interdependencies and design space.

\begin{table}[H]
\centering
\caption {A comparison of the main characteristics of PgNNs, PiNNs, PeNNs, and NOs to model non-linear multiscale phenomena in computational fluid and solid mechanics.}
\label{Difference architectures}
 \begin{tabular}{l c c c c}
 \toprule  
 Feature & {PgNNs} & {PiNNs} & {PeNNs} & {NOs}   \\ 
 \midrule
 Accelerating Capability & \checkmark  & \checkmark & \checkmark  & \checkmark    \\
 \addlinespace
 \addlinespace
  Mesh-free Simulation & \checkmark  & \checkmark & \checkmark & \checkmark    \\
 \addlinespace
 \addlinespace
   Straightforward Network Training  & \checkmark  & \xmark & \xmark  & \xmark  \\
\addlinespace
 \addlinespace
   Training without Labeled Data & \xmark  & \checkmark & \xmark & \xmark   \\
 \addlinespace
 \addlinespace
 Physics-informed Loss Function & \xmark & \checkmark & \checkmark  & \checkmark     \\
  \addlinespace
 \addlinespace
Continuous Solution & \xmark &  \checkmark & \checkmark & \checkmark   \\
 \addlinespace
 \addlinespace
Spatiotemporal Interpolation & \xmark &  \checkmark & \checkmark  & \checkmark  \\ 
 \addlinespace
 \addlinespace
 Physics Encoding & \xmark &  \xmark & \checkmark  & \xmark  \\
 \addlinespace
 \addlinespace
Efficient Operator Learning & \xmark &  \xmark  & \xmark & \checkmark    \\
 \addlinespace
 \addlinespace
Continuous-depth Models & \xmark &  \xmark & \checkmark    &  \xmark \\
  \addlinespace
 \addlinespace
Spatiotemporal Extrapolation & \xmark &  \xmark & \checkmark  & \checkmark  \\ 
  \addlinespace
 \addlinespace
Solution Transferability & \xmark &  \xmark & \checkmark   & \checkmark \\ 
  \addlinespace
 \addlinespace
Efficient Real-time Predictions & \xmark &  \xmark & \xmark   & \checkmark \\ 

 \bottomrule
 \end{tabular}
\end{table}

The combination of scientific computing and deep learning approaches also surpasses traditional computational mechanics solvers in a number of prevalent scenarios in practical engineering. For example, a sparse dataset obtained experimentally for a complex (i.e., hard-to-acquire data) phenomenon cannot be simply integrated with traditional solvers. Whereas using DL, the following tasks can be performed: (i) PgNN-based models can be applied to the sparse data to extract latent interdependencies and conduct spatiotemporal downscaling or upscaling (i.e., interpolated data); (ii) PiNN-based models can be applied to the interpolated data to deduce governing equations and potentially unknown boundary or initial conditions of the phenomena (i.e., strong mathematical form); (iii) PeNN-based models can be used to combine the interpolated data and the strong mathematical form to conduct extrapolation exploration; and (iv) NO-based models can be applied to make real-time predictions of the complex dynamics. Therefore, the combination of DL-based methods and traditional scientific computing methods provides scientists with a cost-effective toolbox to explore problems across different scales that were deemed far-fetched computationally. To this end, several other breakthroughs in DL are required to enable the use of PgNNs, PiNNs, PeNNs, and NOs in large-scale three-dimensional (or multi-dimensional) problems. For instance, the training of complex DL models (e.g., PiNNs,  PeNNs, and NOs) should be accelerated using different parallelization paradigms.    

Table \ref{Difference architectures} compares the main characteristics of the PgNNs, PiNNs, PeNNs, and NOs. The PgNN-based models suffer mainly from their statistical training process, for which they require large datasets. They map carefully curated training datasets  only based on correlations in statistical variations, and hence, their predictions are naturally physics-agnostic. The PiNN-based models suffer mainly from the presence of competing loss terms that may destabilize the training process. PiNN is also a solution learning algorithm with limited generalizability due to its inability to learn the physical operation of a specific phenomenon. Models based on PeNNs and NOs, on the other hand, may experience low convergence rates and require a large volume of paired, structured datasets, leading to highly expensive training. 

Considering the effectiveness of this new challenge of combining scientific computing and DL, future studies can be divided into three distinct categories: (i) \textbf{Improving algorithms}: Developing advanced variants of PgNNs, PiNNs, PeNNs and NOs  that offer simpler implementation with enhanced convergence rate; faster training in multi-dimensional and multi-physics problems; higher accuracy and generalization to unseen conditions while using sparse training datasets, more robust to be used in real time forecasting; better adaptability to multi-spatiotemporal-resolutions, more flexibility to encode various types of governing equations (e.g., all PDE types, closure laws, data-driven laws, etc.), and provide a closer tie with a plethora of traditional solvers; (ii) \textbf{Considering causalities}: Developing a causal training algorithm (e.g., causal Q-learning \cite{molina2020causal}) that restores physical causality during the training of PgNN, PiNN, and PeNN models by re-weighting the governing equations (e.g., PDEs) residual loss at each iteration. This line of research will allow for the development of causality-conforming variants of PgNN, PiNN, and PeNN algorithms that can bring new opportunities for the application of these algorithms to a wider variety of complex scenarios across diverse domains; (iii) \textbf{Expanding applications}: Leveraging the potentials of PgNNs, PiNNs,  PeNNs, and NOs in problems with complex anisotropic materials (e.g., flow in highly heterogeneous porous media, metal and non-metal particulate composites, etc.); problems with multiscale multi-physics phenomena (e.g., magnetorheological fluids, particle-laden fluids, dry powder dynamics, reactive transport, unsaturated soil dynamics,  etc.); problems with multi-resolution objectives and extensive spatiotemporal downscaling or upscaling  (e.g., global and regional climate modeling, geosystem reservoir modeling, etc.); and structural health monitoring (e.g., crack identification and propagation, hydrogen pipeline leakage, CO2 plume detection, etc.);  and (iv) \textbf{Coupling solvers}: Coupling PgNNs, PiNNs, and PeNNs as well as NOs  with open-source computational mechanics packages such as OpenIFEM, OpenFOAM, Palabos, LAMMPS, LIGGGHTS, MOOSE, etc. This line of research will allow for faster surrogate modeling and, hence, faster development of next-generation solvers. It also expedites community and industry adoption of the combined scientific-DL computational paradigm.

\section{Acknowledgements}
S.A.F. would like to acknowledge supports from the Department of Energy Biological and Environmental Research (BER) (award no. DE-SC0023044), National Science Foundation Partnership for Research and Education in Materials (PREM) (award no. DMR-2122041), and Texas State University Multidisciplinary Internal Research Grant (MIRG) (award no. 9000003028).

C.F. would like to acknowledge supports from FEDER funds through the COMPETE 2020 Programme and National Funds through FCT (Portuguese Foundation for Science and Technology) under the projects UID-B/05256/2020, UID-P/05256/2020 and MIT-EXPL/TDI/0038/2019-APROVA-Deep learning for particle-laden viscoelastic flow modelling (POCI-01-0145-FEDER-016665) under the MIT Portugal program.

\section{Conflict of Interest}
The authors declare no conflict of interest.

\def\mybibdoicolor{\color{black}}
\newcommand*{\doi}[1]{\href{\detokenize{#1}} {\raggedright\mybibdoicolor{DOI: \detokenize{#1}}}}

\bibliographystyle{unsrtnat}
\bibliography{references}

\begin{thebibliography}{312}
\providecommand{\natexlab}[1]{#1}
\providecommand{\url}[1]{\texttt{#1}}
\expandafter\ifx\csname urlstyle\endcsname\relax
  \providecommand{\doi}[1]{doi: #1}\else
  \providecommand{\doi}{doi: \begingroup \urlstyle{rm}\Url}\fi

\bibitem[Vinuesa and Brunton(2022)]{vinuesa2022enhancing}
Ricardo Vinuesa and Steven~L Brunton.
\newblock Enhancing computational fluid dynamics with machine learning.
\newblock \emph{Nature Computational Science}, 2\penalty0 (6):\penalty0
  358--366, 2022.
\newblock \doi{https://doi.org/10.1038/s43588-022-00264-7}.

\bibitem[Mianroodi et~al.(2021)Mianroodi, Siboni, and Raabe]{Raabe2021}
J.~R. Mianroodi, N.~H. Siboni, and D.~Raabe.
\newblock Teaching solid mechanics to artificial intelligence - a fast solver
  for heterogeneous materials.
\newblock \emph{NPJ Computational Materials}, 7:\penalty0 99, 2021.
\newblock \doi{https://doi.org/10.1038/s41524-021-00571-z}.

\bibitem[Kim et~al.(2021)Kim, Yang, Park, Gu, and Seunghwa]{Kim2021}
Y.~Kim, C.~Yang, K.~Park, G.~X. Gu, and R.~Seunghwa.
\newblock Deep learning framework for material design space exploration using
  active transfer learning and data augmentation.
\newblock \emph{npj Computational Materials}, 7:\penalty0 140, 2021.
\newblock \doi{https://doi.org/10.1038/s41524-021-00609-2}.

\bibitem[Dino et~al.(2020)Dino, Zeebaree, Salih, Zebari, Ageed, Shukur, Haji,
  and Hasan]{dino2020impact}
H.~I. Dino, S.~R. Zeebaree, A.~A. Salih, R.~R. Zebari, Z.~S. Ageed, H.~M.
  Shukur, L.~M. Haji, and S.~S. Hasan.
\newblock Impact of process execution and physical memory-spaces on os
  performance.
\newblock \emph{Technology Reports of Kansai University}, 62\penalty0
  (5):\penalty0 2391--2401, 2020.

\bibitem[Im et~al.(2021)Im, Lee, and Cho]{im2021surrogate}
S.~Im, J.~Lee, and M.~Cho.
\newblock Surrogate modeling of elasto-plastic problems via long short-term
  memory neural networks and proper orthogonal decomposition.
\newblock \emph{Computer Methods in Applied Mechanics and Engineering},
  385:\penalty0 114030, 2021.
\newblock \doi{https://doi.org/10.1016/j.cma.2021.114030}.

\bibitem[Karniadakis et~al.(2021)Karniadakis, Kevrekidis, Lu, Perdikaris, Wang,
  and Yang]{karniadakis2021physics}
G.~E. Karniadakis, I.~G. Kevrekidis, L.~Lu, P.~Perdikaris, S.~Wang, and
  L.~Yang.
\newblock Physics-informed machine learning.
\newblock \emph{Nature Reviews Physics}, 3\penalty0 (6):\penalty0 422--440,
  2021.
\newblock \doi{https://doi.org/10.1038/s42254-021-00314-5}.

\bibitem[Arman and Anthony(2022)]{arman2022compfluids}
S.~Arman and W.~Anthony.
\newblock Physics-inspired architecture for neural network modeling of forces
  and torques in particle-laden flows.
\newblock \emph{Computers and Fluids}, 238:\penalty0 105379, 2022.
\newblock \doi{https://doi.org/10.1016/j.compfluid.2022.105379}.

\bibitem[Innes et~al.(2019)Innes, Edelman, Fischer, Rackauckas, Saba, Shah, and
  Tebbutt]{innes2019differentiable}
M.~Innes, A.~Edelman, K.~Fischer, C.~Rackauckas, E.~Saba, V.B. Shah, and
  W.~Tebbutt.
\newblock A differentiable programming system to bridge machine learning and
  scientific computing.
\newblock \emph{arXiv preprint arXiv:1907.07587}, 2019.
\newblock \doi{https://doi.org/10.48550/arXiv.1907.07587}.

\bibitem[Brunton et~al.(2020)Brunton, Noack, and
  Koumoutsakos]{brunton2020machine}
S.~L. Brunton, B.~R. Noack, and P.~Koumoutsakos.
\newblock Machine learning for fluid mechanics.
\newblock \emph{Annual Review of Fluid Mechanics}, 52:\penalty0 477--508, 2020.
\newblock \doi{https://doi.org/10.1146/annurev-fluid-010719-060214}.

\bibitem[Cai et~al.(2022{\natexlab{a}})Cai, Mao, Wang, Yin, and
  Karniadakis]{cai2022physics}
S.~Cai, Z.~Mao, Z.~Wang, M.~Yin, and G.~E. Karniadakis.
\newblock Physics-informed neural networks (pinns) for fluid mechanics: A
  review.
\newblock \emph{Acta Mechanica Sinica}, pages 1--12, 2022{\natexlab{a}}.
\newblock \doi{https://doi.org/10.1007/s10409-021-01148-1}.

\bibitem[Kutz(2017)]{Kutz2017}
J.~N. Kutz.
\newblock Deep learning in fluid dynamics.
\newblock \emph{Journal of Fluid Mechanics}, 814:\penalty0 1--4, 2017.
\newblock \doi{https://doi.org/10.1017/jfm.2016.803}.

\bibitem[Shi et~al.(2019)Shi, Tsymbalov, Dao, Suresh, Shapeev, and Li]{Shi2019}
Z.~Shi, E.~Tsymbalov, M.~Dao, S.~Suresh, A.~Shapeev, and J.~Li.
\newblock Deep elastic strain engineering of bandgap through machine learning.
\newblock \emph{Proc. Natl. Acad. Sci.}, 116:\penalty0 4117--4122, 2019.
\newblock \doi{https://doi.org/10.1073/pnas.1818555116}.

\bibitem[Haghighat et~al.(2021{\natexlab{a}})Haghighat, Raissi, Moure, Gomez,
  and Juanes]{haghighat2021physicsa}
E.~Haghighat, M.~Raissi, A.~Moure, H.~Gomez, and R.~Juanes.
\newblock A physics-informed deep learning framework for inversion and
  surrogate modeling in solid mechanics.
\newblock \emph{Computer Methods in Applied Mechanics and Engineering},
  379:\penalty0 113741, 2021{\natexlab{a}}.
\newblock \doi{https://doi.org/10.1016/j.cma.2021.113741}.

\bibitem[Pilania et~al.(2013)Pilania, Wang, X., Rajasekaran, and
  Ramprasad]{Pilania2013}
G.~Pilania, C.~Wang, Jiang X., S.~Rajasekaran, and R.~Ramprasad.
\newblock Accelerating materials property predictions using machine learning.
\newblock \emph{Sci. Rep.}, 3:\penalty0 1--6, 2013.
\newblock \doi{https://doi.org/10.1038/srep02810}.

\bibitem[Butler et~al.(2018)Butler, Davies, Cartwright, Isayev, and
  Walsh]{Butler2018}
K.~T. Butler, D.~W. Davies, H.~Cartwright, O.~Isayev, and A.~Walsh.
\newblock Machine learning for molecular and materials science.
\newblock \emph{Nature}, 559:\penalty0 547--555, 2018.
\newblock \doi{https://doi.org/10.1038/s41586-018-0337-2}.

\bibitem[Brunton and Kutz(2019)]{Brunton2019}
S.~L. Brunton and J.~N. Kutz.
\newblock Methods for data-driven multiscale model discovery for materials.
\newblock \emph{J. Phys. Mater.}, 2:\penalty0 044002, 2019.
\newblock \doi{https://doi.org/10.1088/2515-7639/ab291e}.

\bibitem[Bedolla et~al.(2020)Bedolla, Padierna, and
  Castaneda-Priego]{bedolla2020machine}
E.~Bedolla, L.~C. Padierna, and R.~Castaneda-Priego.
\newblock Machine learning for condensed matter physics.
\newblock \emph{Journal of Physics: Condensed Matter}, 33\penalty0
  (5):\penalty0 053001, 2020.
\newblock \doi{https://doi.org/10.1088/1361-648X/abb895}.

\bibitem[Kochkov et~al.(2021)Kochkov, Smith, Alieva, Wang, Brenner, and
  Hoyer]{Kochkov2021}
D.~Kochkov, J.~A. Smith, A.~Alieva, Q.~Wang, M.~P. Brenner, and S.~Hoyer.
\newblock Machine learning–accelerated computational fluid dynamics.
\newblock \emph{Proceedings of the National Academy of Sciences}, 118:\penalty0
  e2101784118, 2021.
\newblock \doi{https://doi.org/10.1073/pnas.2101784118}.

\bibitem[Tran et~al.(2017)Tran, Hoffman, Saurous, Brevdo, Murphy, and
  Blei]{tran2017deep}
D.~Tran, M.~D. Hoffman, R.~A. Saurous, E.~Brevdo, K.~Murphy, and D.M. Blei.
\newblock Deep probabilistic programming.
\newblock \emph{arXiv preprint arXiv:1701.03757}, 2017.
\newblock \doi{https://doi.org/10.48550/arXiv.1701.03757}.

\bibitem[Jin et~al.(2017)Jin, McCann, Froustey, and Unser]{jin2017deep}
K.~H. Jin, M.T. McCann, E.~Froustey, and M.~Unser.
\newblock Deep convolutional neural network for inverse problems in imaging.
\newblock \emph{IEEE Transactions on Image Processing}, 26\penalty0
  (9):\penalty0 4509--4522, 2017.
\newblock \doi{https://doi.org/10.1109/TIP.2017.2713099}.

\bibitem[Lai et~al.(2022)Lai, Chen, and Huang]{lai2022machine}
Z.~Lai, Q.~Chen, and L.~Huang.
\newblock Machine-learning-enabled discrete element method: Contact detection
  and resolution of irregular-shaped particles.
\newblock \emph{International Journal for Numerical and Analytical Methods in
  Geomechanics}, 46\penalty0 (1):\penalty0 113--140, 2022.
\newblock \doi{https://doi.org/10.1002/nag.3293}.

\bibitem[Faroughi et~al.(2022{\natexlab{a}})Faroughi, Roriz, and
  Fernandes]{faroughi2022meta}
S.~A. Faroughi, A.~I. Roriz, and C.~Fernandes.
\newblock A meta-model to predict the drag coefficient of a particle
  translating in viscoelastic fluids: a machine learning approach.
\newblock \emph{Polymers}, 14\penalty0 (3):\penalty0 430, 2022{\natexlab{a}}.
\newblock \doi{https://doi.org/10.3390/polym14030430}.

\bibitem[Taylor(1715)]{taylor1715methodus}
B.~Taylor.
\newblock \emph{Methodus incrementorum directa \& inversa. Auctore Brook
  Taylor, LL. D. \& Regiae Societatis Secretario}.
\newblock typis Pearsonianis: prostant apud Gul. Innys ad Insignia Principis
  in~…, 1715.

\bibitem[Alder and Wainwright(1957)]{alder1957phase}
B.~J. Alder and T.~E. Wainwright.
\newblock Phase transition for a hard sphere system.
\newblock \emph{The Journal of chemical physics}, 27\penalty0 (5):\penalty0
  1208--1209, 1957.

\bibitem[Clough(1960)]{clough1960finite}
R.~W. Clough.
\newblock The finite element method in plane stress analysis.
\newblock In \emph{Proceedings of 2nd ASCE Conference on Electronic
  Computation, Pittsburgh Pa., Sept. 8 and 9, 1960}, 1960.

\bibitem[Smagorinsky(1963)]{smagorinsky1963general}
J.~Smagorinsky.
\newblock General circulation experiments with the primitive equations: I. the
  basic experiment.
\newblock \emph{Monthly weather review}, 91\penalty0 (3):\penalty0 99--164,
  1963.

\bibitem[Cundall and Strack(1979)]{Cundall197947}
P.~A. Cundall and O.~Strack.
\newblock A discrete numerical model for granular assemblies.
\newblock \emph{G\'eotechnique}, 29:\penalty0 47--65, 1979.
\newblock \doi{https://doi.org/10.1680/geot.1979.29.1.47}.

\bibitem[McDonald(1971)]{mcdonald1971computation}
P.~W. McDonald.
\newblock \emph{The computation of transonic flow through two-dimensional gas
  turbine cascades}, volume 79825.
\newblock American Society of Mechanical Engineers, 1971.

\bibitem[Peskin(1972)]{peskin1972flow}
C.~S. Peskin.
\newblock Flow patterns around heart valves: a numerical method.
\newblock \emph{Journal of computational physics}, 10\penalty0 (2):\penalty0
  252--271, 1972.

\bibitem[Lucy(1977)]{lucy1977numerical}
L.~B. Lucy.
\newblock A numerical approach to the testing of the fission hypothesis.
\newblock \emph{The astronomical journal}, 82:\penalty0 1013--1024, 1977.

\bibitem[D'Humieres et~al.(1986)D'Humieres, Lallemand, and
  Frisch]{d1986lattice}
D.~D'Humieres, P.~Lallemand, and U.~Frisch.
\newblock Lattice gas models for 3d hydrodynamics.
\newblock \emph{EPL (Europhysics Letters)}, 2\penalty0 (4):\penalty0 291, 1986.

\bibitem[Bassi and Rebay(1997)]{bassi1997high}
F.~Bassi and S.~Rebay.
\newblock A high-order accurate discontinuous finite element method for the
  numerical solution of the compressible navier--stokes equations.
\newblock \emph{Journal of computational physics}, 131\penalty0 (2):\penalty0
  267--279, 1997.

\bibitem[Ivakhnenko and Lapa(1967)]{ivakhnenko1967cybernetics}
A.~G. Ivakhnenko and V.~G. Lapa.
\newblock \emph{Cybernetics and forecasting techniques}, volume~8.
\newblock American Elsevier Publishing Company, 1967.

\bibitem[Rumelhart et~al.(1986)Rumelhart, Hinton, and
  Williams]{rumelhart1986learning}
D.~E. Rumelhart, G.~E. Hinton, and R.~J. Williams.
\newblock Learning representations by back-propagating errors.
\newblock \emph{nature}, 323\penalty0 (6088):\penalty0 533--536, 1986.

\bibitem[Andersen et~al.(1990)Andersen, Cook, Karsai, and
  Ramaswamy]{andersen1990artificial}
K.~Andersen, G.~E. Cook, G.~Karsai, and K.~Ramaswamy.
\newblock Artificial neural networks applied to arc welding process modeling
  and control.
\newblock \emph{IEEE Transactions on industry applications}, 26\penalty0
  (5):\penalty0 824--830, 1990.
\newblock \doi{https://doi.org/10.1109/28.60056}.

\bibitem[LeCun et~al.(1998)LeCun, Bottou, Bengio, and
  Haffner]{lecun1998gradient}
Y.~LeCun, L.~Bottou, Y.~Bengio, and P.~Haffner.
\newblock Gradient-based learning applied to document recognition.
\newblock \emph{Proceedings of the IEEE}, 86\penalty0 (11):\penalty0
  2278--2324, 1998.

\bibitem[J. et~al.(2014)J., Jean, Mehdi, Bing, David, Sherjil, and
  Courville~Aaron]{goodfellow2014generative}
Goodfellow~I. J., P.~Jean, M.~Mehdi, X.~Bing, W.~David, O.~Sherjil, and
  C.~Courville~Aaron.
\newblock Generative adversarial nets.
\newblock In \emph{Proceedings of the 27th international conference on neural
  information processing systems}, volume~2, pages 2672--2680, 2014.

\bibitem[Raissi et~al.(2017)Raissi, Perdikaris, and
  Karniadakis]{raissi2017physics}
M.~Raissi, P.~Perdikaris, and G.~E. Karniadakis.
\newblock Physics informed deep learning (part i): Data-driven solutions of
  nonlinear partial differential equations.
\newblock \emph{arXiv preprint arXiv:1711.10561}, 2017.
\newblock \doi{https://doi.org/10.48550/arXiv.1711.10561}.

\bibitem[Lu et~al.(2019)Lu, Jin, and Karniadakis]{lu2019deeponet}
L.~Lu, P.~Jin, and G.~E. Karniadakis.
\newblock Deeponet: Learning nonlinear operators for identifying differential
  equations based on the universal approximation theorem of operators.
\newblock \emph{arXiv preprint arXiv:1910.03193}, 2019.
\newblock \doi{https://doi.org/10.48550/arXiv.1910.03193}.

\bibitem[Rao et~al.({\natexlab{a}})Rao, Sun, and Liu]{rao2021hard}
C.~Rao, H.~Sun, and Y.~Liu.
\newblock Hard encoding of physics for learning spatiotemporal dynamics.
\newblock \emph{arXiv preprint arXiv:2105.00557}, {\natexlab{a}}.
\newblock \doi{https://doi.org/10.48550/arXiv.2105.00557}.

\bibitem[Lienen and Gunnemann(2022)]{Lienen2022}
M.~Lienen and S.~Gunnemann.
\newblock Learning the dynamics of physical systems from sparse observations
  with finite element networks.
\newblock In \emph{The Tenth International Conference on Learning
  Representations}. OpenReview, 2022.
\newblock \doi{https://doi.org/10.48550/arXiv.2203.08852}.

\bibitem[Hasson et~al.(2020)Hasson, Nastase, and Goldstein]{hasson2020direct}
U.~Hasson, S.A. Nastase, and A.~Goldstein.
\newblock Direct fit to nature: an evolutionary perspective on biological and
  artificial neural networks.
\newblock \emph{Neuron}, 105\penalty0 (3):\penalty0 416--434, 2020.
\newblock \doi{https://doi.org/10.1016/j.neuron.2019.12.002}.

\bibitem[Li et~al.(2021{\natexlab{a}})Li, Zheng, Kovachki, Jin, Chen, Liu,
  Azizzadenesheli, and Anandkumar]{li2021physics}
Z.~Li, H.~Zheng, N.~Kovachki, D.~Jin, H.~Chen, B.~Liu, K.~Azizzadenesheli, and
  A.~Anandkumar.
\newblock Physics-informed neural operator for learning partial differential
  equations.
\newblock \emph{arXiv preprint arXiv:2111.03794}, 2021{\natexlab{a}}.
\newblock \doi{https://doi.org/10.48550/arXiv.2111.03794}.

\bibitem[Raissi et~al.(2019)Raissi, Perdikaris, and
  Karniadakis]{raissi2019physics}
M.~Raissi, P.~Perdikaris, and G.~E. Karniadakis.
\newblock Physics-informed neural networks: A deep learning framework for
  solving forward and inverse problems involving nonlinear partial differential
  equations.
\newblock \emph{Journal of Computational Physics}, 378:\penalty0 686--707,
  2019.
\newblock \doi{https://doi.org/10.1016/j.jcp.2018.10.045}.

\bibitem[Nabian and Meidani(2020)]{nabian2020adaptive}
M.~A. Nabian and H.~Meidani.
\newblock Adaptive physics-informed neural networks for markov-chain monte
  carlo.
\newblock \emph{arXiv preprint arXiv:2008.01604}, 2020.
\newblock \doi{https://doi.org/10.48550/arXiv.2008.01604}.

\bibitem[Cuomo et~al.(2022)Cuomo, Di~Cola, Giampaolo, Rozza, Raissi, and
  Piccialli]{cuomo2022scientific}
S.~Cuomo, V.~S. Di~Cola, F.~Giampaolo, G.~Rozza, M.~Raissi, and F.~Piccialli.
\newblock Scientific machine learning through physics-informed neural networks:
  Where we are and what's next.
\newblock \emph{arXiv preprint arXiv:2201.05624}, 2022.
\newblock \doi{https://doi.org/10.48550/arXiv.2201.05624}.

\bibitem[Baydin et~al.(2015)Baydin, Pearlmutter, Radul, and
  Siskind]{baydin2015}
A.~G. Baydin, B.~A. Pearlmutter, A.~A. Radul, and J.~M. Siskind.
\newblock Automatic differentiation in machine learning: a survey.
\newblock \emph{arXiv:1502.05767}, 2015.
\newblock \doi{https://doi.org/10.48550/arXiv.1502.05767}.

\bibitem[Rao et~al.(2020{\natexlab{a}})Rao, Hao, and Yang]{rao2020}
C.~Rao, S.~Hao, and L.~Yang.
\newblock Physics-informed deep learning for incompressible laminar flows.
\newblock \emph{Theoretical and Applied Mechanics Letters}, 10:\penalty0
  207--212, 2020{\natexlab{a}}.
\newblock \doi{https://doi.org/10.1016/j.taml.2020.01.039}.

\bibitem[Faroughi et~al.(2022{\natexlab{b}})Faroughi, Datta, Mahjour, and
  Faroughi]{faroughi2022physics}
Salah~A Faroughi, Pingki Datta, Seyed~Kourosh Mahjour, and Shirko Faroughi.
\newblock Physics-informed neural networks with periodic activation functions
  for solute transport in heterogeneous porous media.
\newblock \emph{arXiv preprint arXiv:2212.08965}, 2022{\natexlab{b}}.
\newblock \doi{https://doi.org/10.48550/arXiv.2212.08965}.

\bibitem[McClenny and Braga-Neto(2020)]{mcclenny2020self}
L.~McClenny and U.~Braga-Neto.
\newblock Self-adaptive physics-informed neural networks using a soft attention
  mechanism.
\newblock \emph{arXiv preprint arXiv:2009.04544}, 2020.
\newblock \doi{https://doi.org/10.48550/arXiv.2009.04544}.

\bibitem[Yadav et~al.(2022)Yadav, Natarajan, and
  Srinivasan]{yadav2022distributed}
G.~K. Yadav, S.~Natarajan, and B.~Srinivasan.
\newblock Distributed pinn for linear elasticity—a unified approach for
  smooth, singular, compressible and incompressible media.
\newblock \emph{International Journal of Computational Methods}, page 2142008,
  2022.

\bibitem[Bauer et~al.(2021)Bauer, Dueben, Hoefler, Quintino, Schulthess, and
  Wedi]{bauer2021digital}
P.~Bauer, P.~D. Dueben, T.~Hoefler, T.~Quintino, T.~C. Schulthess, and N.~P.
  Wedi.
\newblock The digital revolution of earth-system science.
\newblock \emph{Nature Computational Science}, 1\penalty0 (2):\penalty0
  104--113, 2021.
\newblock \doi{https://doi.org/10.1038/s43588-021-00023-0}.

\bibitem[Chen et~al.(2018)Chen, Rubanova, Bettencourt, and
  Duvenaud]{chen2018neural}
R.~T. Chen, Y.~Rubanova, J.~Bettencourt, and D.~K. Duvenaud.
\newblock Neural ordinary differential equations.
\newblock \emph{Advances in neural information processing systems}, 31, 2018.

\bibitem[Chung et~al.(2016)Chung, Lee, and Park]{chung2016deep}
H.~Chung, S.~J. Lee, and J.~G. Park.
\newblock Deep neural network using trainable activation functions.
\newblock In \emph{2016 International Joint Conference on Neural Networks
  (IJCNN)}, pages 348--352. IEEE, 2016.
\newblock \doi{https://doi.org/10.1109/IJCNN.2016.7727219}.

\bibitem[Mattheakis et~al.(2019)Mattheakis, Protopapas, Sondak, Di, and
  Kaxiras]{mattheakis2019physical}
M.~Mattheakis, P.~Protopapas, D.~Sondak, G.~M. Di, and E.~Kaxiras.
\newblock Physical symmetries embedded in neural networks.
\newblock \emph{arXiv preprint arXiv:1904.08991}, 2019.
\newblock \doi{https://doi.org/10.48550/arXiv.1904.08991}.

\bibitem[Lu et~al.(2021{\natexlab{a}})Lu, Jin, Pang, Zhang, and
  Karniadakis]{lu2021learning}
Lu~Lu, Pengzhan Jin, Guofei Pang, Zhongqiang Zhang, and George~Em Karniadakis.
\newblock Learning nonlinear operators via deeponet based on the universal
  approximation theorem of operators.
\newblock \emph{Nature Machine Intelligence}, 3\penalty0 (3):\penalty0
  218--229, 2021{\natexlab{a}}.
\newblock \doi{https://doi.org/10.1038/s42256-021-00302-5}.

\bibitem[Goswami et~al.(2022{\natexlab{a}})Goswami, Bora, Yu, and
  Karniadakis]{goswami2022physicsReview}
S.~Goswami, A.~Bora, Y.~Yu, and G.~E. Karniadakis.
\newblock Physics-informed neural operators.
\newblock \emph{arXiv preprint arXiv:2207.05748}, 2022{\natexlab{a}}.
\newblock \doi{https://doi.org/10.48550/arXiv.2207.05748}.

\bibitem[LeCun et~al.(2015)LeCun, Bengio, and Hinton]{lecun2015deep}
Y.~LeCun, Y.~Bengio, and G.~Hinton.
\newblock Deep learning.
\newblock \emph{nature}, 521\penalty0 (7553):\penalty0 436--444, 2015.
\newblock \doi{https://www.nature.com/articles/nature14539}.

\bibitem[Creswell et~al.(2018)Creswell, White, Dumoulin, Arulkumaran, Sengupta,
  and Bharath]{creswell2018generative}
A.~Creswell, T.~White, V.~Dumoulin, K.~Arulkumaran, B.~Sengupta, and A.~A.
  Bharath.
\newblock Generative adversarial networks: An overview.
\newblock \emph{IEEE Signal Processing Magazine}, 35\penalty0 (1):\penalty0
  53--65, 2018.
\newblock \doi{https://doi.org/10.1109/MSP.2017.2765202}.

\bibitem[Scarselli et~al.(2008)Scarselli, Gori, Tsoi, Hagenbuchner, and
  Monfardini]{scarselli2008graph}
F.~Scarselli, M.~Gori, A.~C. Tsoi, M.~Hagenbuchner, and G.~Monfardini.
\newblock The graph neural network model.
\newblock \emph{IEEE transactions on neural networks}, 20\penalty0
  (1):\penalty0 61--80, 2008.
\newblock \doi{https://doi.org/10.1109/TNN.2008.2005605}.

\bibitem[Goodfellow et~al.(2020)Goodfellow, Pouget-Abadie, Mirza, Xu,
  Warde-Farley, Ozair, Courville, and Bengio]{goodfellow2020generative}
I.~Goodfellow, J.~Pouget-Abadie, M.~Mirza, B.~Xu, D.~Warde-Farley, S.~Ozair,
  A.~Courville, and Y.~Bengio.
\newblock Generative adversarial networks.
\newblock \emph{Communications of the ACM}, 63\penalty0 (11):\penalty0
  139--144, 2020.
\newblock \doi{https://doi.org/10.1145/3422622}.

\bibitem[Rasamoelina et~al.(2020)Rasamoelina, Adjailia, and
  Sin{\v{c}}{\'a}k]{rasamoelina2020review}
A.~D. Rasamoelina, F.~Adjailia, and P.~Sin{\v{c}}{\'a}k.
\newblock A review of activation function for artificial neural network.
\newblock In \emph{2020 IEEE 18th World Symposium on Applied Machine
  Intelligence and Informatics (SAMI)}, pages 281--286. IEEE, 2020.
\newblock \doi{https://doi.org/10.1109/SAMI48414.2020.9108717}.

\bibitem[He et~al.(2020)He, Ma, and Wang]{he2020extract}
C.~He, M.~Ma, and P.~Wang.
\newblock Extract interpretability-accuracy balanced rules from artificial
  neural networks: A review.
\newblock \emph{Neurocomputing}, 387:\penalty0 346--358, 2020.
\newblock \doi{https://doi.org/10.1016/j.neucom.2020.01.036}.

\bibitem[Li et~al.(2014)Li, Zhang, Chen, and Smola]{li2014efficient}
M.~Li, T.~Zhang, Y.~Chen, and A.~J. Smola.
\newblock Efficient mini-batch training for stochastic optimization.
\newblock In \emph{Proceedings of the 20th ACM SIGKDD international conference
  on Knowledge discovery and data mining}, pages 661--670, 2014.
\newblock \doi{https://doi.org/10.1145/2623330.2623612}.

\bibitem[Huang et~al.(2021)Huang, Krugener, Brown, Menhorn, Bungartz, and
  Hartmann]{Huang2021}
K.~Huang, M.~Krugener, A.~Brown, F.~Menhorn, H.~J. Bungartz, and D.~Hartmann.
\newblock Machine learning-based optimal mesh generation in computational fluid
  dynamics.
\newblock \emph{arXiv preprint: 2102.12923v1}, 2021.
\newblock \doi{https://doi.org/10.48550/arXiv.2102.12923}.

\bibitem[Kumar and Kochmann(2022)]{kumar2022machine}
S.~Kumar and D.~M. Kochmann.
\newblock What machine learning can do for computational solid mechanics.
\newblock In \emph{Current Trends and Open Problems in Computational
  Mechanics}, pages 275--285. Springer, 2022.

\bibitem[Guo et~al.(2021)Guo, Yang, Yu, and Buehler]{guo2021artificial}
K.~Guo, Z.~Yang, C-H. Yu, and M.~J. Buehler.
\newblock Artificial intelligence and machine learning in design of mechanical
  materials.
\newblock \emph{Materials Horizons}, 8\penalty0 (4):\penalty0 1153--1172, 2021.
\newblock \doi{https://doi.org/10.1039/D0MH01451F}.

\bibitem[Zhang et~al.(2020{\natexlab{a}})Zhang, Wang, Jimack, and
  Wang]{Zhang2020}
Z.~Zhang, Y.~Wang, P.~K. Jimack, and H.~Wang.
\newblock Meshing{N}et: a new mesh generation method based on deep learning.
\newblock \emph{arXiv preprint: 2004.07016v1}, 2020{\natexlab{a}}.
\newblock \doi{https://doi.org/10.48550/arXiv.2004.07016}.

\bibitem[Wu et~al.(2022)Wu, Liu, An, Huang, and Lyu]{Tingfan2022}
T.~Wu, X.~Liu, W.~An, Z.~Huang, and H.~Lyu.
\newblock A mesh optimization method using machine learning technique and
  variational mesh adaptation.
\newblock \emph{Chinese Journal of Aeronautics}, 35:\penalty0 27--41, 2022.
\newblock \doi{https://doi.org/10.1016/j.cja.2021.05.018}.

\bibitem[Mendizabal et~al.(2020)Mendizabal, M{\'a}rquez-Neila, and
  Cotin]{mendizabal2020simulation}
A.~Mendizabal, P.~M{\'a}rquez-Neila, and S.~Cotin.
\newblock Simulation of hyperelastic materials in real-time using deep
  learning.
\newblock \emph{Medical image analysis}, 59:\penalty0 101569, 2020.
\newblock \doi{https://doi.org/10.1016/j.media.2019.101569}.

\bibitem[Lu et~al.(2021{\natexlab{b}})Lu, Gao, Dietiker, Shahnam, and
  Rogers]{lu2021machine}
L.~Lu, X.~Gao, J-F. Dietiker, M.~Shahnam, and W.~A. Rogers.
\newblock Machine learning accelerated discrete element modeling of granular
  flows.
\newblock \emph{Chemical Engineering Science}, 245:\penalty0 116832,
  2021{\natexlab{b}}.
\newblock \doi{https://doi.org/10.1016/j.ces.2021.116832}.

\bibitem[Li et~al.(2021{\natexlab{b}})Li, Meidani, Yadav, and
  Farimani]{li2021graph}
Z.~Li, K.~Meidani, P.~Yadav, and A.~B. Farimani.
\newblock Graph neural networks accelerated molecular dynamics.
\newblock \emph{arXiv preprint arXiv:2112.03383}, 2021{\natexlab{b}}.
\newblock \doi{https://doi.org/10.48550/arXiv.2112.03383}.

\bibitem[Menke et~al.(2021)Menke, Maes, and Geiger]{menke2021upscaling}
H.~P. Menke, J.~Maes, and S.~Geiger.
\newblock Upscaling the porosity--permeability relationship of a microporous
  carbonate for darcy-scale flow with machine learning.
\newblock \emph{Scientific Reports}, 11\penalty0 (1):\penalty0 1--10, 2021.
\newblock \doi{https://doi.org/10.1038/s41598-021-82029-2}.

\bibitem[Cheng et~al.(2022)Cheng, Chen, Anastasiou, Angeli, Matar, Guo, Pain,
  and Arcucci]{cheng2022generalised}
S.~Cheng, J.~Chen, C.~Anastasiou, P.~Angeli, O.~K. Matar, Y.~Guo, C.~C. Pain,
  and R.~Arcucci.
\newblock Generalised latent assimilation in heterogeneous reduced spaces with
  machine learning surrogate models.
\newblock \emph{arXiv preprint arXiv:2204.03497}, 2022.
\newblock \doi{https://doi.org/10.48550/arXiv.2204.03497}.

\bibitem[Zawawi et~al.(2018)Zawawi, Saleha, Salwa, Hassan, Zahari, Ramli, and
  Muda]{zawawi2018review}
M.~H. Zawawi, A.~Saleha, A.~Salwa, N.~H. Hassan, N.~M. Zahari, M.~Z. Ramli, and
  Z.~C. Muda.
\newblock A review: Fundamentals of computational fluid dynamics (cfd).
\newblock In \emph{AIP conference proceedings}, volume 2030, page 020252. AIP
  Publishing LLC, 2018.
\newblock \doi{https://doi.org/10.1063/1.5066893}.

\bibitem[He and Tafti(2019)]{he2019supervised}
L.~He and D.~K. Tafti.
\newblock A supervised machine learning approach for predicting variable drag
  forces on spherical particles in suspension.
\newblock \emph{Powder technology}, 345:\penalty0 379--389, 2019.
\newblock \doi{https://doi.org/10.1016/j.powtec.2019.01.013}.

\bibitem[Zhu et~al.(2020)Zhu, Tang, and Luo]{zhu2020machine}
L-T. Zhu, J-X. Tang, and Z-H. Luo.
\newblock Machine learning to assist filtered two-fluid model development for
  dense gas--particle flows.
\newblock \emph{AIChE Journal}, 66\penalty0 (6):\penalty0 e16973, 2020.
\newblock \doi{https://doi.org/10.1002/aic.16973}.

\bibitem[Roriz et~al.(2021)Roriz, Faroughi, McKinley, and
  Fernandes]{roriz2021ml}
A.~I. Roriz, S.~A. Faroughi, G.~H. McKinley, and C.~Fernandes.
\newblock Ml driven models to predict the drag coefficient of a sphere
  translating in shear-thinning viscoelastic fluids.
\newblock 2021.

\bibitem[Loiro et~al.(2021)Loiro, Fernandes, McKinley, and
  Faroughi]{loiro2021digital}
C.~Loiro, C.~Fernandes, G.~H. McKinley, and S.~A. Faroughi.
\newblock Digital-twin for particle-laden viscoelastic fluids: Ml-based models
  to predict the drag coefficient of random arrays of spheres.
\newblock 2021.

\bibitem[Webb et~al.(2020)Webb, Jackson, Gil, and de~Pablo]{webb2020targeted}
M.~A. Webb, N.~E. Jackson, P.~S. Gil, and J.~J. de~Pablo.
\newblock Targeted sequence design within the coarse-grained polymer genome.
\newblock \emph{Science advances}, 6\penalty0 (43):\penalty0 eabc6216, 2020.
\newblock \doi{https://doi.org/10.1126/sciadv.abc6216}.

\bibitem[Srivastava et~al.(2014)Srivastava, Hinton, Krizhevsky, Sutskever, and
  Salakhutdinov]{srivastava2014dropout}
N.~Srivastava, G.~Hinton, A.~Krizhevsky, I.~Sutskever, and R.~Salakhutdinov.
\newblock Dropout: a simple way to prevent neural networks from overfitting.
\newblock \emph{The journal of machine learning research}, 15\penalty0
  (1):\penalty0 1929--1958, 2014.

\bibitem[Bejani and Ghatee(2021)]{bejani2021systematic}
M.~M. Bejani and M.~Ghatee.
\newblock A systematic review on overfitting control in shallow and deep neural
  networks.
\newblock \emph{Artificial Intelligence Review}, 54\penalty0 (8):\penalty0
  6391--6438, 2021.
\newblock \doi{https://doi.org/10.1007/s10462-021-09975-1}.

\bibitem[Ying(2019)]{ying2019overview}
X.~Ying.
\newblock An overview of overfitting and its solutions.
\newblock In \emph{Journal of physics: Conference series}, volume 1168, page
  022022. IOP Publishing, 2019.
\newblock \doi{https://doi.org/10.1088/1742-6596/1168/2/022022}.

\bibitem[Cati et~al.(2022)Cati, aus~der Wiesche, and
  D{\"u}zg{\"u}n]{cati2022numerical}
Y.~Cati, S.~aus~der Wiesche, and M.~D{\"u}zg{\"u}n.
\newblock Numerical model of the railway brake disk for the temperature and
  axial thermal stress analyses.
\newblock \emph{Journal of Thermal Science and Engineering Applications},
  14\penalty0 (10):\penalty0 101014, 2022.
\newblock \doi{https://doi.org/10.1115/1.4054213}.

\bibitem[Chen et~al.(2020)Chen, Liu, Pang, Chen, Chi, and Gong]{Chen2020}
X.~Chen, J.~Liu, Y.~Pang, J.~Chen, L.~Chi, and C.~Gong.
\newblock Developing a new mesh quality evaluation method based on
  convolutional neural network.
\newblock \emph{Engineering Applications of Computational Fluid Mechanics},
  14:\penalty0 391--400, 2020.
\newblock \doi{https://doi.org/10.1080/19942060.2020.1720820}.

\bibitem[Maddu et~al.(2021)Maddu, Sturm, Cheeseman, M{\"u}ller, and
  Sbalzarini]{maddu2021stencil}
S.~Maddu, D.~Sturm, B.~L. Cheeseman, C.~L. M{\"u}ller, and I.~F. Sbalzarini.
\newblock Stencil-net: Data-driven solution-adaptive discretization of partial
  differential equations.
\newblock \emph{arXiv preprint arXiv:2101.06182}, 2021.
\newblock \doi{https://doi.org/10.48550/arXiv.2101.06182}.

\bibitem[Bar-Sinai et~al.(2019)Bar-Sinai, Hoyer, Hickey, and
  Brenner]{bar2019learning}
Y.~Bar-Sinai, S.~Hoyer, J.~Hickey, and M.~P. Brenner.
\newblock Learning data-driven discretizations for partial differential
  equations.
\newblock \emph{Proceedings of the National Academy of Sciences}, 116\penalty0
  (31):\penalty0 15344--15349, 2019.
\newblock \doi{https://doi.org/10.1073/pnas.1814058116}.

\bibitem[Bernardin et~al.(2010)Bernardin, Bossy, Chauvin, Jabir, and
  Rousseau]{bernardin2010stochastic}
F.~Bernardin, M.~Bossy, C.~Chauvin, J-F. Jabir, and A.~Rousseau.
\newblock Stochastic lagrangian method for downscaling problems in
  computational fluid dynamics.
\newblock \emph{ESAIM: Mathematical Modelling and Numerical Analysis},
  44\penalty0 (5):\penalty0 885--920, 2010.
\newblock \doi{https://doi.org/10.1051/m2an/2010046}.

\bibitem[Wei et~al.(2016)Wei, Dong, Chen, Xiao, and Li]{wei2016effect}
X.~Wei, C.~Dong, Z.~Chen, K.~Xiao, and X.~Li.
\newblock The effect of hydrogen on the evolution of intergranular cracking: a
  cross-scale study using first-principles and cohesive finite element methods.
\newblock \emph{RSC advances}, 6\penalty0 (33):\penalty0 27282--27292, 2016.
\newblock \doi{https://doi.org/10.1039/C5RA26061B}.

\bibitem[Shu(1998)]{shu1998essentially}
C.~Shu.
\newblock Essentially non-oscillatory and weighted essentially non-oscillatory
  schemes for hyperbolic conservation laws.
\newblock \emph{Advanced numerical approximation of nonlinear hyperbolic
  equations}, pages 325--432, 1998.

\bibitem[Zhang et~al.(2021{\natexlab{a}})Zhang, Jimack, and
  Wang]{zhang2021meshingnet3d}
Z.~Zhang, P.~K. Jimack, and H.~Wang.
\newblock Meshingnet3d: Efficient generation of adapted tetrahedral meshes for
  computational mechanics.
\newblock \emph{Advances in Engineering Software}, 157:\penalty0 103021,
  2021{\natexlab{a}}.
\newblock \doi{https://doi.org/10.1016/j.advengsoft.2021.103021}.

\bibitem[Triantafyllidis and Labridis(2002)]{triantafyllidis2002finite}
D.~G. Triantafyllidis and D.~P. Labridis.
\newblock A finite-element mesh generator based on growing neural networks.
\newblock \emph{IEEE Transactions on neural networks}, 13\penalty0
  (6):\penalty0 1482--1496, 2002.
\newblock \doi{https://doi.org/10.1109/TNN.2002.804223}.

\bibitem[Srasuay et~al.(2010)Srasuay, Chumthong, and
  Ruangsinchaiwanich]{srasuay2010mesh}
K.~Srasuay, A.~Chumthong, and S.~Ruangsinchaiwanich.
\newblock Mesh generation of fem by ann on iron—core transformer.
\newblock In \emph{2010 International Conference on Electrical Machines and
  Systems}, pages 1885--1890. IEEE, 2010.

\bibitem[Lee and Chen(1993)]{lee1993fluid}
M.~J. Lee and J.~T. Chen.
\newblock Fluid property predictions with the aid of neural networks.
\newblock \emph{Industrial \& engineering chemistry research}, 32\penalty0
  (5):\penalty0 995--997, 1993.

\bibitem[Yang et~al.(2016)Yang, Yang, and Xiao]{Yang2016}
C.~Yang, X.~Yang, and X.~Xiao.
\newblock Data-driven projection method in fluid simulation.
\newblock \emph{Computer Animation and Virtual Worlds}, 27:\penalty0 415--424,
  2016.
\newblock \doi{https://doi.org/10.1002/cav.1695}.

\bibitem[Tompson et~al.(2016)Tompson, Schlachter, Sprechmann, and
  Perlin]{Jonathan2016}
J.~Tompson, K.~Schlachter, P.~Sprechmann, and K.~Perlin.
\newblock Accelerating {E}ulerian fluid simulation with convolutional networks,
  2016.

\bibitem[Jacobs(1980)]{Jacobs1980}
{\relax D.A.H.}~Jacobs.
\newblock Preconditioned conjugate gradient methods for solving systems of
  algebraic equations.
\newblock Technical Report RD/L/N193/80, Central Electricity Research
  Laboratories, 1980.

\bibitem[Chen et~al.(2019{\natexlab{a}})Chen, Viquerat, and Hachem]{chen2019u}
J.~Chen, J.~Viquerat, and E.~Hachem.
\newblock U-net architectures for fast prediction of incompressible laminar
  flows.
\newblock \emph{arXiv preprint arXiv:1910.13532}, 2019{\natexlab{a}}.
\newblock \doi{https://doi.org/10.48550/arXiv.1910.13532}.

\bibitem[Deng et~al.(2019)Deng, He, Liu, and Kim]{Deng2019}
Z.~Deng, C.~He, Y.~Liu, and K.~Kim.
\newblock Super-resolution reconstruction of turbulent velocity fields using a
  generative adversarial network-based artificial intelligence framework.
\newblock \emph{Physics of Fluids}, 31:\penalty0 125111, 2019.
\newblock \doi{https://doi.org/10.1063/1.5127031}.

\bibitem[Ling et~al.(2016)Ling, Kurzawski, and Templeton]{Ling2016}
J.~Ling, A.~Kurzawski, and J.~Templeton.
\newblock Reynolds averaged turbulence modelling using deep neural networks
  with embedded invariance.
\newblock \emph{Journal of Fluid Mechanics}, 807:\penalty0 155--166, 2016.
\newblock \doi{https://doi.org/10.1017/jfm.2016.615}.

\bibitem[L{\'e}vy-Leblond(1971)]{levy1971galilei}
J.~L{\'e}vy-Leblond.
\newblock Galilei group and galilean invariance.
\newblock In \emph{Group theory and its applications}, pages 221--299.
  Elsevier, 1971.
\newblock \doi{https://doi.org/10.1016/B978-0-12-455152-7.50011-2}.

\bibitem[Maulik et~al.(2019)Maulik, San, Rasheed, and Vedula]{Maulik2019}
R.~Maulik, O.~San, A.~Rasheed, and P.~Vedula.
\newblock Subgrid modelling for two-dimensional turbulence using neural
  networks.
\newblock \emph{Journal of Fluid Mechanics}, 858:\penalty0 122--144, 2019.
\newblock \doi{https://doi.org/10.1017/jfm.2018.770}.

\bibitem[Kraichnan(1967)]{Kraichnan1967}
R.~H. Kraichnan.
\newblock Inertial ranges in two-dimensional turbulence.
\newblock \emph{Phys. Fluids}, 10:\penalty0 1417--1423, 1967.
\newblock \doi{https://doi.org/10.1063/1.1762301}.

\bibitem[Kim and Lee(2020)]{Kim2020}
J.~Kim and C.~Lee.
\newblock Prediction of turbulent heat transfer using convolutional neural
  networks.
\newblock \emph{Journal of Fluid Mechanics}, 882:\penalty0 A18, 2020.
\newblock \doi{https://doi.org/10.1017/jfm.2019.814}.

\bibitem[Kingma and Ba(2014)]{kingma2014adam}
D.~P. Kingma and J.~Ba.
\newblock Adam: A method for stochastic optimization.
\newblock \emph{arXiv preprint arXiv:1412.6980}, 2014.
\newblock \doi{https://doi.org/10.48550/arXiv.1412.6980}.

\bibitem[Hoang(2020)]{hoang2020image}
N-D. Hoang.
\newblock Image processing-based spall object detection using gabor filter,
  texture analysis, and adaptive moment estimation (adam) optimized logistic
  regression models.
\newblock \emph{Advances in Civil Engineering}, 2020, 2020.
\newblock \doi{https://doi.org/10.1155/2020/8829715}.

\bibitem[Priyadarshini and Cotton(2021)]{priyadarshini2021novel}
I.~Priyadarshini and C.~Cotton.
\newblock A novel lstm--cnn--grid search-based deep neural network for
  sentiment analysis.
\newblock \emph{The Journal of Supercomputing}, 77\penalty0 (12):\penalty0
  13911--13932, 2021.
\newblock \doi{https://doi.org/10.1007/s11227-021-03838-w}.

\bibitem[Sun et~al.(2021)Sun, Ding, Zhang, and Jia]{sun2021improved}
Y.~Sun, S.~Ding, Z.~Zhang, and W.~Jia.
\newblock An improved grid search algorithm to optimize svr for prediction.
\newblock \emph{Soft Computing}, 25\penalty0 (7):\penalty0 5633--5644, 2021.
\newblock \doi{https://doi.org/10.1007/s00500-020-05560-w}.

\bibitem[Yousif et~al.(2022)Yousif, Yu, and Lim]{Yousif2022}
M.~Yousif, L.~Yu, and H.~Lim.
\newblock Physics-guided deep learning for generating turbulent inflow
  conditions.
\newblock \emph{Journal of Fluid Mechanics}, 936:\penalty0 A21, 2022.
\newblock \doi{https://doi.org/10.1017/jfm.2022.61}.

\bibitem[Shi et~al.(2016)Shi, Caballero, Husz{\'a}r, Totz, Aitken, Bishop,
  Rueckert, and Wang]{shi2016real}
W.~Shi, J.~Caballero, F.~Husz{\'a}r, J.~Totz, A.~P. Aitken, R.~Bishop,
  D.~Rueckert, and Z.~Wang.
\newblock Real-time single image and video super-resolution using an efficient
  sub-pixel convolutional neural network.
\newblock In \emph{Proceedings of the IEEE conference on computer vision and
  pattern recognition}, pages 1874--1883, 2016.

\bibitem[Talab et~al.(2019)Talab, Awang, and Najim]{talab2019super}
M.~A. Talab, S.~Awang, and S.~M. Najim.
\newblock Super-low resolution face recognition using integrated efficient
  sub-pixel convolutional neural network (espcn) and convolutional neural
  network (cnn).
\newblock In \emph{2019 IEEE international conference on automatic control and
  intelligent systems (I2CACIS)}, pages 331--335. IEEE, 2019.
\newblock \doi{https://doi.org/10.1109/I2CACIS.2019.8825083}.

\bibitem[Huang et~al.(2015)Huang, Xu, and Yu]{huang2015bidirectional}
Z.~Huang, W.~Xu, and K.~Yu.
\newblock Bidirectional lstm-crf models for sequence tagging.
\newblock \emph{arXiv preprint arXiv:1508.01991}, 2015.
\newblock \doi{https://doi.org/10.48550/arXiv.1508.01991}.

\bibitem[Sherstinsky(2020)]{sherstinsky2020fundamentals}
A.~Sherstinsky.
\newblock Fundamentals of recurrent neural network (rnn) and long short-term
  memory (lstm) network.
\newblock \emph{Physica D: Nonlinear Phenomena}, 404:\penalty0 132306, 2020.
\newblock \doi{https://doi.org/10.1016/j.physd.2019.132306}.

\bibitem[Kou and Zhang(2021)]{Kou2021}
J.~Kou and W.~Zhang.
\newblock Data-driven modeling for unsteady aerodynamics and aeroelasticity.
\newblock \emph{Progress in Aerospace Sciences}, 125:\penalty0 100725, 2021.
\newblock \doi{https://doi.org/10.1016/j.paerosci.2021.100725}.

\bibitem[Wang et~al.(2022{\natexlab{a}})Wang, Gong, Fan, Li, and
  Qian]{WANG2022302}
Z.~Wang, K.~Gong, W.~Fan, C.~Li, and W.~Qian.
\newblock Prediction of swirling flow field in combustor based on deep
  learning.
\newblock \emph{Acta Astronautica}, 201:\penalty0 302--316, 2022{\natexlab{a}}.
\newblock \doi{https://doi.org/10.1016/j.actaastro.2022.09.022}.

\bibitem[Chowdhary et~al.(2022)Chowdhary, Hoang, Lee, Ray, Weirs, and
  Carnes]{CHOWDHARY2022115396}
K.~Chowdhary, C.~Hoang, K.~Lee, J.~Ray, V.~G. Weirs, and B.~Carnes.
\newblock Calibrating hypersonic turbulence flow models with the {HIFiRE}-1
  experiment using data-driven machine-learned models.
\newblock \emph{Computer Methods in Applied Mechanics and Engineering},
  401:\penalty0 115396, 2022.
\newblock \doi{https://doi.org/10.1016/j.cma.2022.115396}.

\bibitem[Bond and Daniel(2008)]{bond2008guaranteed}
B.~N. Bond and L.~Daniel.
\newblock Guaranteed stable projection-based model reduction for indefinite and
  unstable linear systems.
\newblock In \emph{2008 IEEE/ACM International Conference on Computer-Aided
  Design}, pages 728--735. IEEE, 2008.
\newblock \doi{https://doi.org/10.1109/ICCAD.2008.4681657}.

\bibitem[Beli et~al.(2018)Beli, Mencik, Silva, and Arruda]{beli2018projection}
D.~Beli, J-M. Mencik, P.~B. Silva, and J.~Arruda.
\newblock A projection-based model reduction strategy for the wave and
  vibration analysis of rotating periodic structures.
\newblock \emph{Computational Mechanics}, 62\penalty0 (6):\penalty0 1511--1528,
  2018.
\newblock \doi{https://doi.org/10.1007/s00466-018-1576-7}.

\bibitem[Siddiqui et~al.(2022)Siddiqui, De~Troyer, Decuyper, Csurcsia,
  Schoukens, and Runacres]{SIDDIQUI2022103706}
M.~F. Siddiqui, T.~De~Troyer, J.~Decuyper, P.~Z. Csurcsia, J.~Schoukens, and
  M.C. Runacres.
\newblock A data-driven nonlinear state-space model of the unsteady lift force
  on a pitching wing.
\newblock \emph{Journal of Fluids and Structures}, 114:\penalty0 103706, 2022.
\newblock \doi{https://doi.org/10.1016/j.jfluidstructs.2022.103706}.

\bibitem[Wang et~al.(2022{\natexlab{b}})Wang, Kou, and Zhang]{Wang2022}
X.~Wang, J.~Kou, and W.~Zhang.
\newblock Unsteady aerodynamic prediction for iced airfoil based on multi-task
  learning.
\newblock \emph{Physics of Fluids}, 34:\penalty0 087117, 2022{\natexlab{b}}.
\newblock \doi{https://doi.org/10.1063/5.0101991}.

\bibitem[Stevens and Colonius(2020)]{Stevens2020}
B.~Stevens and T.~Colonius.
\newblock Enhancement of shock-capturing methods via machine learning.
\newblock \emph{Theoretical and Computational Fluid Dynamics}, 34:\penalty0
  483--496, 2020.
\newblock \doi{https://doi.org/10.1007/s00162-020-00531-1}.

\bibitem[Pawar and Faroughi(2022)]{pawar2022complex}
N.~Pawar and S.~A. Faroughi.
\newblock Complex fluids latent space exploration towards accelerated
  predictive modeling.
\newblock \emph{Bulletin of the American Physical Society}, 2022.

\bibitem[Fernandes et~al.(2022{\natexlab{a}})Fernandes, Faroughi, Ferr{\'a}s,
  and Afonso]{fernandes2022advanced}
C.~Fernandes, S.~A. Faroughi, L.~L. Ferr{\'a}s, and A.~M. Afonso.
\newblock Advanced polymer simulation and processing, 2022{\natexlab{a}}.

\bibitem[Faroughi et~al.(2020)Faroughi, Fernandes, N{\'o}brega, and
  McKinley]{faroughi2020closure}
S.~A. Faroughi, C.~Fernandes, J.~M. N{\'o}brega, and G.~H. McKinley.
\newblock A closure model for the drag coefficient of a sphere translating in a
  viscoelastic fluid.
\newblock \emph{Journal of Non-Newtonian Fluid Mechanics}, 277:\penalty0
  104218, 2020.
\newblock \doi{https://doi.org/10.1016/j.jnnfm.2019.104218}.

\bibitem[Fernandes et~al.(2019)Fernandes, Faroughi, Carneiro, N\'obrega, and
  McKinley]{Fernandes2019}
C.~Fernandes, S.~A. Faroughi, O.~S. Carneiro, J.~Miguel N\'obrega, and G.~H.
  McKinley.
\newblock Fully-resolved simulations of particle-laden viscoelastic fluids
  using an immersed boundary method.
\newblock \emph{Journal of Non-Newtonian Fluid Mechanics}, 266:\penalty0
  80--94, 2019.
\newblock \doi{https://doi.org/10.1016/j.jnnfm.2019.02.007}.

\bibitem[Lin et~al.(2017)Lin, Wu, Lin, Wen, and Li]{lin2017ensemble}
W.~Lin, Z.~Wu, L.~Lin, A.~Wen, and J.~Li.
\newblock An ensemble random forest algorithm for insurance big data analysis.
\newblock \emph{Ieee access}, 5:\penalty0 16568--16575, 2017.
\newblock \doi{https://doi.org/10.1109/ACCESS.2017.2738069}.

\bibitem[Chen et~al.(2015)Chen, He, Benesty, Khotilovich, Tang, Cho, Chen,
  et~al.]{chen2015xgboost}
T.~Chen, T.~He, M.~Benesty, V.~Khotilovich, Y.~Tang, H.~Cho, K.~Chen, et~al.
\newblock Xgboost: extreme gradient boosting.
\newblock \emph{R package version 0.4-2}, 1\penalty0 (4):\penalty0 1--4, 2015.

\bibitem[Lennon et~al.(2022)Lennon, McKinley, and Swan]{lennon2022MLcomplex}
K.~R. Lennon, G.~H. McKinley, and J.~W. Swan.
\newblock Scientific machine learning for modeling and simulating complex
  fluids.
\newblock \emph{arXiv preprint arXiv:2210.04431v1}, 2022.
\newblock \doi{https://doi.org/10.48550/arXiv.2210.04431}.

\bibitem[Cai et~al.(2022{\natexlab{b}})Cai, Chen, and Liu]{Cai2022}
Z.~Cai, J.~Chen, and M.~Liu.
\newblock Least-squares {R}e{LU} neural network ({LSNN}) method for scalar
  nonlinear hyperbolic conservation law.
\newblock \emph{Applied Numerical Mathematics}, 174:\penalty0 163--176,
  2022{\natexlab{b}}.
\newblock \doi{https://doi.org/10.1016/j.apnum.2022.01.002}.

\bibitem[Haber et~al.(2022)Haber, Viquerat, Larcher, Ryckelynck, Alves, Patil,
  and Hachem]{Haber2022}
G.~E. Haber, J.~Viquerat, A.~Larcher, D.~Ryckelynck, J.~Alves, A.~Patil, and
  E.~Hachem.
\newblock Deep learning model to assist multiphysics conjugate problems.
\newblock \emph{Physics of Fluids}, 34:\penalty0 015131, 2022.
\newblock \doi{https://doi.org/10.1063/5.0077723}.

\bibitem[Lara and Ferrer(2022)]{Lara2022}
F.~M. Lara and E.~Ferrer.
\newblock Accelerating high order discontinuous {G}alerkin solvers using neural
  networks: 1{D} {B}urgers’ equation.
\newblock \emph{Computers \& Fluids}, 235:\penalty0 105274, 2022.
\newblock \doi{https://doi.org/10.1016/j.compfluid.2021.105274}.

\bibitem[List et~al.(2022)List, Chen, and Thuerey]{List2022}
B.~List, L.~Chen, and N.~Thuerey.
\newblock Learned turbulence modelling with differentiable fluid solvers:
  Physics-based loss functions and optimisation horizons.
\newblock \emph{Journal of Fluid Mechanics}, 949:\penalty0 A25, 2022.
\newblock \doi{https://doi.org/10.1017/jfm.2022.738}.

\bibitem[Vollant et~al.(2017)Vollant, Balarac, and Corre]{Vollant2017}
A.~Vollant, G.~Balarac, and C.~Corre.
\newblock Subgrid-scale scalar flux modelling based on optimal estimation
  theory and machine-learning procedures.
\newblock \emph{Journal of Turbulence}, 18:\penalty0 854--878, 2017.
\newblock \doi{https://doi.org/10.1080/14685248.2017.1334907}.

\bibitem[Beck et~al.(2018)Beck, Flad, and Munz]{Beck2018}
A.~D. Beck, D.~G. Flad, and C.~D. Munz.
\newblock Deep neural networks for data-driven turbulence models.
\newblock \emph{arXiv preprint arXiv:1806.04482}, 2018.
\newblock \doi{https://doi.org/10.48550/arXiv.1806.04482}.

\bibitem[Sekar et~al.(2019)Sekar, Jiang, Shu, and Khoo]{Sekar2019}
V.~Sekar, Q.~H. Jiang, C.~Shu, and B.~C. Khoo.
\newblock Fast flow field prediction over airfoils using deep learning
  approach.
\newblock \emph{Physics of Fluids}, 31:\penalty0 057103, 2019.
\newblock \doi{https://doi.org/10.1063/1.5094943}.

\bibitem[Zhu et~al.(2019{\natexlab{a}})Zhu, Zhang, Kou, and Liu]{Zhu2019}
L.~Zhu, W.~W. Zhang, J.~Kou, and Y.~Liu.
\newblock Machine learning methods for turbulence modeling in subsonic flows
  around airfoils.
\newblock \emph{Physics of Fluids}, 31:\penalty0 015105, 2019{\natexlab{a}}.
\newblock \doi{https://doi.org/10.1063/1.5061693}.

\bibitem[Tadesse et~al.(2012)Tadesse, Patel, Chaudhary, and
  Nagpal]{tadesse2012neural}
Z.~Tadesse, K.~Patel, S.~Chaudhary, and A.~Nagpal.
\newblock Neural networks for prediction of deflection in composite bridges.
\newblock \emph{Journal of Constructional Steel Research}, 68\penalty0
  (1):\penalty0 138--149, 2012.
\newblock \doi{https://doi.org/10.1016/j.jcsr.2011.08.003}.

\bibitem[G{\"u}neyisi et~al.(2014)G{\"u}neyisi, D'Aniello, Landolfo,
  Mermerda{\c{s}}, et~al.]{guneyisi2014prediction}
E.~M. G{\"u}neyisi, M.~D'Aniello, R.~Landolfo, K.~Mermerda{\c{s}}, et~al.
\newblock Prediction of the flexural overstrength factor for steel beams using
  artificial neural network.
\newblock \emph{Steel and Composite Structures}, 17\penalty0 (3):\penalty0
  215--236, 2014.
\newblock \doi{http://dx.doi.org/10.12989/scs.2014.17.3.215}.

\bibitem[Hung et~al.(2019)Hung, Viet, and Van]{hung2019deep}
T.~V. Hung, V.~Q. Viet, and T.~D. Van.
\newblock A deep learning-based procedure for estimation of ultimate load
  carrying of steel trusses using advanced analysis.
\newblock \emph{Journal of Science and Technology in Civil Engineering
  (STCE)-HUCE}, 13\penalty0 (3):\penalty0 113--123, 2019.
\newblock \doi{https://doi.org/10.31814/stce.nuce2019-13(3)-11}.

\bibitem[Chen et~al.(2019{\natexlab{b}})Chen, Li, Chen, Ren, Wang, and
  Li]{chen2019applicationshell}
G.~Chen, T.~Li, Q.~Chen, S.~Ren, C.~Wang, and S.~Li.
\newblock Application of deep learning neural network to identify collision
  load conditions based on permanent plastic deformation of shell structures.
\newblock \emph{Computational Mechanics}, 64\penalty0 (2):\penalty0 435--449,
  2019{\natexlab{b}}.
\newblock \doi{https://doi.org/10.1007/s00466-019-01706-2}.

\bibitem[Hosseinpour et~al.(2020)Hosseinpour, Sharifi, and
  Sharifi]{hosseinpour2020neural}
M.~Hosseinpour, Y.~Sharifi, and H.~Sharifi.
\newblock Neural network application for distortional buckling capacity
  assessment of castellated steel beams.
\newblock In \emph{Structures}, volume~27, pages 1174--1183. Elsevier, 2020.
\newblock \doi{https://doi.org/10.1016/j.istruc.2020.07.027}.

\bibitem[Trahair and Bradford(2017)]{trahair2017behaviour}
N.~S. Trahair and M.~A. Bradford.
\newblock \emph{The behaviour and design of steel structures to AS 4100}.
\newblock CRC Press, 2017.

\bibitem[White et~al.(2006)White, Surovek, Alemdar, Chang, Kim, and
  Kuchenbecker]{white2006stability}
D.~W. White, A.~E. Surovek, B.~N. Alemdar, C-J. Chang, Y.~D. Kim, and G.~H.
  Kuchenbecker.
\newblock Stability analysis and design of steel building frames using the 2005
  aisc specification.
\newblock \emph{Steel Structures}, 6\penalty0 (2):\penalty0 71--91, 2006.

\bibitem[Eur(2005)]{Eurocode3}
Design of steel structures part 1–1: General rules and rules for buildings.
\newblock \emph{European Committee for Standardization (ECS), Brussels
  Belgium.}, 2005.

\bibitem[White et~al.(2019)White, Arrighi, Kudo, and
  Watts]{white2019multiscale}
D.~A. White, W.~J. Arrighi, J.~Kudo, and S.~E. Watts.
\newblock Multiscale topology optimization using neural network surrogate
  models.
\newblock \emph{Computer Methods in Applied Mechanics and Engineering},
  346:\penalty0 1118--1135, 2019.
\newblock \doi{https://doi.org/10.1016/j.cma.2018.09.007}.

\bibitem[Zhang et~al.(2019)Zhang, Li, Xiao, Gao, Chu, and
  Zhang]{zhang2019concurrent}
Y.~Zhang, H.~Li, M.~Xiao, L.~Gao, S.~Chu, and J.~Zhang.
\newblock Concurrent topology optimization for cellular structures with
  nonuniform microstructures based on the kriging metamodel.
\newblock \emph{Structural and Multidisciplinary Optimization}, 59\penalty0
  (4):\penalty0 1273--1299, 2019.
\newblock \doi{https://doi.org/10.1007/s00158-018-2130-0}.

\bibitem[Sigmund and Maute(2013)]{sigmund2013topology}
O.~Sigmund and K.~Maute.
\newblock Topology optimization approaches.
\newblock \emph{Structural and Multidisciplinary Optimization}, 48\penalty0
  (6):\penalty0 1031--1055, 2013.
\newblock \doi{https://doi.org/10.1007/s00158-013-0978-6}.

\bibitem[Abueidda et~al.(2020)Abueidda, Koric, and Sobh]{abueidda2020topology}
D.~W. Abueidda, S.~Koric, and N.~A. Sobh.
\newblock Topology optimization of 2d structures with nonlinearities using deep
  learning.
\newblock \emph{Computers \& Structures}, 237:\penalty0 106283, 2020.
\newblock \doi{https://doi.org/10.1016/j.compstruc.2020.106283}.

\bibitem[Yu et~al.(2019)Yu, Hur, Jung, and Jang]{yu2019deep}
Y.~Yu, T.~Hur, J.~Jung, and I.~G. Jang.
\newblock Deep learning for determining a near-optimal topological design
  without any iteration.
\newblock \emph{Structural and Multidisciplinary Optimization}, 59\penalty0
  (3):\penalty0 787--799, 2019.
\newblock \doi{https://doi.org/10.1007/s00158-018-2101-5}.

\bibitem[Banga et~al.(2018)Banga, Gehani, Bhilare, Patel, and
  Kara]{banga20183d}
S.~Banga, H.~Gehani, S.~Bhilare, S.~Patel, and L.~Kara.
\newblock 3d topology optimization using convolutional neural networks.
\newblock \emph{arXiv preprint arXiv:1808.07440}, 2018.
\newblock \doi{https://doi.org/10.48550/arXiv.1808.07440}.

\bibitem[Li et~al.(2019)Li, Huang, Li, Zheng, and Hong]{li2019non}
B.~Li, C.~Huang, X.~Li, S.~Zheng, and J.~Hong.
\newblock Non-iterative structural topology optimization using deep learning.
\newblock \emph{Computer-Aided Design}, 115:\penalty0 172--180, 2019.
\newblock \doi{https://doi.org/10.1016/j.cad.2019.05.038}.

\bibitem[Takano and Alaghband(2019)]{takano2019srgan}
N.~Takano and G.~Alaghband.
\newblock Srgan: Training dataset matters.
\newblock \emph{arXiv preprint arXiv:1903.09922}, 2019.
\newblock \doi{https://doi.org/10.48550/arXiv.1903.09922}.

\bibitem[Nagano and Kikuta(2018)]{nagano2018srgan}
Y.~Nagano and Y.~Kikuta.
\newblock Srgan for super-resolving low-resolution food images.
\newblock In \emph{Proceedings of the Joint Workshop on Multimedia for Cooking
  and Eating Activities and Multimedia Assisted Dietary Management}, pages
  33--37, 2018.
\newblock \doi{https://doi.org/10.1145/3230519.3230587}.

\bibitem[Kumar et~al.(2020)Kumar, Tan, Zheng, and Kochmann]{kumar2020inverse}
S.~Kumar, S.~Tan, L.~Zheng, and D.~M. Kochmann.
\newblock Inverse-designed spinodoid metamaterials.
\newblock \emph{npj Computational Materials}, 6\penalty0 (1):\penalty0 1--10,
  2020.
\newblock \doi{https://doi.org/10.1038/s41524-020-0341-6}.

\bibitem[Ni and Gao(2021)]{ni2021deep}
B.~Ni and H.~Gao.
\newblock A deep learning approach to the inverse problem of modulus
  identification in elasticity.
\newblock \emph{MRS Bulletin}, 46\penalty0 (1):\penalty0 19--25, 2021.
\newblock \doi{https://doi.org/10.1557/s43577-020-00006-y}.

\bibitem[Messner(2020)]{messner2020convolutional}
M.~C. Messner.
\newblock Convolutional neural network surrogate models for the mechanical
  properties of periodic structures.
\newblock \emph{Journal of Mechanical Design}, 142\penalty0 (2):\penalty0
  024503, 2020.
\newblock \doi{https://doi.org/10.1115/1.4045040}.

\bibitem[Tcherniak(2002)]{tcherniak2002topology}
D.~Tcherniak.
\newblock Topology optimization of resonating structures using simp method.
\newblock \emph{International Journal for Numerical Methods in Engineering},
  54\penalty0 (11):\penalty0 1605--1622, 2002.
\newblock \doi{https://doi.org/10.1002/nme.484}.

\bibitem[Lininger et~al.(2021)Lininger, Hinczewski, and
  Strangi]{lininger2021general}
A.~Lininger, M.~Hinczewski, and G.~Strangi.
\newblock General inverse design of thin-film metamaterials with convolutional
  neural networks.
\newblock \emph{arXiv preprint arXiv:2104.01952}, 2021.
\newblock \doi{https://doi.org/10.48550/arXiv.2104.01952}.

\bibitem[L{\"o}per et~al.(2014)L{\"o}per, Stuckelberger, Niesen, Werner,
  Filipi{\v{c}}, Moon, Yum, Topi{\v{c}}, De~Wolf, and Ballif]{loper2014complex}
P.~L{\"o}per, M.~Stuckelberger, B.~Niesen, J.~Werner, M.~Filipi{\v{c}},
  S.~Moon, J-H. Yum, M.~Topi{\v{c}}, S.~De~Wolf, and C.~Ballif.
\newblock Complex refractive index spectra of ch3nh3pbi3 perovskite thin films
  determined by spectroscopic ellipsometry and spectrophotometry.
\newblock \emph{The journal of physical chemistry letters}, 6\penalty0
  (1):\penalty0 66--71, 2014.
\newblock \doi{https://doi.org/10.1021/jz502471h}.

\bibitem[Smith and Smith(2020)]{smith2020conditional}
K.~E. Smith and A~O Smith.
\newblock Conditional gan for timeseries generation.
\newblock \emph{arXiv preprint arXiv:2006.16477}, 2020.
\newblock \doi{https://doi.org/10.48550/arXiv.2006.16477}.

\bibitem[Balaji et~al.(2019)Balaji, Min, Bai, Chellappa, and
  Graf]{balaji2019conditional}
Y.~Balaji, M.~R. Min, B.~Bai, R.~Chellappa, and H.~P. Graf.
\newblock Conditional gan with discriminative filter generation for
  text-to-video synthesis.
\newblock In \emph{IJCAI}, volume~1, page~2, 2019.

\bibitem[Oberai et~al.(2003)Oberai, Gokhale, and
  Feij{\'o}o]{oberai2003solution}
A.~A. Oberai, N.~H. Gokhale, and G.~R. Feij{\'o}o.
\newblock Solution of inverse problems in elasticity imaging using the adjoint
  method.
\newblock \emph{Inverse problems}, 19\penalty0 (2):\penalty0 297, 2003.

\bibitem[Liang et~al.(2018)Liang, Liu, Martin, and Sun]{liang2018deep}
L.~Liang, M.~Liu, C.~Martin, and W.~Sun.
\newblock A deep learning approach to estimate stress distribution: a fast and
  accurate surrogate of finite-element analysis.
\newblock \emph{Journal of The Royal Society Interface}, 15\penalty0
  (138):\penalty0 20170844, 2018.
\newblock \doi{https://doi.org/10.1098/rsif.2017.0844}.

\bibitem[Mozaffar et~al.(2019)Mozaffar, Bostanabad, Chen, Ehmann, Cao, and
  Bessa]{Mozaffar2019}
M.~Mozaffar, R.~Bostanabad, W.~Chen, K.~Ehmann, J.~Cao, and M.~A. Bessa.
\newblock Deep learning predicts path-dependent plasticity.
\newblock \emph{PNAS}, 116:\penalty0 26414--26420, 2019.
\newblock \doi{https://doi.org/10.1073/pnas.1911815116}.

\bibitem[Chatterjee(2000)]{chatterjee2000introduction}
A.~Chatterjee.
\newblock An introduction to the proper orthogonal decomposition.
\newblock \emph{Current science}, pages 808--817, 2000.
\newblock \doi{https://www.jstor.org/stable/24103957}.

\bibitem[Liang et~al.(2002)Liang, Lee, Lim, Lin, Lee, and Wu]{liang2002proper}
Y.~C. Liang, H.~P. Lee, S.~P. Lim, W.~Z. Lin, K.~H. Lee, and C.~Wu.
\newblock Proper orthogonal decomposition and its applications—part i:
  Theory.
\newblock \emph{Journal of Sound and vibration}, 252\penalty0 (3):\penalty0
  527--544, 2002.
\newblock \doi{https://doi.org/10.1006/jsvi.2001.4041}.

\bibitem[Long et~al.(2021)Long, Zhao, Jiang, Li, and Liu]{long2021deep}
X.~Y. Long, S.~K. Zhao, C.~Jiang, W.~P. Li, and C.~H. Liu.
\newblock Deep learning-based planar crack damage evaluation using
  convolutional neural networks.
\newblock \emph{Engineering Fracture Mechanics}, 246:\penalty0 107604, 2021.
\newblock \doi{https://doi.org/10.1016/j.engfracmech.2021.107604}.

\bibitem[Zhu et~al.(2021)Zhu, Ohsaki, and Guo]{zhu2021prediction}
S.~Zhu, M.~Ohsaki, and X.~Guo.
\newblock Prediction of non-linear buckling load of imperfect reticulated shell
  using modified consistent imperfection and machine learning.
\newblock \emph{Engineering Structures}, 226:\penalty0 111374, 2021.
\newblock \doi{https://doi.org/10.1016/j.engstruct.2020.111374}.

\bibitem[Miller and Ziemia{\'n}ski(2020)]{miller2020optimization}
B.~Miller and L.~Ziemia{\'n}ski.
\newblock Optimization of dynamic behavior of thin-walled laminated cylindrical
  shells by genetic algorithms and deep neural networks supported by modal
  shape identification.
\newblock \emph{Advances in Engineering Software}, 147:\penalty0 102830, 2020.
\newblock \doi{https://doi.org/10.1016/j.advengsoft.2020.102830}.

\bibitem[Nie et~al.(2020)Nie, Jiang, and Kara]{nie2020stress}
Z.~Nie, H.~Jiang, and L.~B. Kara.
\newblock Stress field prediction in cantilevered structures using
  convolutional neural networks.
\newblock \emph{Journal of Computing and Information Science in Engineering},
  20\penalty0 (1):\penalty0 011002, 2020.
\newblock \doi{https://doi.org/10.1115/1.4044097}.

\bibitem[Pizarro and Massone(2021)]{pizarro2021structural}
P.~N. Pizarro and L.~M. Massone.
\newblock Structural design of reinforced concrete buildings based on deep
  neural networks.
\newblock \emph{Engineering Structures}, 241:\penalty0 112377, 2021.
\newblock \doi{https://doi.org/10.1016/j.engstruct.2021.112377}.

\bibitem[Pathirage et~al.(2018)Pathirage, Li, Li, Hao, Liu, and
  Ni]{pathirage2018structural}
C.~S. Pathirage, J.~Li, L.~Li, H.~Hao, W.~Liu, and P.~Ni.
\newblock Structural damage identification based on autoencoder neural networks
  and deep learning.
\newblock \emph{Engineering structures}, 172:\penalty0 13--28, 2018.
\newblock \doi{https://doi.org/10.1016/j.engstruct.2018.05.109}.

\bibitem[Jiang and Zhang(2020)]{jiang2020real}
S.~Jiang and J.~Zhang.
\newblock Real-time crack assessment using deep neural networks with
  wall-climbing unmanned aerial system.
\newblock \emph{Computer-Aided Civil and Infrastructure Engineering},
  35\penalty0 (6):\penalty0 549--564, 2020.
\newblock \doi{https://doi.org/10.1111/mice.12519}.

\bibitem[Perez-Ramirez et~al.(2019)Perez-Ramirez, Amezquita-Sanchez,
  Valtierra-Rodriguez, Adeli, Dominguez-Gonzalez, and
  Romero-Troncoso]{perez2019recurrent}
C.~A. Perez-Ramirez, J.~P. Amezquita-Sanchez, M.~Valtierra-Rodriguez, H.~Adeli,
  A.~Dominguez-Gonzalez, and R.~J. Romero-Troncoso.
\newblock Recurrent neural network model with bayesian training and mutual
  information for response prediction of large buildings.
\newblock \emph{Engineering Structures}, 178:\penalty0 603--615, 2019.
\newblock \doi{https://doi.org/10.1016/j.engstruct.2018.10.065}.

\bibitem[Truong et~al.(2020)Truong, Dinh-Cong, Lee, and
  Nguyen-Thoi]{truong2020effective}
T.~T. Truong, D.~Dinh-Cong, J.~Lee, and T.~Nguyen-Thoi.
\newblock An effective deep feedforward neural networks (dfnn) method for
  damage identification of truss structures using noisy incomplete modal data.
\newblock \emph{Journal of Building Engineering}, 30:\penalty0 101244, 2020.
\newblock \doi{https://doi.org/10.1016/j.jobe.2020.101244}.

\bibitem[Raissi(2018)]{raissi2018deephidden}
M.~Raissi.
\newblock Deep hidden physics models: Deep learning of nonlinear partial
  differential equations.
\newblock \emph{The Journal of Machine Learning Research}, 19\penalty0
  (1):\penalty0 932--955, 2018.

\bibitem[Biros et~al.(2011)Biros, Ghattas, Heinkenschloss, Keyes, Mallick,
  Tenorio, van Bloemen~Waanders, Willcox, Marzouk, and Biegler]{biros2011large}
G.~Biros, O.~Ghattas, M.~Heinkenschloss, D.~Keyes, B.~Mallick, L.~Tenorio,
  B.~van Bloemen~Waanders, K.~Willcox, Y.~Marzouk, and L.~Biegler.
\newblock \emph{Large-scale inverse problems and quantification of
  uncertainty}.
\newblock John Wiley \& Sons, 2011.

\bibitem[Vogel(2002)]{vogel2002computational}
C.~R. Vogel.
\newblock \emph{Computational methods for inverse problems}.
\newblock SIAM, 2002.

\bibitem[Franssen et~al.(2009)Franssen, Hendricks, Riva, Bakr, Van~der Wiel,
  Stauffer, and Guadagnini]{franssen2009comparison}
H.~J. Franssen, A.~A. Hendricks, M.~Riva, M.~Bakr, N.~Van~der Wiel,
  F.~Stauffer, and A.~Guadagnini.
\newblock A comparison of seven methods for the inverse modelling of
  groundwater flow. application to the characterisation of well catchments.
\newblock \emph{Advances in Water Resources}, 32\penalty0 (6):\penalty0
  851--872, 2009.
\newblock \doi{https://doi.org/10.1016/j.advwatres.2009.02.011}.

\bibitem[Li et~al.(2020{\natexlab{a}})Li, Kovachki, Azizzadenesheli, Liu,
  Bhattacharya, Stuart, and Anandkumar]{li2020fourier}
Z.~Li, N.~Kovachki, K.~Azizzadenesheli, B.~Liu, K.~Bhattacharya, A.~Stuart, and
  A.~Anandkumar.
\newblock Fourier neural operator for parametric partial differential
  equations.
\newblock \emph{arXiv preprint arXiv:2010.08895}, 2020{\natexlab{a}}.
\newblock \doi{https://doi.org/10.48550/arXiv.2010.08895}.

\bibitem[Randjbaran et~al.(2015)Randjbaran, Zahari, Vaghei, and
  Karamizadeh]{randjbaran2015review}
E.~Randjbaran, R.~Zahari, R.~Vaghei, and F.~Karamizadeh.
\newblock A review paper on comparison of numerical techniques for finding
  approximate solutions to boundary value problems on post-buckling in
  functionally graded materials.
\newblock \emph{Trends Journal of Sciences Research}, 2\penalty0 (1):\penalty0
  1--6, 2015.
\newblock \doi{https://www.tjsr.org/journal/index.php/tjsr/article/view/9}.

\bibitem[Triebel(2015)]{triebel2015hybrid}
H.~Triebel.
\newblock \emph{Hybrid function spaces, heat and Navier-Stokes equations}.
\newblock 2015.

\bibitem[Durran(2013)]{durran2013numerical}
D.~R. Durran.
\newblock \emph{Numerical methods for wave equations in geophysical fluid
  dynamics}, volume~32.
\newblock Springer Science \& Business Media, 2013.

\bibitem[Prato(2004)]{prato2004stochastic}
G.~D. Prato.
\newblock The stochastic burgers equation.
\newblock In \emph{Kolmogorov Equations for Stochastic PDEs}, pages 131--153.
  Springer, 2004.

\bibitem[Medkov{\'a}(2018)]{medkova2018laplace}
D.~Medkov{\'a}.
\newblock The laplace equation.
\newblock \emph{Boundary value problems on bounded and unbounded Lipschitz
  domains. Springer, Cham}, 2018.

\bibitem[Genovese et~al.(2006)Genovese, Deutsch, Neelov, Goedecker, and
  Beylkin]{genovese2006efficient}
L.~Genovese, T.~Deutsch, A.~Neelov, S.~Goedecker, and G.~Beylkin.
\newblock Efficient solution of poisson’s equation with free boundary
  conditions.
\newblock \emph{The Journal of chemical physics}, 125\penalty0 (7):\penalty0
  074105, 2006.
\newblock \doi{https://doi.org/10.1063/1.2335442}.

\bibitem[Jagtap et~al.(2020)Jagtap, Kharazmi, and
  Karniadakis]{jagtap2020conservative}
A.~D. Jagtap, E.~Kharazmi, and G.~E. Karniadakis.
\newblock Conservative physics-informed neural networks on discrete domains for
  conservation laws: Applications to forward and inverse problems.
\newblock \emph{Computer Methods in Applied Mechanics and Engineering},
  365:\penalty0 113028, 2020.
\newblock \doi{https://doi.org/10.1016/j.cma.2020.113028}.

\bibitem[Jagtap and Karniadakis(2021)]{jagtap2021extended}
A.~D. Jagtap and G.~E. Karniadakis.
\newblock Extended physics-informed neural networks (xpinns): A generalized
  space-time domain decomposition based deep learning framework for nonlinear
  partial differential equations.
\newblock In \emph{AAAI Spring Symposium: MLPS}, 2021.

\bibitem[Mahmoudabadbozchelou and
  Jamali(2021)]{mahmoudabadbozchelou2021rheology}
M.~Mahmoudabadbozchelou and S.~Jamali.
\newblock Rheology-informed neural networks (rhinns) for forward and inverse
  metamodelling of complex fluids.
\newblock \emph{Scientific reports}, 11\penalty0 (1):\penalty0 1--13, 2021.
\newblock \doi{https://doi.org/10.1038/s41598-021-91518-3}.

\bibitem[Katsikis et~al.(2022)Katsikis, Muradova, and
  Stavroulakis]{katsikis2022gentle}
D.~Katsikis, A.~D. Muradova, and G.~E. Stavroulakis.
\newblock A gentle introduction to physics informed neural networks, with
  applications in static rod and beam problems.
\newblock \emph{J Adv App Comput Math}, 9:\penalty0 103--128, 2022.
\newblock \doi{https://doi.org/10.15377/2409-5761.2022.09.8}.

\bibitem[Salvati et~al.(2022)Salvati, Tognan, Laurenti, Pelegatti, and
  De~Bona]{salvati2022defect}
E.~Salvati, A.~Tognan, L.~Laurenti, M.~Pelegatti, and F.~De~Bona.
\newblock A defect-based physics-informed machine learning framework for
  fatigue finite life prediction in additive manufacturing.
\newblock \emph{Materials \& Design}, page 111089, 2022.
\newblock \doi{https://doi.org/10.1016/j.matdes.2022.111089}.

\bibitem[Datta et~al.(2022)Datta, Pawar, and Faroughi]{datta2022physics}
P.~Datta, N.~Pawar, and S.~A. Faroughi.
\newblock A physics-informed neural network to model the flow of dry particles.
\newblock In \emph{Fall Meeting 2022}. AGU, 2022.

\bibitem[Nguyen et~al.(2022)Nguyen, Raissi, and Seshaiyer]{nguyen2022modeling}
L.~Nguyen, M.~Raissi, and P.~Seshaiyer.
\newblock Modeling, analysis and physics informed neural network approaches for
  studying the dynamics of covid-19 involving human-human and human-pathogen
  interaction.
\newblock \emph{Computational and Mathematical Biophysics}, 10\penalty0
  (1):\penalty0 1--17, 2022.
\newblock \doi{https://doi.org/10.1515/cmb-2022-0001}.

\bibitem[Shaier et~al.(2022)Shaier, Raissi, and Seshaiyer]{shaier2022data}
S.~Shaier, M.~Raissi, and P.~Seshaiyer.
\newblock Data-driven approaches for predicting spread of infectious diseases
  through dinns: Disease informed neural networks.
\newblock \emph{Letters in Biomathematics}, 9\penalty0 (1):\penalty0 71--105,
  2022.

\bibitem[Dissanayake and Phan-Thien(1994)]{dissanayake1994neural}
M.~Dissanayake and N.~Phan-Thien.
\newblock Neural-network-based approximations for solving partial differential
  equations.
\newblock \emph{communications in Numerical Methods in Engineering},
  10\penalty0 (3):\penalty0 195--201, 1994.
\newblock \doi{https://doi.org/10.1002/cnm.1640100303}.

\bibitem[Owhadi(2015)]{owhadi2015bayesian}
H.~Owhadi.
\newblock Bayesian numerical homogenization.
\newblock \emph{Multiscale Modeling \& Simulation}, 13\penalty0 (3):\penalty0
  812--828, 2015.
\newblock \doi{https://doi.org/10.1137/140974596}.

\bibitem[Raissi et~al.(2018)Raissi, Yazdani, and Karniadakis]{raissi2018hidden}
M.~Raissi, A.~Yazdani, and G.~E. Karniadakis.
\newblock Hidden fluid mechanics: A navier-stokes informed deep learning
  framework for assimilating flow visualization data.
\newblock \emph{arXiv preprint arXiv:1808.04327}, 2018.
\newblock \doi{https://doi.org/10.48550/arXiv.1808.04327}.

\bibitem[Iserles(2009)]{iserles2009first}
A.~Iserles.
\newblock \emph{A first course in the numerical analysis of differential
  equations}.
\newblock Number~44. Cambridge university press, 2009.

\bibitem[Nabian et~al.(2021)Nabian, Gladstone, and
  Meidani]{nabian2021efficient}
M.~A. Nabian, R.~J. Gladstone, and H.~Meidani.
\newblock Efficient training of physics-informed neural networks via importance
  sampling.
\newblock \emph{Computer-Aided Civil and Infrastructure Engineering},
  36\penalty0 (8):\penalty0 962--977, 2021.
\newblock \doi{https://doi.org/10.1111/mice.12685}.

\bibitem[Su and Gardner(1969)]{su1969korteweg}
C.~H. Su and C.~S. Gardner.
\newblock Korteweg-de vries equation and generalizations. iii. derivation of
  the korteweg-de vries equation and burgers equation.
\newblock \emph{Journal of Mathematical Physics}, 10\penalty0 (3):\penalty0
  536--539, 1969.
\newblock \doi{https://doi.org/10.1063/1.1664873}.

\bibitem[Constantin and Foias(2020)]{constantin2020navier}
P.~Constantin and C.~Foias.
\newblock \emph{Navier-stokes equations}.
\newblock University of Chicago Press, 2020.

\bibitem[Stiasny et~al.(2021)Stiasny, Misyris, and
  Chatzivasileiadis]{stiasny2021physics}
J.~Stiasny, G.~S. Misyris, and S.~Chatzivasileiadis.
\newblock Physics-informed neural networks for non-linear system identification
  for power system dynamics.
\newblock In \emph{2021 IEEE Madrid PowerTech}, pages 1--6. IEEE, 2021.
\newblock \doi{https://doi.org/10.1109/PowerTech46648.2021.9495063}.

\bibitem[Mao et~al.(2020)Mao, Jagtap, and Karniadakis]{mao2020physics}
Z.~Mao, A.~D. Jagtap, and G.E. Karniadakis.
\newblock Physics-informed neural networks for high-speed flows.
\newblock \emph{Computer Methods in Applied Mechanics and Engineering},
  360:\penalty0 112789, 2020.

\bibitem[Jagtap et~al.(2022{\natexlab{a}})Jagtap, Mao, Adams, and
  Karniadakis]{jagtap2022physics}
A.~D. Jagtap, Z.~Mao, N.~Adams, and G.~E. Karniadakis.
\newblock Physics-informed neural networks for inverse problems in supersonic
  flows.
\newblock \emph{arXiv preprint arXiv:2202.11821}, 2022{\natexlab{a}}.
\newblock \doi{https://doi.org/10.48550/arXiv.2202.11821}.

\bibitem[Zhang et~al.(2020{\natexlab{b}})Zhang, Dey, Kakkar, Dasgupta, and
  Chakraborty]{zhang2020frequency}
T.~Zhang, B.~Dey, P.~Kakkar, A~Dasgupta, and A.~Chakraborty.
\newblock Frequency-compensated pinns for fluid-dynamic design problems.
\newblock \emph{arXiv preprint arXiv:2011.01456}, 2020{\natexlab{b}}.
\newblock \doi{https://doi.org/10.48550/arXiv.2011.01456}.

\bibitem[Tancik et~al.(2020)Tancik, Srinivasan, Mildenhall, Fridovich-Keil,
  Raghavan, Singhal, Ramamoorthi, Barron, and Ng]{tancik2020fourier}
M.~Tancik, P.~Srinivasan, B.~Mildenhall, S.~Fridovich-Keil, N.~Raghavan,
  U.~Singhal, R.~Ramamoorthi, J.~Barron, and R.~Ng.
\newblock Fourier features let networks learn high frequency functions in low
  dimensional domains.
\newblock \emph{Advances in Neural Information Processing Systems},
  33:\penalty0 7537--7547, 2020.

\bibitem[Cheng and Zhang(2021)]{cheng2021deep}
C.~Cheng and G-T. Zhang.
\newblock Deep learning method based on physics informed neural network with
  resnet block for solving fluid flow problems.
\newblock \emph{Water}, 13\penalty0 (4):\penalty0 423, 2021.
\newblock \doi{https://doi.org/10.3390/w13040423}.

\bibitem[Lou et~al.(2021)Lou, Meng, and Karniadakis]{lou2021physics}
Q.~Lou, X.~Meng, and G.~E. Karniadakis.
\newblock Physics-informed neural networks for solving forward and inverse flow
  problems via the boltzmann-bgk formulation.
\newblock \emph{Journal of Computational Physics}, 447:\penalty0 110676, 2021.
\newblock \doi{https://doi.org/10.1016/j.jcp.2021.110676}.

\bibitem[Wessels et~al.(2020)Wessels, Wei{\ss}enfels, and
  Wriggers]{wessels2020neural}
H.~Wessels, C.~Wei{\ss}enfels, and P.~Wriggers.
\newblock The neural particle method--an updated lagrangian physics informed
  neural network for computational fluid dynamics.
\newblock \emph{Computer Methods in Applied Mechanics and Engineering},
  368:\penalty0 113127, 2020.
\newblock \doi{https://doi.org/10.1016/j.cma.2020.113127}.

\bibitem[Mahmoudabadbozchelou et~al.(2022)Mahmoudabadbozchelou, Karniadakis,
  and Jamali]{mahmoudabadbozchelou2022nn}
M.~Mahmoudabadbozchelou, G.~E. Karniadakis, and S.~Jamali.
\newblock nn-pinns: Non-newtonian physics-informed neural networks for complex
  fluid modeling.
\newblock \emph{Soft Matter}, 18\penalty0 (1):\penalty0 172--185, 2022.
\newblock \doi{https://doi.org/10.1039/D1SM01298C}.

\bibitem[Haghighat et~al.(2021{\natexlab{b}})Haghighat, Amini, and
  Juanes]{haghighat2021physicsb}
E.~Haghighat, D.~Amini, and R.~Juanes.
\newblock Physics-informed neural network simulation of multiphase
  poroelasticity using stress-split sequential training.
\newblock \emph{arXiv preprint arXiv:2110.03049}, 2021{\natexlab{b}}.
\newblock \doi{https://doi.org/10.48550/arXiv.2110.03049}.

\bibitem[Almajid and Abu-Al-Saud(2022)]{almajid2022prediction}
M.~M. Almajid and M.~O. Abu-Al-Saud.
\newblock Prediction of porous media fluid flow using physics informed neural
  networks.
\newblock \emph{Journal of Petroleum Science and Engineering}, 208:\penalty0
  109205, 2022.
\newblock \doi{https://doi.org/10.1016/j.petrol.2021.109205}.

\bibitem[Depina et~al.(2022)Depina, Jain, Mar, and
  Gotovac]{depina2022application}
I.~Depina, S.~Jain, V.~S. Mar, and H.~Gotovac.
\newblock Application of physics-informed neural networks to inverse problems
  in unsaturated groundwater flow.
\newblock \emph{Georisk: Assessment and Management of Risk for Engineered
  Systems and Geohazards}, 16\penalty0 (1):\penalty0 21--36, 2022.
\newblock \doi{https://doi.org/10.1080/17499518.2021.1971251}.

\bibitem[Tartakovsky et~al.(2018)Tartakovsky, Marrero, Perdikaris, Tartakovsky,
  and Barajas-S]{tartakovsky2018learning}
A.~M. Tartakovsky, C.~O. Marrero, P.~Perdikaris, G.~D. Tartakovsky, and
  D.~Barajas-S.
\newblock Learning parameters and constitutive relationships with physics
  informed deep neural networks.
\newblock \emph{arXiv preprint arXiv:1808.03398}, 2018.
\newblock \doi{https://doi.org/10.48550/arXiv.1808.03398}.

\bibitem[Thakur et~al.(2022)Thakur, Raissi, and
  Ardekani]{thakur2022viscoelasticnet}
S.~Thakur, M.~Raissi, and A.~M. Ardekani.
\newblock Viscoelasticnet: A physics informed neural network framework for
  stress discovery and model selection.
\newblock \emph{arXiv preprint arXiv:2209.06972}, 2022.
\newblock \doi{https://doi.org/10.48550/arXiv.2209.06972}.

\bibitem[Fernandes et~al.(2022{\natexlab{b}})Fernandes, Faroughi, Ribeiro,
  Isabel, and McKinley]{fernandes2022finite}
C.~Fernandes, S.~A. Faroughi, R.~Ribeiro, A.~Isabel, and G.~H. McKinley.
\newblock Finite volume simulations of particle-laden viscoelastic fluid flows:
  Application to hydraulic fracture processes.
\newblock \emph{Engineering with Computers}, pages 1--27, 2022{\natexlab{b}}.
\newblock \doi{https://doi.org/10.1007/s00366-022-01626-5}.

\bibitem[Chiu et~al.(2022)Chiu, Wong, Ooi, Dao, and Ong]{chiu2022can}
P.~Chiu, J.~C. Wong, C.~Ooi, M.~H. Dao, and Y.~Ong.
\newblock Can-pinn: A fast physics-informed neural network based on
  coupled-automatic--numerical differentiation method.
\newblock \emph{Computer Methods in Applied Mechanics and Engineering},
  395:\penalty0 114909, 2022.
\newblock \doi{https://doi.org/10.1016/j.cma.2022.114909}.

\bibitem[Van Der~Hoeven(2004)]{van2004truncated}
J.~Van Der~Hoeven.
\newblock The truncated fourier transform and applications.
\newblock In \emph{Proceedings of the 2004 international symposium on Symbolic
  and algebraic computation}, pages 290--296, 2004.
\newblock \doi{https://doi.org/10.1145/1005285.1005327}.

\bibitem[Raynaud et~al.(2022)Raynaud, Houde, and
  Gosselin]{raynaud2022modalpinn}
G.~Raynaud, S.~Houde, and F.~P. Gosselin.
\newblock Modalpinn: an extension of physics-informed neural networks with
  enforced truncated fourier decomposition for periodic flow reconstruction
  using a limited number of imperfect sensors.
\newblock \emph{Journal of Computational Physics}, page 111271, 2022.
\newblock \doi{https://doi.org/10.1016/j.jcp.2022.111271}.

\bibitem[Oldenburg et~al.(2022)Oldenburg, Borowski, {\"O}ner, Schmitz, and
  Stiehm]{oldenburg2022geometry}
J.~Oldenburg, F.~Borowski, A.~{\"O}ner, K.~Schmitz, and M.~Stiehm.
\newblock Geometry aware physics informed neural network surrogate for solving
  navier-stokes equation (gapinn).
\newblock 2022.
\newblock \doi{https://doi.org/10.21203/rs.3.rs-1466550/v1}.

\bibitem[Wandel et~al.(2022)Wandel, Weinmann, Neidlin, and
  Klein]{wandel2022spline}
N.~Wandel, M.~Weinmann, M.~Neidlin, and R.~Klein.
\newblock Spline-pinn: Approaching pdes without data using fast,
  physics-informed hermite-spline cnns.
\newblock In \emph{Proceedings of the AAAI Conference on Artificial
  Intelligence}, volume~36, pages 8529--8538, 2022.
\newblock \doi{https://doi.org/10.1609/aaai.v36i8.20830}.

\bibitem[Jin et~al.(2021)Jin, Cai, Li, and Karniadakis]{jin2021nsfnets}
X.~Jin, S.~Cai, H.~Li, and G.~E. Karniadakis.
\newblock Nsfnets (navier-stokes flow nets): Physics-informed neural networks
  for the incompressible navier-stokes equations.
\newblock \emph{Journal of Computational Physics}, 426:\penalty0 109951, 2021.
\newblock \doi{https://doi.org/10.1016/j.jcp.2020.109951}.

\bibitem[Cheng et~al.(2021)Cheng, Xu, Li, and Zhang]{cheng2021deep1}
C.~Cheng, P.~Xu, Y.~Li, and G.~Zhang.
\newblock Deep learning based on pinn for solving 2 d0f vortex induced
  vibration of cylinder with high reynolds number.
\newblock \emph{arXiv preprint arXiv:2106.01545}, 2021.
\newblock \doi{https://doi.org/10.48550/arXiv.2106.01545}.

\bibitem[Eivazi et~al.(2022)Eivazi, Tahani, Schlatter, and
  Vinuesa]{eivazi2022physics}
H.~Eivazi, M.~Tahani, P.~Schlatter, and R.~Vinuesa.
\newblock Physics-informed neural networks for solving reynolds-averaged
  navier--stokes equations.
\newblock \emph{Physics of Fluids}, 34\penalty0 (7):\penalty0 075117, 2022.
\newblock \doi{https://doi.org/10.1063/5.0095270}.

\bibitem[Wang et~al.(2017)Wang, Wu, and Xiao]{wang2017physics}
J.~Wang, J.~Wu, and H.~Xiao.
\newblock Physics-informed machine learning approach for reconstructing
  reynolds stress modeling discrepancies based on dns data.
\newblock \emph{Physical Review Fluids}, 2\penalty0 (3):\penalty0 034603, 2017.
\newblock \doi{https://doi.org/10.1103/PhysRevFluids.2.034603}.

\bibitem[Zhang et~al.(2022{\natexlab{a}})Zhang, Zhu, Wang, Ju, Qian, Ye, and
  Yang]{zhang2022gw}
X.~Zhang, Y.~Zhu, J.~Wang, L.~Ju, Y.~Qian, M.~Ye, and J.~Yang.
\newblock Gw-pinn: A deep learning algorithm for solving groundwater flow
  equations.
\newblock \emph{Advances in Water Resources}, page 104243, 2022{\natexlab{a}}.
\newblock \doi{https://doi.org/10.1016/j.advwatres.2022.104243}.

\bibitem[Aliakbari et~al.(2022)Aliakbari, Mahmoudi, Vadasz, and
  Arzani]{aliakbari2022predicting}
M.~Aliakbari, M.~Mahmoudi, P.~Vadasz, and A.~Arzani.
\newblock Predicting high-fidelity multiphysics data from low-fidelity fluid
  flow and transport solvers using physics-informed neural networks.
\newblock \emph{International Journal of Heat and Fluid Flow}, 96:\penalty0
  109002, 2022.
\newblock \doi{https://doi.org/10.1016/j.ijheatfluidflow.2022.109002}.

\bibitem[Tartakovsky et~al.(2020)Tartakovsky, Marrero, Perdikaris, Tartakovsky,
  and Barajas-Solano]{tartakovsky2020physics}
A.~M. Tartakovsky, C.~O. Marrero, P.~Perdikaris, G.~D. Tartakovsky, and
  D.~Barajas-Solano.
\newblock Physics-informed deep neural networks for learning parameters and
  constitutive relationships in subsurface flow problems.
\newblock \emph{Water Resources Research}, 56\penalty0 (5):\penalty0
  e2019WR026731, 2020.
\newblock \doi{https://doi.org/10.1029/2019WR026731}.

\bibitem[Kashefi and Mukerji(2022{\natexlab{a}})]{kashefi2022prediction}
Ali Kashefi and Tapan Mukerji.
\newblock Prediction of fluid flow in porous media by sparse observations and
  physics-informed pointnet.
\newblock \emph{arXiv preprint arXiv:2208.13434}, 2022{\natexlab{a}}.
\newblock \doi{https://doi.org/10.48550/arXiv.2208.13434}.

\bibitem[Kissas et~al.(2020)Kissas, Yang, Hwuang, Witschey, Detre, and
  Perdikaris]{kissas2020machine}
G.~Kissas, Y.~Yang, E.~Hwuang, W.~R. Witschey, J.~A. Detre, and P.~Perdikaris.
\newblock Machine learning in cardiovascular flows modeling: Predicting
  arterial blood pressure from non-invasive 4d flow mri data using
  physics-informed neural networks.
\newblock \emph{Computer Methods in Applied Mechanics and Engineering},
  358:\penalty0 112623, 2020.
\newblock \doi{https://doi.org/10.1016/j.cma.2019.112623}.

\bibitem[Arzani et~al.(2021)Arzani, Wang, and D'Souza]{arzani2021uncovering}
A.~Arzani, J-X. Wang, and R.~M. D'Souza.
\newblock Uncovering near-wall blood flow from sparse data with
  physics-informed neural networks.
\newblock \emph{Physics of Fluids}, 33\penalty0 (7):\penalty0 071905, 2021.
\newblock \doi{https://doi.org/10.1063/5.0055600}.

\bibitem[Jagtap et~al.(2022{\natexlab{b}})Jagtap, Mitsotakis, and
  Karniadakis]{jagtap2022deep}
A.~D. Jagtap, D.~Mitsotakis, and G.~E. Karniadakis.
\newblock Deep learning of inverse water waves problems using multi-fidelity
  data: Application to {S}erre--{G}reen--{N}aghdi equations.
\newblock \emph{Ocean Engineering}, 248:\penalty0 110775, 2022{\natexlab{b}}.
\newblock \doi{https://doi.org/10.1016/j.oceaneng.2022.110775}.

\bibitem[Kashefi and Mukerji(2022{\natexlab{b}})]{KASHEFI2022111510}
Ali Kashefi and Tapan Mukerji.
\newblock Physics-informed pointnet: A deep learning solver for steady-state
  incompressible flows and thermal fields on multiple sets of irregular
  geometries.
\newblock \emph{Journal of Computational Physics}, 468:\penalty0 111510,
  2022{\natexlab{b}}.
\newblock ISSN 0021-9991.
\newblock \doi{https://doi.org/10.1016/j.jcp.2022.111510}.

\bibitem[Haghighat et~al.(2020)Haghighat, Raissi, Moure, Gomez, and
  Juanes]{haghighat2020deep}
E.~Haghighat, M.~Raissi, A.~Moure, H.~Gomez, and R.~Juanes.
\newblock A deep learning framework for solution and discovery in solid
  mechanics: linear elasticity.
\newblock \emph{arXiv preprint arXiv:2003.02751}, 2020.

\bibitem[Shukla et~al.(2021)Shukla, Jagtap, Blackshire, Sparkman, and
  Karniadakis]{shukla2021physics}
K.~Shukla, A.~D. Jagtap, J.~L. Blackshire, D.~Sparkman, and G.~E. Karniadakis.
\newblock A physics-informed neural network for quantifying the microstructural
  properties of polycrystalline nickel using ultrasound data: A promising
  approach for solving inverse problems.
\newblock \emph{IEEE Signal Processing Magazine}, 39\penalty0 (1):\penalty0
  68--77, 2021.
\newblock \doi{https://doi.org/10.1109/MSP.2021.3118904}.

\bibitem[Henkes et~al.(2022)Henkes, Wessels, and Mahnken]{henkes2022physics}
A.~Henkes, H.~Wessels, and R.~Mahnken.
\newblock Physics informed neural networks for continuum micromechanics.
\newblock \emph{Computer Methods in Applied Mechanics and Engineering},
  393:\penalty0 114790, 2022.
\newblock \doi{https://doi.org/10.1016/j.cma.2022.114790}.

\bibitem[Zhang and Gu(2021)]{zhang2021physics}
Z.~Zhang and G.~X. Gu.
\newblock Physics-informed deep learning for digital materials.
\newblock \emph{Theoretical and Applied Mechanics Letters}, 11\penalty0
  (1):\penalty0 100220, 2021.
\newblock \doi{https://doi.org/10.1016/j.taml.2021.100220}.

\bibitem[Rao et~al.(2020{\natexlab{b}})Rao, Sun, and Liu]{rao2020physics}
C.~Rao, H.~Sun, and Y.~Liu.
\newblock Physics informed deep learning for computational elastodynamics
  without labeled data.
\newblock \emph{arXiv preprint arXiv:2006.08472}, 2020{\natexlab{b}}.
\newblock \doi{https://doi.org/10.48550/arXiv.2006.08472}.

\bibitem[Fang and Zhan(2019)]{fang2019deep}
Z.~Fang and J.~Zhan.
\newblock Deep physical informed neural networks for metamaterial design.
\newblock \emph{IEEE Access}, 8:\penalty0 24506--24513, 2019.
\newblock \doi{https://doi.org/10.1109/ACCESS.2019.2963375}.

\bibitem[Lax and Nelson(1976)]{lax1976maxwell}
M.~Lax and D.~F. Nelson.
\newblock Maxwell equations in material form.
\newblock \emph{Physical Review B}, 13\penalty0 (4):\penalty0 1777, 1976.
\newblock \doi{https://doi.org/10.1103/PhysRevB.13.1777}.

\bibitem[Zhang et~al.(2020{\natexlab{c}})Zhang, Yin, and
  Karniadakis]{zhang2020physics}
E.~Zhang, M.~Yin, and G.~E. Karniadakis.
\newblock Physics-informed neural networks for nonhomogeneous material
  identification in elasticity imaging.
\newblock \emph{arXiv preprint arXiv:2009.04525}, 2020{\natexlab{c}}.
\newblock \doi{https://doi.org/10.48550/arXiv.2009.04525}.

\bibitem[Abueidda et~al.(2022{\natexlab{a}})Abueidda, Koric, Guleryuz, and
  Sobh]{abueidda2022enhanced}
D.~W. Abueidda, S.~Koric, E.~Guleryuz, and N.~A. Sobh.
\newblock Enhanced physics-informed neural networks for hyperelasticity.
\newblock \emph{arXiv preprint arXiv:2205.14148}, 2022{\natexlab{a}}.
\newblock \doi{https://doi.org/10.48550/arXiv.2205.14148}.

\bibitem[Abueidda et~al.(2022{\natexlab{b}})Abueidda, Koric, Al-Rub, Parrott,
  James, and Sobh]{abueidda2022deep}
D.~W. Abueidda, S.~Koric, R.~A. Al-Rub, C.~M. Parrott, K.~A. James, and N.~A.
  Sobh.
\newblock A deep learning energy method for hyperelasticity and
  viscoelasticity.
\newblock \emph{European Journal of Mechanics-A/Solids}, 95:\penalty0 104639,
  2022{\natexlab{b}}.
\newblock \doi{https://doi.org/10.1016/j.euromechsol.2022.104639}.

\bibitem[Yuan et~al.(2022)Yuan, Ni, Deng, and Hao]{yuan2022pinn}
L.~Yuan, Y.~Ni, X.~Deng, and S.~Hao.
\newblock A-pinn: Auxiliary physics informed neural networks for forward and
  inverse problems of nonlinear integro-differential equations.
\newblock \emph{Journal of Computational Physics}, 462:\penalty0 111260, 2022.
\newblock \doi{https://doi.org/10.1016/j.jcp.2022.111260}.

\bibitem[Lu et~al.(2021{\natexlab{c}})Lu, Meng, Mao, and
  Karniadakis]{lu2021deepxde}
L.~Lu, X.~Meng, Z.~Mao, and G.~E. Karniadakis.
\newblock Deepxde: A deep learning library for solving differential equations.
\newblock \emph{SIAM Review}, 63\penalty0 (1):\penalty0 208--228,
  2021{\natexlab{c}}.
\newblock \doi{https://doi.org/10.1137/19M1274067}.

\bibitem[Arora(2022)]{arora2022physrnet}
R.~Arora.
\newblock Physrnet: Physics informed super-resolution network for application
  in computational solid mechanics.
\newblock \emph{arXiv preprint arXiv:2206.15457}, 2022.
\newblock \doi{https://doi.org/10.48550/arXiv.2206.15457}.

\bibitem[Madenci et~al.(2019)Madenci, Barut, and
  Dorduncu]{madenci2019peridynamic}
E.~Madenci, A.~Barut, and M.~Dorduncu.
\newblock \emph{Peridynamic differential operator for numerical analysis},
  volume~10.
\newblock Springer, 2019.

\bibitem[Haghighat et~al.(2021{\natexlab{c}})Haghighat, Bekar, Madenci, and
  Juanes]{haghighat2021nonlocal}
E.~Haghighat, A.~C. Bekar, E.~Madenci, and R.~Juanes.
\newblock A nonlocal physics-informed deep learning framework using the
  peridynamic differential operator.
\newblock \emph{Computer Methods in Applied Mechanics and Engineering},
  385:\penalty0 114012, 2021{\natexlab{c}}.
\newblock \doi{https://doi.org/10.1016/j.cma.2021.114012}.

\bibitem[Huang et~al.(2006)Huang, Zhu, and Siew]{huang2006extreme}
G.~Huang, Q.~Zhu, and C.~Siew.
\newblock Extreme learning machine: theory and applications.
\newblock \emph{Neurocomputing}, 70\penalty0 (1-3):\penalty0 489--501, 2006.
\newblock \doi{https://doi.org/10.1016/j.neucom.2005.12.126}.

\bibitem[Yan et~al.(2022)Yan, Vescovini, and Dozio]{yan2022framework}
C.~A. Yan, R.~Vescovini, and L.~Dozio.
\newblock A framework based on physics-informed neural networks and extreme
  learning for the analysis of composite structures.
\newblock \emph{Computers \& Structures}, 265:\penalty0 106761, 2022.
\newblock \doi{https://doi.org/10.1016/j.compstruc.2022.106761}.

\bibitem[Rezaei et~al.(2022)Rezaei, Harandi, Moeineddin, Xu, and
  Reese]{rezaei2022mixed}
S.~Rezaei, A.~Harandi, A.~Moeineddin, B.~Xu, and S.~Reese.
\newblock A mixed formulation for physics-informed neural networks as a
  potential solver for engineering problems in heterogeneous domains:
  comparison with finite element method.
\newblock \emph{arXiv preprint arXiv:2206.13103}, 2022.

\bibitem[Mallampati and Almekkawy(2021)]{mallampati2021measuring}
A.~Mallampati and M.~Almekkawy.
\newblock Measuring tissue elastic properties using physics based neural
  networks.
\newblock In \emph{2021 IEEE UFFC Latin America Ultrasonics Symposium (LAUS)},
  pages 1--4. IEEE, 2021.
\newblock \doi{https://doi.org/10.1109/LAUS53676.2021.9639231}.

\bibitem[Li et~al.(2021{\natexlab{c}})Li, Bazant, and
  Zhu]{li2021physicselastic}
W.~Li, M.~Z. Bazant, and J.~Zhu.
\newblock A physics-guided neural network framework for elastic plates:
  Comparison of governing equations-based and energy-based approaches.
\newblock \emph{Computer Methods in Applied Mechanics and Engineering},
  383:\penalty0 113933, 2021{\natexlab{c}}.
\newblock \doi{https://doi.org/10.1016/j.cma.2021.113933}.

\bibitem[Vahab et~al.(2021)Vahab, Haghighat, Khaleghi, and
  Khalili]{vahab2021physics}
M.~Vahab, E.~Haghighat, M.~Khaleghi, and N.~Khalili.
\newblock A physics informed neural network approach to solution and
  identification of biharmonic equations of elasticity.
\newblock \emph{arXiv preprint arXiv:2108.07243}, 2021.
\newblock \doi{https://doi.org/10.48550/arXiv.2108.07243}.

\bibitem[Raj et~al.(2021)Raj, Kumbhar, and Annabattula]{raj2021physics}
M.~Raj, P.~Kumbhar, and R.~K. Annabattula.
\newblock Physics-informed neural networks for solving thermo-mechanics
  problems of functionally graded material.
\newblock \emph{arXiv preprint arXiv:2111.10751}, 2021.
\newblock \doi{https://doi.org/10.48550/arXiv.2111.10751}.

\bibitem[Zhang et~al.(2022{\natexlab{b}})Zhang, Dao, Karniadakis, and
  Suresh]{zhang2022analyses}
E.~Zhang, M.~Dao, G.~E. Karniadakis, and S.~Suresh.
\newblock Analyses of internal structures and defects in materials using
  physics-informed neural networks.
\newblock \emph{Science advances}, 8\penalty0 (7):\penalty0 eabk0644,
  2022{\natexlab{b}}.
\newblock \doi{https://doi.org/10.1126/sciadv.abk0644}.

\bibitem[Bastek and Kochmann(2022)]{bastek2022physics}
J.~Bastek and D.~M. Kochmann.
\newblock Physics-informed neural networks for shell structures.
\newblock \emph{arXiv preprint arXiv:2207.14291}, 2022.
\newblock \doi{https://doi.org/10.48550/arXiv.2207.14291}.

\bibitem[Zhang et~al.(2021{\natexlab{b}})Zhang, Gong, and
  Xuan]{zhang2021physicsstain}
X.~Zhang, J.~Gong, and F~Xuan.
\newblock A physics-informed neural network for creep-fatigue life prediction
  of components at elevated temperatures.
\newblock \emph{Engineering Fracture Mechanics}, 258:\penalty0 108130,
  2021{\natexlab{b}}.
\newblock \doi{https://doi.org/10.1016/j.engfracmech.2021.108130}.

\bibitem[Haghighat et~al.(2021{\natexlab{d}})Haghighat, Bekar, Madenci, and
  Juanes]{haghighat2021deepstructuralmech}
E.~Haghighat, A.~C. Bekar, E.~Madenci, and R.~Juanes.
\newblock Deep learning for solution and inversion of structural mechanics and
  vibrations.
\newblock \emph{arXiv preprint arXiv:2105.09477}, 2021{\natexlab{d}}.
\newblock \doi{https://doi.org/10.48550/arXiv.2105.09477}.

\bibitem[Zheng et~al.(2022)Zheng, Li, Qi, Gao, Liu, and Yuan]{zheng2022physics}
B.~Zheng, T.~Li, H.~Qi, L.~Gao, X.~Liu, and L.~Yuan.
\newblock Physics-informed machine learning model for computational fracture of
  quasi-brittle materials without labelled data.
\newblock \emph{International Journal of Mechanical Sciences}, 223:\penalty0
  107282, 2022.
\newblock \doi{https://doi.org/10.1016/j.ijmecsci.2022.107282}.

\bibitem[Arora et~al.(2022)Arora, Kakkar, Dey, and
  Chakraborty]{arora2022physics}
R.~Arora, P.~Kakkar, B.~Dey, and A.~Chakraborty.
\newblock Physics-informed neural networks for modeling rate-and
  temperature-dependent plasticity.
\newblock \emph{arXiv preprint arXiv:2201.08363}, 2022.
\newblock \doi{https://doi.org/10.48550/arXiv.2201.08363}.

\bibitem[Bai et~al.(2022)Bai, Jeong, Batuwatta, Xiao, Wang, Rathnayaka,
  Alzubaidi, Liu, and Gu]{bai2022introduction}
J.~Bai, H.~Jeong, C.~P. Batuwatta, S.~Xiao, Q.~Wang, C.~M. Rathnayaka,
  L.~Alzubaidi, G.~Liu, and Y.~Gu.
\newblock An introduction to programming physics-informed neural network-based
  computational solid mechanics.
\newblock \emph{arXiv preprint arXiv:2210.09060}, 2022.
\newblock \doi{https://doi.org/10.48550/arXiv.2210.09060}.

\bibitem[Dwivedi and Srinivasan(2020)]{dwivedi2020solution}
V.~Dwivedi and B.~Srinivasan.
\newblock Solution of biharmonic equation in complicated geometries with
  physics informed extreme learning machine.
\newblock \emph{Journal of Computing and Information Science in Engineering},
  20\penalty0 (6), 2020.
\newblock \doi{https://doi.org/10.1115/1.4046892}.

\bibitem[Dwivedi et~al.(2021)Dwivedi, Parashar, and
  Srinivasan]{dwivedi2021distributed}
V.~Dwivedi, N.~Parashar, and B.~Srinivasan.
\newblock Distributed learning machines for solving forward and inverse
  problems in partial differential equations.
\newblock \emph{Neurocomputing}, 420:\penalty0 299--316, 2021.
\newblock \doi{https://doi.org/10.1016/j.neucom.2020.09.006}.

\bibitem[Shin et~al.(2020)Shin, Darbon, and Karniadakis]{shin2020convergence}
Yeonjong Shin, Jerome Darbon, and George~Em Karniadakis.
\newblock On the convergence of physics informed neural networks for linear
  second-order elliptic and parabolic type pdes.
\newblock \emph{arXiv preprint arXiv:2004.01806}, 2020.
\newblock \doi{https://doi.org/10.48550/arXiv.2004.01806}.

\bibitem[Fuks and Tchelepi(2020)]{fuks2020limitations}
O.~Fuks and H.~A. Tchelepi.
\newblock Limitations of physics informed machine learning for nonlinear
  two-phase transport in porous media.
\newblock \emph{Journal of Machine Learning for Modeling and Computing},
  1\penalty0 (1), 2020.
\newblock \doi{https://doi.org/10.1615/JMachLearnModelComput.2020033905}.

\bibitem[Long et~al.(2018)Long, She, and Mukhopadhyay]{long2018hybridnet}
Y.~Long, X.~She, and S.~Mukhopadhyay.
\newblock Hybridnet: integrating model-based and data-driven learning to
  predict evolution of dynamical systems.
\newblock In \emph{Conference on Robot Learning}, pages 551--560. PMLR, 2018.

\bibitem[Zhu et~al.(2019{\natexlab{b}})Zhu, Zabaras, Koutsourelakis, and
  Perdikaris]{zhu2019physics}
Y.~Zhu, N.~Zabaras, P.~Koutsourelakis, and P.~Perdikaris.
\newblock Physics-constrained deep learning for high-dimensional surrogate
  modeling and uncertainty quantification without labeled data.
\newblock \emph{Journal of Computational Physics}, 394:\penalty0 56--81,
  2019{\natexlab{b}}.
\newblock \doi{https://doi.org/10.1016/j.jcp.2019.05.024}.

\bibitem[Geneva and Zabaras(2020)]{geneva2020modeling}
N.~Geneva and N.~Zabaras.
\newblock Modeling the dynamics of pde systems with physics-constrained deep
  auto-regressive networks.
\newblock \emph{Journal of Computational Physics}, 403:\penalty0 109056, 2020.
\newblock \doi{https://doi.org/10.1016/j.jcp.2019.109056}.

\bibitem[Wang et~al.(2020)Wang, Kashinath, Mustafa, Albert, , and
  Yu]{wang2020towards}
R.~Wang, K.~Kashinath, M.~Mustafa, A.~Albert, , and R.~Yu.
\newblock Towards physics-informed deep learning for turbulent flow prediction.
\newblock In \emph{Proceedings of the 26th ACM SIGKDD International Conference
  on Knowledge Discovery \& Data Mining}, pages 1457--1466, 2020.
\newblock \doi{https://doi.org/10.1145/3394486.3403198}.

\bibitem[Ranade et~al.(2021)Ranade, Hill, and
  Pathak]{ranade2021discretizationnet}
R.~Ranade, C.~Hill, and J.~Pathak.
\newblock Discretizationnet: A machine-learning based solver for navier--stokes
  equations using finite volume discretization.
\newblock \emph{Computer Methods in Applied Mechanics and Engineering},
  378:\penalty0 113722, 2021.
\newblock \doi{https://doi.org/10.1016/j.cma.2021.113722}.

\bibitem[Gao et~al.(2021)Gao, Sun, and Wang]{gao2021phygeonet}
H.~Gao, L.~Sun, and J.~Wang.
\newblock Phygeonet: Physics-informed geometry-adaptive convolutional neural
  networks for solving parameterized steady-state pdes on irregular domain.
\newblock \emph{Journal of Computational Physics}, 428:\penalty0 110079, 2021.
\newblock \doi{https://doi.org/10.1016/j.jcp.2020.110079}.

\bibitem[Rao et~al.({\natexlab{b}})Rao, Ren, Liu, and Sun]{rao2022discovering}
Chengping Rao, Pu~Ren, Yang Liu, and Hao Sun.
\newblock Discovering nonlinear pdes from scarce data with physics-encoded
  learning.
\newblock \emph{arXiv preprint arXiv:2201.12354}, {\natexlab{b}}.
\newblock \doi{https://doi.org/10.48550/arXiv.2201.12354}.

\bibitem[Wang(1994)]{wang1994deterministic}
J.~Wang.
\newblock A deterministic annealing neural network for convex programming.
\newblock \emph{Neural networks}, 7\penalty0 (4):\penalty0 629--641, 1994.
\newblock \doi{https://doi.org/10.1016/0893-6080(94)90041-8}.

\bibitem[Rangarajan et~al.(1996)Rangarajan, Gold, and
  Mjolsness]{rangarajan1996novel}
Anand Rangarajan, Steven Gold, and Eric Mjolsness.
\newblock A novel optimizing network architecture with applications.
\newblock \emph{Neural Computation}, 8\penalty0 (5):\penalty0 1041--1060, 1996.
\newblock \doi{https://doi.org/10.1162/neco.1996.8.5.1041}.

\bibitem[Cranmer et~al.(2020)Cranmer, Greydanus, Hoyer, Battaglia, Spergel, and
  Ho]{cranmer2020lagrangian}
M.~Cranmer, S.~Greydanus, S.~Hoyer, P.~Battaglia, D.~Spergel, and S.~Ho.
\newblock Lagrangian neural networks.
\newblock \emph{arXiv preprint arXiv:2003.04630}, 2020.
\newblock \doi{https://doi.org/10.48550/arXiv.2003.04630}.

\bibitem[Allen-Blanchette et~al.(2020)Allen-Blanchette, Veer, Majumdar, and
  Leonard]{allen2020lagnetvip}
C.~Allen-Blanchette, S.~Veer, A.~Majumdar, and N.~E. Leonard.
\newblock Lagnetvip: A lagrangian neural network for video prediction.
\newblock \emph{arXiv preprint arXiv:2010.12932}, 2020.
\newblock \doi{https://doi.org/10.48550/arXiv.2010.12932}.

\bibitem[Chen et~al.(2019{\natexlab{c}})Chen, Zhang, Arjovsky, and
  Bottou]{chen2019symplectic}
Z.~Chen, J.~Zhang, M.~Arjovsky, and L.~Bottou.
\newblock Symplectic recurrent neural networks.
\newblock \emph{arXiv preprint arXiv:1909.13334}, 2019{\natexlab{c}}.
\newblock \doi{https://doi.org/10.48550/arXiv.1909.13334}.

\bibitem[DiPietro et~al.(2020)DiPietro, Xiong, and Zhu]{dipietro2020sparse}
D.~DiPietro, S.~Xiong, and B.~Zhu.
\newblock Sparse symplectically integrated neural networks.
\newblock \emph{Advances in Neural Information Processing Systems},
  33:\penalty0 6074--6085, 2020.

\bibitem[Trask et~al.(2022)Trask, Huang, and Hu]{trask2022enforcing}
N.~Trask, A.~Huang, and X.~Hu.
\newblock Enforcing exact physics in scientific machine learning: a data-driven
  exterior calculus on graphs.
\newblock \emph{Journal of Computational Physics}, 456:\penalty0 110969, 2022.
\newblock \doi{https://doi.org/10.1016/j.jcp.2022.110969}.

\bibitem[Lin et~al.(2013)Lin, Chen, and Yan]{lin2013network}
M.~Lin, Q.~Chen, and S.~Yan.
\newblock Network in network.
\newblock \emph{arXiv preprint arXiv:1312.4400}, 2013.
\newblock \doi{https://doi.org/10.48550/arXiv.1312.4400}.

\bibitem[Shi et~al.(2015)Shi, Chen, Wang, Yeung, Wong, and
  Woo]{shi2015convolutional}
X.~Shi, Z.~Chen, H.~Wang, D.~Yeung, W.~Wong, and W.~Woo.
\newblock Convolutional lstm network: A machine learning approach for
  precipitation nowcasting.
\newblock \emph{Advances in neural information processing systems}, 28, 2015.

\bibitem[He et~al.(2016)He, Zhang, Ren, and Sun]{he2016deep}
Kaiming He, Xiangyu Zhang, Shaoqing Ren, and Jian Sun.
\newblock Deep residual learning for image recognition.
\newblock In \emph{Proceedings of the IEEE conference on computer vision and
  pattern recognition}, pages 770--778, 2016.

\bibitem[Ren et~al.(2022)Ren, Rao, Liu, Wang, and Sun]{ren2022phycrnet}
P.~Ren, C.~Rao, Y.~Liu, J.~Wang, and Hao Sun.
\newblock Phycrnet: Physics-informed convolutional-recurrent network for
  solving spatiotemporal pdes.
\newblock \emph{Computer Methods in Applied Mechanics and Engineering},
  389:\penalty0 114399, 2022.
\newblock \doi{https://doi.org/10.1016/j.cma.2021.114399}.

\bibitem[Baydin et~al.(2018)Baydin, Pearlmutter, Radul, and
  Siskind]{baydin2018automatic}
A.~G. Baydin, B.~A Pearlmutter, A.~A. Radul, and J.~M. Siskind.
\newblock Automatic differentiation in machine learning: a survey.
\newblock \emph{Journal of Marchine Learning Research}, 18:\penalty0 1--43,
  2018.

\bibitem[Rackauckas et~al.(2019)Rackauckas, Innes, Ma, Bettencourt, White, and
  Dixit]{rackauckas2019diffeqflux}
C.~Rackauckas, M.~Innes, Y.~Ma, J.~Bettencourt, L.~White, and V.~Dixit.
\newblock Diffeqflux. jl-a julia library for neural differential equations.
\newblock \emph{arXiv preprint arXiv:1902.02376}, 2019.
\newblock \doi{https://doi.org/10.48550/arXiv.1902.02376}.

\bibitem[Pontryagin(1987)]{pontryagin1987mathematical}
L.~S. Pontryagin.
\newblock \emph{Mathematical theory of optimal processes}.
\newblock CRC press, 1987.

\bibitem[Ma et~al.(2021)Ma, Dixit, Innes, Guo, and
  Rackauckas]{ma2021comparison}
Y.~Ma, V.~Dixit, M.~J. Innes, X.~Guo, and C.~Rackauckas.
\newblock A comparison of automatic differentiation and continuous sensitivity
  analysis for derivatives of differential equation solutions.
\newblock In \emph{2021 IEEE High Performance Extreme Computing Conference
  (HPEC)}, pages 1--9. IEEE, 2021.
\newblock \doi{https://doi.org/10.1109/HPEC49654.2021.9622796}.

\bibitem[Poli et~al.(2020)Poli, Massaroli, Yamashita, Asama, and
  Park]{poli2020torchdyn}
M.~Poli, S.~Massaroli, A.~Yamashita, H.~Asama, and J.~Park.
\newblock Torchdyn: A neural differential equations library.
\newblock \emph{arXiv preprint arXiv:2009.09346}, 2020.
\newblock \doi{https://doi.org/10.48550/arXiv.2009.09346}.

\bibitem[Lai et~al.(2021)Lai, Mylonas, Nagarajaiah, and
  Chatzi]{lai2021structural}
Z.~Lai, C.~Mylonas, S.~Nagarajaiah, and E.~Chatzi.
\newblock Structural identification with physics-informed neural ordinary
  differential equations.
\newblock \emph{Journal of Sound and Vibration}, 508:\penalty0 116196, 2021.
\newblock \doi{https://doi.org/10.1016/j.jsv.2021.116196}.

\bibitem[Roehrl et~al.(2020)Roehrl, Runkler, Brandtstetter, Tokic, and
  Obermayer]{roehrl2020modeling}
M.~A. Roehrl, T.~A. Runkler, V.~Brandtstetter, M.~Tokic, and S.~Obermayer.
\newblock Modeling system dynamics with physics-informed neural networks based
  on lagrangian mechanics.
\newblock \emph{IFAC-PapersOnLine}, 53\penalty0 (2):\penalty0 9195--9200, 2020.
\newblock \doi{https://doi.org/10.1016/j.ifacol.2020.12.2182}.

\bibitem[Dulny et~al.(2021)Dulny, Hotho, and Krause]{dulny2021neuralpde}
A.~Dulny, A.~Hotho, and A.~Krause.
\newblock Neuralpde: Modelling dynamical systems from data.
\newblock \emph{arXiv preprint arXiv:2111.07671}, 2021.
\newblock \doi{https://doi.org/10.48550/arXiv.2111.07671}.

\bibitem[K. et~al.(2016)K., X., S., and Jian]{he2016identity}
He~K., Zhang X., Ren S., and Sun Jian.
\newblock Identity mappings in deep residual networks.
\newblock In \emph{European conference on computer vision}, pages 630--645.
  Springer, 2016.

\bibitem[Goswami et~al.(2020)Goswami, Anitescu, Chakraborty, and
  Rabczuk]{goswami2020transfer}
S.~Goswami, C.~Anitescu, S.~Chakraborty, and T.~Rabczuk.
\newblock Transfer learning enhanced physics informed neural network for
  phase-field modeling of fracture.
\newblock \emph{Theoretical and Applied Fracture Mechanics}, 106:\penalty0
  102447, 2020.
\newblock \doi{https://doi.org/10.1016/j.tafmec.2019.102447}.

\bibitem[Bhattacharya et~al.(2020)Bhattacharya, Hosseini, Kovachk, and
  Stuart]{bhattacharya2020model}
K.~Bhattacharya, B.~Hosseini, N.B. Kovachk, and A.M. Stuart.
\newblock Model reduction and neural networks for parametric pdes.
\newblock \emph{arXiv preprint arXiv:2005.03180}, 2020.
\newblock \doi{https://doi.org/10.48550/arXiv.2005.03180}.

\bibitem[Li et~al.(2020{\natexlab{b}})Li, Kovachki, Azizzadenesheli, Liu,
  Bhattacharya, Stuart, and Anandkumar]{li2020neuralb}
Z.~Li, N.~Kovachki, K.~Azizzadenesheli, Burigede Liu, K.~Bhattacharya,
  A.~Stuart, and A.~Anandkumar.
\newblock Neural operator: Graph kernel network for partial differential
  equations.
\newblock \emph{arXiv preprint arXiv:2003.03485}, 2020{\natexlab{b}}.
\newblock \doi{https://doi.org/10.48550/arXiv.2003.03485}.

\bibitem[Migus et~al.(2022)Migus, Yin, Mazari, and Gallinari]{migus2022multi}
Leon Migus, Yuan Yin, Jocelyn~Ahmed Mazari, and Patrick Gallinari.
\newblock Multi-scale physical representations for approximating pde solutions
  with graph neural operators.
\newblock \emph{arXiv preprint arXiv:2206.14687}, 2022.
\newblock \doi{https://doi.org/10.48550/arXiv.2206.14687}.

\bibitem[Chen and Chen(1995)]{chen1995universal}
T.~Chen and H.~Chen.
\newblock Universal approximation to nonlinear operators by neural networks
  with arbitrary activation functions and its application to dynamical systems.
\newblock \emph{IEEE Transactions on Neural Networks}, 6\penalty0 (4):\penalty0
  911--917, 1995.
\newblock \doi{https://doi.org/10.1109/72.392253}.

\bibitem[Lin et~al.(2021)Lin, Li, Lu, Cai, Maxey, and
  Karniadakis]{lin2021operator}
C.~Lin, Z.~Li, L.~Lu, S.~Cai, M.~Maxey, and G.~E. Karniadakis.
\newblock Operator learning for predicting multiscale bubble growth dynamics.
\newblock \emph{The Journal of Chemical Physics}, 154\penalty0 (10):\penalty0
  104118, 2021.
\newblock \doi{https://doi.org/10.1063/5.0041203}.

\bibitem[Oommen et~al.(2022)Oommen, Shukla, Goswami, Dingreville, and
  Karniadakis]{oommen2022learning}
V.~Oommen, K.~Shukla, S.~Goswami, R.~Dingreville, and G.~E. Karniadakis.
\newblock Learning two-phase microstructure evolution using neural operators
  and autoencoder architectures.
\newblock \emph{arXiv preprint arXiv:2204.07230}, 2022.
\newblock \doi{https://doi.org/10.48550/arXiv.2204.07230}.

\bibitem[Wang et~al.(2021)Wang, Wang, and Perdikaris]{wang2021learning}
S.~Wang, H.~Wang, and P.~Perdikaris.
\newblock Learning the solution operator of parametric partial differential
  equations with physics-informed deeponets.
\newblock \emph{Science advances}, 7\penalty0 (40):\penalty0 eabi8605, 2021.
\newblock \doi{https://doi.org/10.1126/sciadv.abi8605}.

\bibitem[Goswami et~al.(2022{\natexlab{b}})Goswami, Yin, Yu, and
  Karniadakis]{goswami2022physicscrack}
S.~Goswami, M.~Yin, Y.~Yu, and G.~E. Karniadakis.
\newblock A physics-informed variational deeponet for predicting crack path in
  quasi-brittle materials.
\newblock \emph{Computer Methods in Applied Mechanics and Engineering},
  391:\penalty0 114587, 2022{\natexlab{b}}.
\newblock \doi{https://doi.org/10.1016/j.cma.2022.114587}.

\bibitem[DeVore(2017)]{devore2017theoretical}
R.A. DeVore.
\newblock The theoretical foundation of reduced basis methods.
\newblock \emph{Model reduction and approximation: theory and algorithms},
  15:\penalty0 137, 2017.

\bibitem[Zhu and Zabaras(2018)]{zhu2018bayesian}
Y.~Zhu and N.~Zabaras.
\newblock Bayesian deep convolutional encoder--decoder networks for surrogate
  modeling and uncertainty quantification.
\newblock \emph{Journal of Computational Physics}, 366:\penalty0 415--447,
  2018.
\newblock \doi{https://doi.org/10.1016/j.jcp.2018.04.018}.

\bibitem[Grady et~al.(2022)Grady, Khan, Louboutin, Yin, Witte, Chandra, Hewett,
  and Herrmann]{grady2022towards}
T.~J. Grady, R.~Khan, M.~Louboutin, Z.~Yin, P.~A. Witte, R.~Chandra, R.~J.
  Hewett, and F.~J. Herrmann.
\newblock Towards large-scale learned solvers for parametric pdes with
  model-parallel fourier neural operators.
\newblock \emph{arXiv preprint arXiv:2204.01205}, 2022.
\newblock \doi{https://doi.org/10.48550/arXiv.2204.01205}.

\bibitem[Bui et~al.(2018)Bui, Adjiman, Bardow, Anthony, Boston, Brown, Fennell,
  Fuss, Galindo, Hackett, and others.]{bui2018carbon}
M.~Bui, C.~S. Adjiman, A.~Bardow, E.~J. Anthony, A.~Boston, S.~Brown, P.~S.
  Fennell, S.~Fuss, A.~Galindo, L.~A. Hackett, and others.
\newblock Carbon capture and storage (ccs): the way forward.
\newblock \emph{Energy \& Environmental Science}, 11\penalty0 (5):\penalty0
  1062--1176, 2018.
\newblock \doi{https://doi.org/10.1039/C7EE02342A}.

\bibitem[Wen et~al.(2022)Wen, Li, Azizzadenesheli, Anandkumar, and
  Benson]{wen2022u}
G.~Wen, Z.~Li, K.~Azizzadenesheli, A.~Anandkumar, and S.M. Benson.
\newblock U-fno—an enhanced fourier neural operator-based deep-learning model
  for multiphase flow.
\newblock \emph{Advances in Water Resources}, 163:\penalty0 104180, 2022.
\newblock \doi{https://doi.org/10.1016/j.advwatres.2022.104180}.

\bibitem[You et~al.(2022{\natexlab{a}})You, Zhang, Ross, Lee, and
  Yu]{you2022learning}
H.~You, Q.~Zhang, C.J. Ross, C-H. Lee, and Y.~Yu.
\newblock Learning deep implicit fourier neural operators (ifnos) with
  applications to heterogeneous material modeling.
\newblock \emph{arXiv preprint arXiv:2203.08205}, 2022{\natexlab{a}}.
\newblock \doi{https://doi.org/10.48550/arXiv.2203.08205}.

\bibitem[Kovachki et~al.(2021)Kovachki, Li, Liu, Azizzadenesheli, Bhattacharya,
  Stuart, and Anandkumar]{kovachki2021neural}
N.~Kovachki, Z.~Li, B.~Liu, K.~Azizzadenesheli, K.~Bhattacharya, A.~Stuart, and
  A.~Anandkumar.
\newblock Neural operator: Learning maps between function spaces.
\newblock \emph{arXiv preprint arXiv:2108.08481}, 2021.
\newblock \doi{https://doi.org/10.48550/arXiv.2108.08481}.

\bibitem[Lu et~al.(2022)Lu, Meng, Cai, Mao, Goswami, Zhang, and
  Karniadakis]{lu2022comprehensive}
L.~Lu, X.~Meng, S.~Cai, Z.~Mao, S.~Goswami, Z.~Zhang, and G.~E. Karniadakis.
\newblock A comprehensive and fair comparison of two neural operators (with
  practical extensions) based on fair data.
\newblock \emph{Computer Methods in Applied Mechanics and Engineering},
  393:\penalty0 114778, 2022.
\newblock \doi{https://doi.org/10.1016/j.cma.2022.114778}.

\bibitem[You et~al.(2022{\natexlab{b}})You, Yu, D'Elia, Gao, and
  Silling]{you2022nonlocal}
H.~You, Y.~Yu, M.~D'Elia, T.~Gao, and S.~Silling.
\newblock Nonlocal kernel network (nkn): a stable and resolution-independent
  deep neural network.
\newblock \emph{arXiv preprint arXiv:2201.02217}, 2022{\natexlab{b}}.
\newblock \doi{https://doi.org/10.48550/arXiv.2201.02217}.

\bibitem[Molina et~al.(2020)Molina, Avelino, Morales, and
  Sucar]{molina2020causal}
A.~M. Molina, I.~F. Avelino, E.~F. Morales, and L.~E. Sucar.
\newblock Causal based q-learning.
\newblock \emph{Research in Computing Science}, 149:\penalty0 95--104, 2020.

\end{thebibliography}
\end{document}